\documentclass[letterpaper]{article} % DO NOT CHANGE THIS
\usepackage{aaai24}  % DO NOT CHANGE THIS
\usepackage{times}  % DO NOT CHANGE THIS
\usepackage{helvet}  % DO NOT CHANGE THIS
\usepackage{courier}  % DO NOT CHANGE THIS
\usepackage[hyphens]{url}  % DO NOT CHANGE THIS
\usepackage{graphicx} % DO NOT CHANGE THIS
\usepackage{natbib}  % DO NOT CHANGE THIS AND DO NOT ADD ANY OPTIONS TO IT
\usepackage{caption} % DO NOT CHANGE THIS AND DO NOT ADD ANY OPTIONS TO IT
\usepackage{algorithm}
\usepackage{algorithmic}

\usepackage{newfloat}
\usepackage{listings}
\DeclareCaptionStyle{ruled}{labelfont=normalfont,labelsep=colon,strut=off} % DO NOT CHANGE THIS
\floatstyle{ruled}
\newfloat{listing}{tb}{lst}{}
\floatname{listing}{Listing}
\usepackage[group-separator={,}]{siunitx}
\usepackage{amsmath}
\usepackage{amsfonts}

\usepackage[capitalize]{cleveref}
\crefname{section}{Sec.}{Secs.}
\Crefname{section}{Section}{Sections}
\Crefname{table}{Table}{Tables}
\crefname{table}{Tab.}{Tabs.}
\Crefname{figure}{Figure}{Figures}
\crefname{figure}{Fig.}{Figs.}
\Crefname{equation}{Equation}{Equations}
\crefname{equation}{Eq.}{Eqs.}

\usepackage{xspace}
\makeatletter
\DeclareRobustCommand\onedot{\futurelet\@let@token\@onedot}
\def\@onedot{\ifx\@let@token.\else.\null\fi\xspace}

\def\ie{\emph{i.e}\onedot}

\def\etal{\emph{et al}\onedot}
\makeatother

\usepackage{xcolor}
\usepackage{colortbl}
\usepackage{multirow}
\usepackage{tabularx}
\usepackage{booktabs} % To thicken table lines
\usepackage{makecell}
\usepackage{graphbox}
\usepackage{footmisc}
\usepackage{bm}
\usepackage{caption}
\usepackage{arydshln}
\usepackage{dashrule}
\usepackage{enumitem}
\setlist[itemize]{noitemsep,nolistsep}

\usepackage{algorithm}
\usepackage{algorithmic}
\usepackage[labelformat=simple]{subcaption}

\Crefname{figure}{Figure}{Figures}
\crefname{figure}{Fig.}{Figs.}
\Crefname{equation}{Equation}{Equations}
\crefname{equation}{Eq.}{Eqs.}
\Crefname{algorithm}{Algorithm}{Algorithms}
\crefname{algorithm}{Alg.}{Algs.}

\DeclareMathOperator*{\argmin}{arg\,min}

\newcommand{\ceil}[1]{\left\lceil #1 \right\rceil}

\usepackage{dsfont}
\newcommand{\R}{\mathbb{R}}

\newcommand{\Z}{\mathbb{Z}_{0}}

\newcommand{\codebook}{\mathcal{C}}
\newcommand{\loss}{\mathcal{L}}
\newcommand{\token}{S}

\newcommand{\bs}{\textbf{s}}
\newcommand{\sz}{\textbf{s}_0}
\newcommand{\stm}{\textbf{s}_{t-1}}
\newcommand{\st}{\textbf{s}_t}
\newcommand{\sT}{\textbf{s}_T}
\newcommand{\mt}{\mathbf{m}_t}
\newcommand{\mtm}{\mathbf{m}_{t-1}}

\colorlet{lightpink}{pink!35}
\colorlet{lightcyan}{cyan!20}
\colorlet{red}{red!80}
\colorlet{blue}{blue!80}
\colorlet{green}{green!60!black}
\colorlet{algemp}{cyan!10}

\colorlet{gan}{teal!20}
\colorlet{dm}{pink!35}

\usepackage{pifont}

\newcommand{\best}[1]{\textcolor{red}{\textbf{#1}}}
\newcommand{\second}[1]{\textcolor{blue}{\textbf{#1}}}

\newcolumntype{C}[1]{>{\centering\arraybackslash}p{#1}}
\newcolumntype{L}[1]{>{\raggedleft\arraybackslash}p{#1}}
\newcolumntype{R}[1]{>{\raggedright\arraybackslash}p{#1}}

\renewcommand{\paragraph}[1]{\vspace{2pt} \noindent \textbf{#1}}
\newcommand{\red}[1]{\textcolor{red}{#1}}

\newcommand{\RNum}[1]{\uppercase\expandafter{\romannumeral #1\relax}}

\newcommand{\Comment}[1]{\hfill{$\triangleright$ #1}}

\graphicspath{{figures/}}
\usepackage[switch]{lineno}
\title{Iterative Token Evaluation and Refinement for Real-World Super-Resolution}
\author{
    Chaofeng Chen\textsuperscript{\rm 1}, \,
    Shangchen Zhou\textsuperscript{\rm 1}, \,
    Liang Liao\textsuperscript{\rm 1}, \,
    Haoning Wu\textsuperscript{\rm 1}, \, \\
    Wenxiu Sun\textsuperscript{\rm 2}, \,
    Qiong Yan\textsuperscript{\rm 2}, \,
    Weisi Lin\textsuperscript{\rm 1}
}
\affiliations{
    \textsuperscript{\rm 1} S-Lab, Nanyang Technological University \quad
    \textsuperscript{\rm 2} SenseTime Research \\
    {\textcolor{red}{\tt\small \url{https://github.com/chaofengc/ITER}}}
}

\begin{document}

\twocolumn[{%
\renewcommand\twocolumn[1][]{#1}%
\maketitle
\begin{center}
    \centering
    \vspace{-1.3cm} 
    \captionsetup{type=figure}
    \includegraphics[width=.99\textwidth]{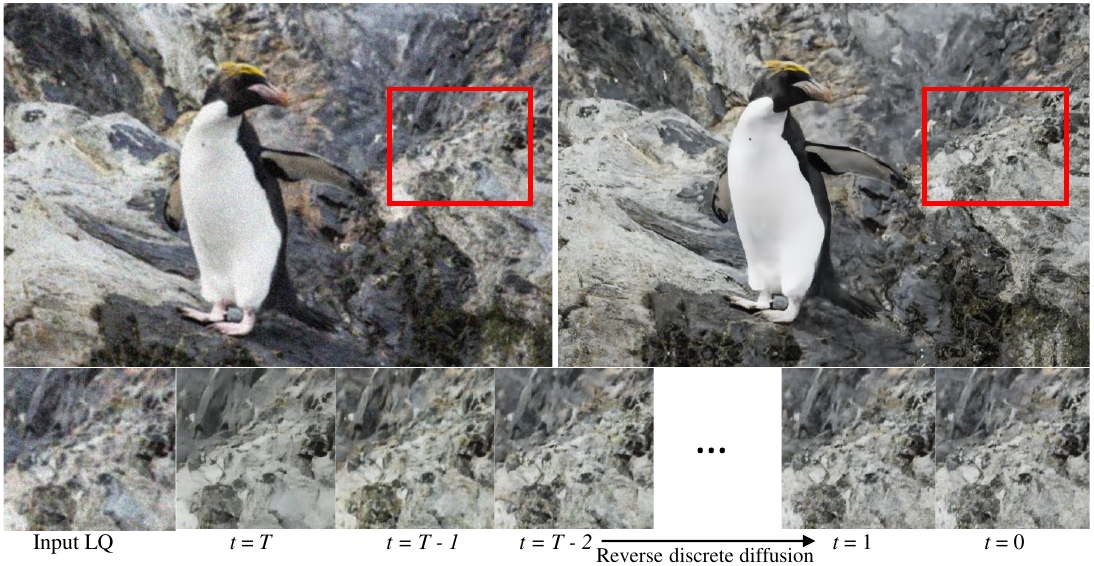} 
    \captionof{figure}{Example result with the proposed ITER. Left top: input LQ image; Right top: SR result with ITER. $t$ is the iterative step index of the reverse discrete diffusion process, and $t=T$ is the initial distortion removal result. The textures are gradually enriched with iterative refinement. To obtain satisfactory results, our ITER requires only a total iteration step of $T\leq8$.} \label{fig:teaser}
\end{center}%
}]

%%%%%%%%% ABSTRACT
\begin{abstract}
    Real-world image super-resolution (RWSR) is a long-standing problem as low-quality (LQ) images often have complex and unidentified degradations. Existing methods such as Generative Adversarial Networks (GANs) or continuous diffusion models present their own issues including GANs being difficult to train while continuous diffusion models requiring numerous inference steps. In this paper, we propose an Iterative Token Evaluation and Refinement (ITER) framework for RWSR, which utilizes a discrete diffusion model operating in the discrete token representation space, \ie, indexes of features extracted from a VQGAN codebook pre-trained with high-quality (HQ) images. We show that ITER is easier to train than GANs and more efficient than continuous diffusion models. Specifically, we divide RWSR into two sub-tasks, \textit{i.e.}, distortion removal and texture generation. Distortion removal involves simple HQ token prediction with LQ images, while texture generation uses a discrete diffusion model to iteratively refine the distortion removal output with a token refinement network. In particular, we propose to include a token evaluation network in the discrete diffusion process. It learns to evaluate which tokens are good restorations and helps to improve the iterative refinement results. Moreover, the evaluation network can first check status of the distortion removal output and then adaptively select total refinement steps needed, thereby maintaining a good balance between distortion removal and texture generation. Extensive experimental results show that ITER is easy to train and performs well within just 8 iterative steps. Our codes will be available publicly.
\end{abstract}

%%%%%%%%% BODY TEXT
\section{Introduction} \label{sec:intro}

Single-image super-resolution (SISR) aims to restore high-quality (HQ) outputs from low-quality (LQ) inputs that have been degraded through processes such as downsampling, blurring, noise, and compression. Previous studies \cite{liang2021swinir,Zamir2021Restormer,chen2023hat} have achieved remarkable progress in enhancing LQ images degraded by a single predefined type of degradation, thanks to the emergence of increasingly powerful deep networks. However, in real-world LQ images, multiple unknown degradations are typically present, making previous methods unsuitable for such complex scenarios.

Real-world super-resolution (RWSR) is particularly ill-posed because details are usually corrupted or completely lost due to complex degradations. In general, the RWSR can be divided into two subtasks: \textit{distortion removal} and \textit{conditioned texture generation}. Many existing approaches, such as \cite{wang2018esrgan,zhang2019ranksrgan}, follow the seminal SRGAN \cite{ledig2017srgan} and rely on Generative Adversarial Networks (GANs). Typically, these methods require the joint optimization of various constraints for the two subtasks: 1) reconstruction loss for distortion removal, which is usually composed of pixel-wise L1/L2 loss and feature space perceptual loss; 2) adversarial loss for texture generation. Effective training of these models often involves tedious fine-tuning of hyper-parameters between restoration and generation abilities. Moreover, most models have a fixed preference for restoration and generation and cannot be flexibly adapted to LQ inputs with different degradation levels. Recently, approaches such as SR3 \cite{saharia2022sr3} and LDM \cite{rombach2022latentdiffusion} have turned to the popular diffusion model (DM) for realistic generative ability. Although DMs are easier to train and more powerful than GANs, they require hundreds or even thousands of iterative steps to generate outputs. Additionally, current DM-based methods have only been shown to be effective on images with moderate distortions. Their performance on severely distorted real-world LQ images remains to be validated. 

In this paper, we introduce a new framework for RWSR based on a conditioned discrete diffusion model, called Iterative Token Evaluation and Refinement (ITER). ITER incorporates several critical designs to address the challenges of RWSR. Firstly, we formulate the RWSR task as a discrete token space problem, utilizing a pretrained codebook of VQGAN \cite{esser2021taming}, instead of pixel space regression. This approach offers two advantages: 1) A small discrete proxy space reduces the ambiguity of image restoration, as demonstrated in \cite{zhou2022codeformer}; 2) Generative sampling in a limited discrete space requires fewer iteration steps than denoising diffusion sampling in an infinite continuous space, as shown in \cite{bond2021unleashing,gu2021vqdm,chang2022maskgit}. Secondly, in contrast to previous GAN and DM methods, we explicitly separate the two sub-tasks of RWSR and address them with token restoration and token refinement modules, respectively. For the first task, we use a simple token restoration network to predict HQ tokens from LQ images. For the second task, we use a conditioned discrete diffusion model to iteratively refine outputs from the token restoration network. This approach facilitates optimizing each module and enables flexible trade-offs between restoration and generation. Finally, and most importantly, we propose to include a token evaluation block in the condition diffusion process. Unlike previous discrete diffusion models \cite{bond2021unleashing,chang2022maskgit} which directly rely on token prediction probability to select tokens to keep in each de-masking step, we introduce a evaluation block to check whether each tokens are correctly refined or not. This allows our model to better select good tokens in each step during iterative refinement process, and therefore improve the final results. Additionally, the token evaluation block enables us to adaptively select the total refinement steps to balance restoration and texture generation by evaluating the initially restored tokens. We can use fewer refinement steps for good initial restoration results to avoid over-textured outputs.
The experiments demonstrate that our proposed ITER framework can effectively remove distortions and generate realistic textures without tedious GAN training in an efficient manner, requiring \emph{less than 8 iterative refinement steps.} Please refer to \cref{fig:teaser} for an example. In summary, our contributions are as follows:

\begin{itemize}
    \item We propose a novel framework, ITER, that addresses the two sub-tasks of RWSR in discrete token space. Compared to GAN, ITER is much easier to train and more flexible at inference time. Compared to DM-based methods, it requires fewer iteration steps and has demonstrated effectiveness on real-world LQ inputs with complex degradations.
    \item We propose an iterative evaluation and refinement approach for texture generation. The newly introduced token evaluation block allows the model to make better decisions on which tokens to refine during the iterative refinement process. Furthermore, by evaluating the quality of initially restored tokens, ITER is able to adaptively balance distortion removal and the texture generation in the final results by using different refinement steps. Besides, the user can also manually control the visual effects of outputs through a threshold value without the need for retraining the model.
\end{itemize}
   
\section{Related Works}

In this section, we provide a brief overview of SISR and generative models utilized in SR. We also recommend recent literature reviews \cite{anwar2020deepsurvey,liu2022blindsrsurvey,liu2023survey} for more comprehensive summaries.

\paragraph{Single Image Super-Resolution.} Recent SISR for bicubic downsampled LQ images has made remarkable progress with the improvement of network architectures. Methods such as \cite{kim2016accurate,kim2016deeply,lim2017enhanced,ledig2017srgan,zhang2018residual} introduced deeper and wider networks with more skip connections, showing the power of residual learning \cite{he2016deep}. Attention mechanisms, including channel attention \cite{zhang2018image}, spatial attention \cite{niu2020HAN,ChenSPARNet}, and non-local attention \cite{zhang2019rnan,Mei_2021_CVPR_NLSA,zhou2020cross}, have also been found to be beneficial. Recent works employing vision transformers \cite{chen2020IPT,liang2021swinir,zhang2022ELAN,chen2023hat} have surpassed CNN-based networks by a large margin, thanks to the ability to model relationships in a large receptive field.

Latest works have focused more on the challenging task of RWSR. Some methods \cite{fritsche2019frequency,wei2021unsupervised,wan2020bringing,maeda2020unpaired,ji2020realsr,wang2021unsupervised,zhang2021blind,mou2022mmrealsr,jie2022DASR} implicitly learn degradation representations from LQ inputs and perform well in distortion removal. However, their generalization ability is limited due to the complexity of the real-world degradation space. BSRGAN~\cite{zhang2021designing} and Real-ESRGAN~\cite{wang2021real} adopt manually designed large degradation spaces to synthesize LQ inputs and have proven to be effective. Li \etal \cite{Li_2022_ReDegNet} proposed learning degradations from real LQ-HQ face pairs and then synthesizing training datasets. Although these methods improve distortion removal, they rely on unstable adversarial training to generate missing details, which may result in unrealistic textures.

\paragraph{Generative Models for Super-Resolution.} Many works employ GAN networks to generate missing textures for real LQ images. StyleGAN~\cite{karras2020analyzing} works well for real face SR \cite{yang2021gan,wang2021towards,chan2021glean}. Pan \etal \cite{pan2020exploiting} used a BigGAN generator \cite{brock2018biggan} for natural image restoration. The recent VQGAN\cite{esser2021taming} demonstrates superior performance in image synthesis and is shown to be effective in real SR of both face \cite{zhou2022codeformer} and natural images \cite{chen2022femasr}. 

The latest works with diffusion models \cite{saharia2022sr3,rombach2022latentdiffusion,gao2023implicit,wang2023stablesr} are more powerful than GAN, but they are based on continuous feature space and require many iterative sampling steps. In this work, we take advantage of the discrete diffusion models \cite{gu2021vqdm,bond2021unleashing,chang2022maskgit}, which is powerful in texture generation and efficient at inference time. To the best of our knowledge, we are the first work to show the potential of discrete diffusion models on image restoration.

\section{Methodology}

In this work, we propose a new iterative token sampling approach for texture generation in RWSR. Our pipeline operates in the discrete representation space pre-trained by VQGAN, which has been shown to be effective in image restoration \cite{chen2022femasr,zhou2022codeformer}. Our framework consists of three stages:
\begin{itemize}[leftmargin=10pt]
    \item \textbf{Stage I: HQ images to discrete tokens.} Different from previous works based on continuous latent diffusion models, our method is based on discrete latent space. Therefore, we need to pretrain a vector-quantized auto-encoder (VQVAE) \cite{esser2021taming} with discrete codebook to encode input HQ images $I_h$, such that $I_h$ can be transformed to discrete tokens, denoted as $S_h$.  
    \item \textbf{Stage II: LQ images to tokens with distortion removal.} Instead of directly encoding LQ images $I_l$ with pretrained VQVAE, we propose to train a separate distortion removal encoder for $I_l$. It helps to remove obvious distortions in LQ input $I_l$ and encode it to a relatively clean discrete token space $S_l$. 
    \item \textbf{Stage III: Texture generation with discrete diffusion.} After obtaining the discrete representations $S_l$ and $S_h$, we formulate the texture generation as a discrete diffusion model between $S_l$ and $S_h$. The key difference with our method is that we include an additional token evaluation block to improve the decision-making process for which tokens to refine during the reverse diffusion process. In such manner, the proposed ITER not only generates realistic textures but also permits adaptable control over the texture strength in the final output.
\end{itemize}
Details are given in the following sections.

\subsection{HQ images to discrete tokens} \label{sec:vqgan-swin}

Following VQGAN \cite{esser2021taming}, the encoder $E_H$ takes the input high-quality (HQ) image $I_h\in \R^{H \times W \times 3}$ in RGB space and encodes it to latent features $Z_h \in \R^{m \times n \times d}$. Subsequently, $Z_h$ is quantized into discrete features $Z_c \in \R^{m \times n \times d}$ by identifying its nearest neighbors in the learnable codebook $\codebook=\{c_k \in \R^d\}_{k=0}^{N-1}$:
\begin{equation}
Z_c^{(i, j)} = \argmin_{c_k \in \codebook} \| Z_h^{(i, j)} - c_k\|_2.
\end{equation}
The corresponding indices $k \in \{0, \ldots, N - 1\}$ determine the token representation of the inputs $\token_h \in \Z^{m\times n}$. Finally, the decoder reconstructs the image from the latent $I_{rec} = D_H(Z_c) = D_H(E_H(I_h))$. Instead of using the original VQGAN \cite{esser2021taming}, we replace the non-local attention with Swin Transformer blocks \cite{liu2021swin} to reduce memory cost for large resolution inputs. More details can be found in the supplementary material.

\subsection{LQ images to tokens with distortion removal} \label{sec:dist-remove}

\begin{figure}[!t]
    \centering
    \includegraphics[width=0.99\linewidth]{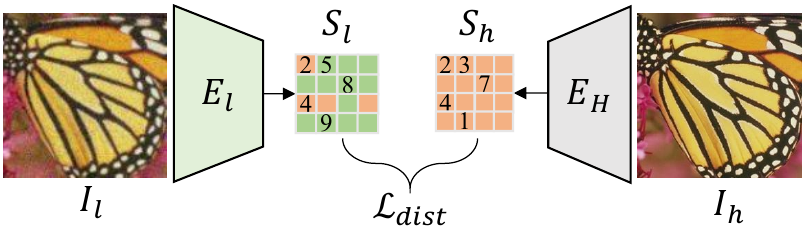}
    \caption{Training of $E_l$ to encode $I_l$ to token space $S_l$.}
    \label{fig:token_classify}
\end{figure}

It is straightforward to also encode $I_l$ with pretrained $E_H$ in the first stage. However, since $I_l$ contains complex distortions, the encoded tokens are also noisy, increasing the difficulties of restoration in the following stage. Inspired by recent works \cite{chen2022femasr, zhou2022codeformer}, we realize that a straightforward token prediction can eliminate evident distortions. Hence, we introduce a preprocess subtask to remove distortions when encoding $I_l$ into token space. Specifically, we employ an LQ encoder $E_l$ to directly predict the HQ code indexes $S_h$ as illustrated in \cref{fig:token_classify}:
\begin{equation}
S_l = E_l(I_l), \quad \loss_{dist} = - S_h^i \log (S_l^i), \label{eq:token_pred}
\end{equation}
Through this approach, $I_l$ can be encoded into a comparatively clean token space with the learned $E_l$. 

\subsection{Texture generation with discrete diffusion} \label{sec:iter}

Although the distortions in $S_l$ are effectively removed, generating missing details through \cref{eq:token_pred} is a challenging task because the generation of diverse natural textures is highly ill-posed and essentially a one-to-many endeavor. To address this issue, we propose an iterative token evaluation and refinement approach, named as ITER, for RWSR, following the generative sampling pipeline outlined in \cite{chang2022maskgit,lezama2022tokencritic}. As ITER is based on the discrete diffusion model \cite{bond2021unleashing,gu2021vqdm}, we will first provide a brief overview of it.

\paragraph{Discrete Diffusion Model.} Given an initial image token $\textbf{s}_0 \in \Z$, the forward diffusion process establishes a Markov chain $q(\bs_{1:T}|\sz) = \prod_{t=1}^T q(\st|\stm)$, which progressively corrupts $\textbf{s}_0$ by randomly masking $\sz$ over $T$ steps until $\sT$ is entirely obscured. Conversely, the reverse process is a generative model that incrementally ``unmasks'' $\sT$ to the data distribution $p(\bs_{0:T}) = p(\sT) \prod_{t=1}^T p_\theta(\stm|\st)$. According to \cite{bond2021unleashing, chang2022maskgit, lezama2022tokencritic}, the ``unmasking'' transit distribution $p_\theta$ can be approximated by learning to predict the authentic $\sz$, given any arbitrarily masked version $\st$:
\begin{equation}
\argmin_{\theta} -\log p_{\theta}(\sz|\st).
\end{equation}
Following \cite{chang2022maskgit}, during the forward process, $\st$ is obtained by randomly masking $\sz$ at a ratio of $\gamma(r)$, where $r \in \text{Uniform}(0, 1]$, and $\gamma(\cdot)$ represents the mask scheduling function. In the reverse process, $\st$ is sampled according to the prediction probability $p_\theta(\st|\mathbf{s}_{t+1}, \sT)$. The masking ratio is computed using the predefined total sampling step $T$, \ie, $\gamma(\frac{t}{T})$ where $t \in \{T, \ldots, 1\}$. 

\begin{figure}[!t]
    \centering
    \includegraphics[width=\linewidth]{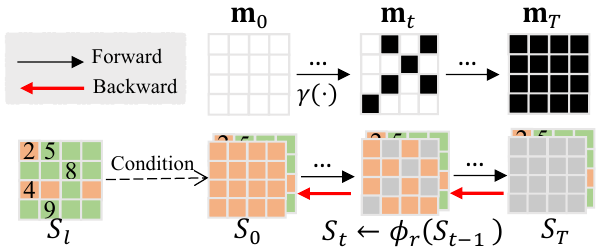}
    \caption{Illustration of forward and backward diffusion process with the conditioned discrete diffusion model. The condition inputs of $\phi_r$ are omitted here for simplicity.}
    \label{fig:iter_diffusion}
\end{figure}

\begin{algorithm}[t]
    \caption{Training of ITER} \label{alg:train}
    \begin{flushleft}
    \textbf{Input:} $S_l$, $S_h$, schedule function $\gamma(\cdot)$, learning rate $\eta$, networks $\phi_r$ and $\phi_e$ 
    \end{flushleft}
    \begin{algorithmic}[1]
    \REPEAT
    \STATE $r \sim \text{Uniform}(0, 1]$
    \STATE $N \gets$ token numbers in $S_h$
    \STATE $\mt \gets \text{RandomMask}(\ceil{\gamma(r) \cdot N})$
    \STATE $S_t \gets S_h \odot \mt + (1 - \mt) \odot S_T$
    \STATE $\theta_r \gets \theta_r - \eta \nabla_{\theta_r} \loss_r$ \Comment{Update $\phi_r$}
    \STATE $\theta_e \gets \theta_e - \eta \nabla_{\theta_e} \loss_e$ \Comment{Update $\phi_e$}
    \UNTIL{converge}
    \end{algorithmic}
\end{algorithm}

\begin{algorithm}[t]
    \caption{Adaptive Inference of ITER} \label{alg:test}
    \begin{flushleft}
    \textbf{Input:} $I_l, T=8, \gamma(\cdot)$, networks $E_l$, $D_H$, $\phi_r$ and $\phi_e$
    \end{flushleft}
    \begin{algorithmic}[1]
    \STATE $S_l \gets E_l(I_l)$ \Comment{Initial restoration}
    \STATE $N \gets$ token numbers in $S_l$
    \STATE $T_s \gets$ $T$ \newline
    \hspace*{-\fboxsep}\colorbox{algemp}{\parbox{\linewidth}{%
    \IF{use adaptive inference}
        \STATE $\mathbf{m}_s \gets \phi_e(S_l)$ with $\alpha$, \cref{eq:ms}
        \WHILE{$\ceil{\Bigl (1 - \gamma \bigl( \frac{T_s - 1}{T} \bigr)\Bigr) \cdot N} < \sum \mathbf{m}_s$}
        \STATE $T_s \gets$ $T_s - 1$ \Comment{Find start time step}
        \ENDWHILE
        \STATE Initialize with \cref{eq:adaptive_init}
    \ENDIF
    }}
    \FOR{$t = T_s \cdots 1$}
    \STATE $k \gets \ceil{\Bigl (1 - \gamma \bigl( \frac{t - 1}{T} \bigr)\Bigr) \cdot N}$  \Comment{Number to sample}
    \STATE $S_{t-1} \gets \text{sample } p_{\phi_r}(S_{t-1}|S_t, S_l, \mt)$ \Comment{Refine} 
    \STATE $\mtm \gets \text{sample}~k~\text{from}~p_{\phi_e}(\mtm=1|S_{t-1})$ \Comment{Evaluate}
    \STATE $S_{t-1} \gets S_{t-1} \odot \mtm + S_T \odot (1 - \mtm)$  
    \ENDFOR \\
    \RETURN $I_{sr} \gets D_H(S_0)$ \Comment{Get SR result.}
    \end{algorithmic}
\end{algorithm}

\paragraph{Network Training.} As depicted in \cref{fig:iter_diffusion}, the proposed ITER model is a conditioned version of the discrete diffusion model. It is a Markov chain that goes from ground truth tokens $S_h$ (\ie, $S_0$) to fully masked tokens $S_T$ while being conditioned on $S_l$. The reverse diffusion step $p_\theta(\stm|\st)$ is learned with the refinement network $\phi_r$ using the following objective function:
\begin{align}
    \loss_r = - S_h \log \bigl( \phi_r(S_t, S_l, \mt) \bigr), \label{eq:loss_refine}
\end{align}
where $\mt$ is the random mask in corresponding forward diffusion step, and tells $\phi_e$ which tokens need to be refined.

The difference is that we introduce an extra token evaluation network $\phi_e$ to learn which tokens are good tokens for both $S_t$ and $S_l$ with the objective function below:
\begin{align}
    \loss_e = - \mt \log \bigl( \phi_e(S_t) \bigr) - \mathbf{m}_l \log \bigl( \phi_e(S_l) \bigr), \label{eq:loss_eva} 
\end{align}
where $\mathbf{m}_l$ are the ground truth sampling masks for $S_l$. 

\subsection{Adaptive inference of ITER} \label{sec:adaptive} 

As illustrated in \Cref{alg:test}, the inference process of ITER can be a standard reverse diffusion from $S_T$ to $S_0$ with the condition $S_l$. However, in our framework, the initially restored tokens $S_l$ already contain good tokens and may not require the entire reverse process. With the aid of the token evaluation network $\phi_e$, it is possible to select the appropriate starting time step $T_s$ for the reverse diffusion process by assessing the number of good tokens in $S_l$ using $\mathbf{m}_l = \phi_e(S_l)$, as shown below:
\begin{equation}
    \mathbf{m}_s^i = \left \{ \begin{array}{cc}
          1 & \mbox{\text{if} $p_{\phi_e}(\mathbf{m}_l^i=1) \geq \alpha$}; \\ 
          0 & \mbox{otherwise}. 
    \end{array} \right. \label{eq:ms}
\end{equation}
where $\alpha$ is the threshold value, and $\mathbf{m}_s$ is the binary mask for the starting time step $T_s$. We can quickly determine the appropriate $T_s$ by comparing the mask ratio indicated by $\gamma(\cdot)$, see \Cref{alg:test} for further details. We can then initialize $S_t$ and $\mt$ using the following equations:
\begin{equation}
S_t=\mathbf{m}_s \odot S_l + (1 - \mathbf{m}_s) \odot S_T, \quad \mt=\mathbf{m}_s. \label{eq:adaptive_init}
\end{equation}
Finally, we follow the typical reverse diffusion process to compute the ``unmasking" distribution $p_{\phi_r}$, where $t \in \{T_s, \ldots, 1\}$. The final outcome is obtained by $I{sr}=D_H(S_0)$. The proposed adaptive inference strategy not only makes ITER more efficient but also avoids disrupting the initial good tokens in $S_l$.

\begin{figure*}[!t]
    \centering
    \newcommand{\imgwidth}{0.196\linewidth}
    \includegraphics[width=\imgwidth]{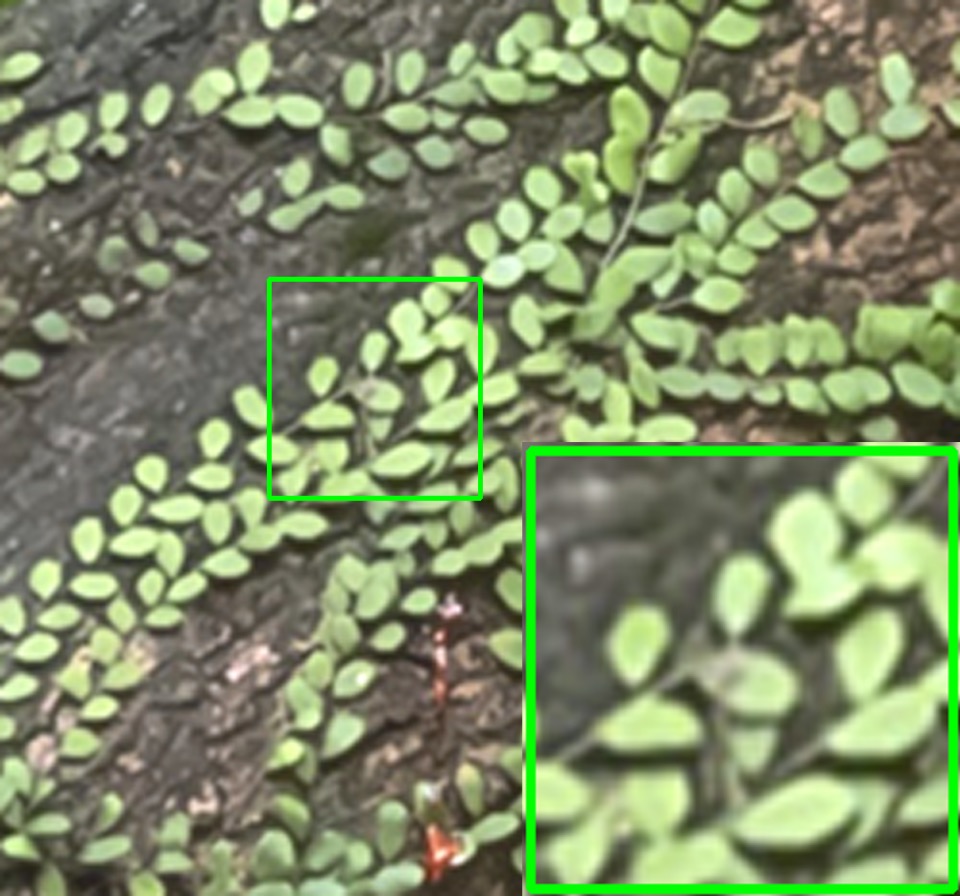}
    \includegraphics[width=\imgwidth]{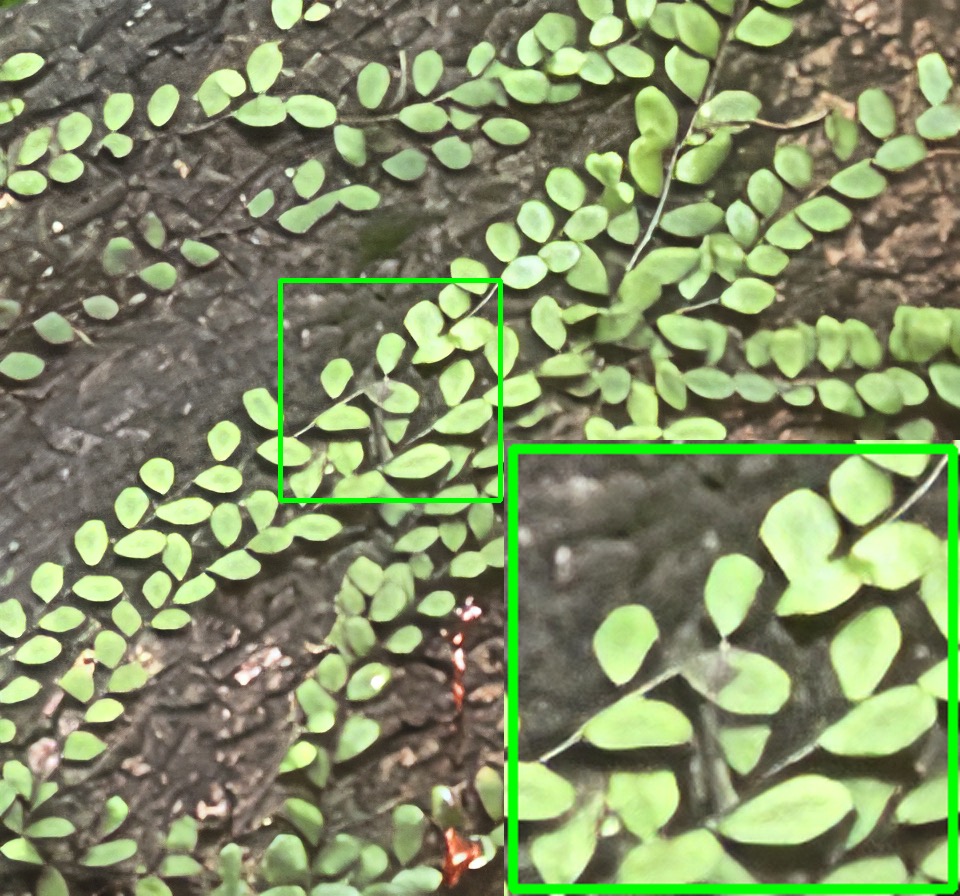}
    \includegraphics[width=\imgwidth]{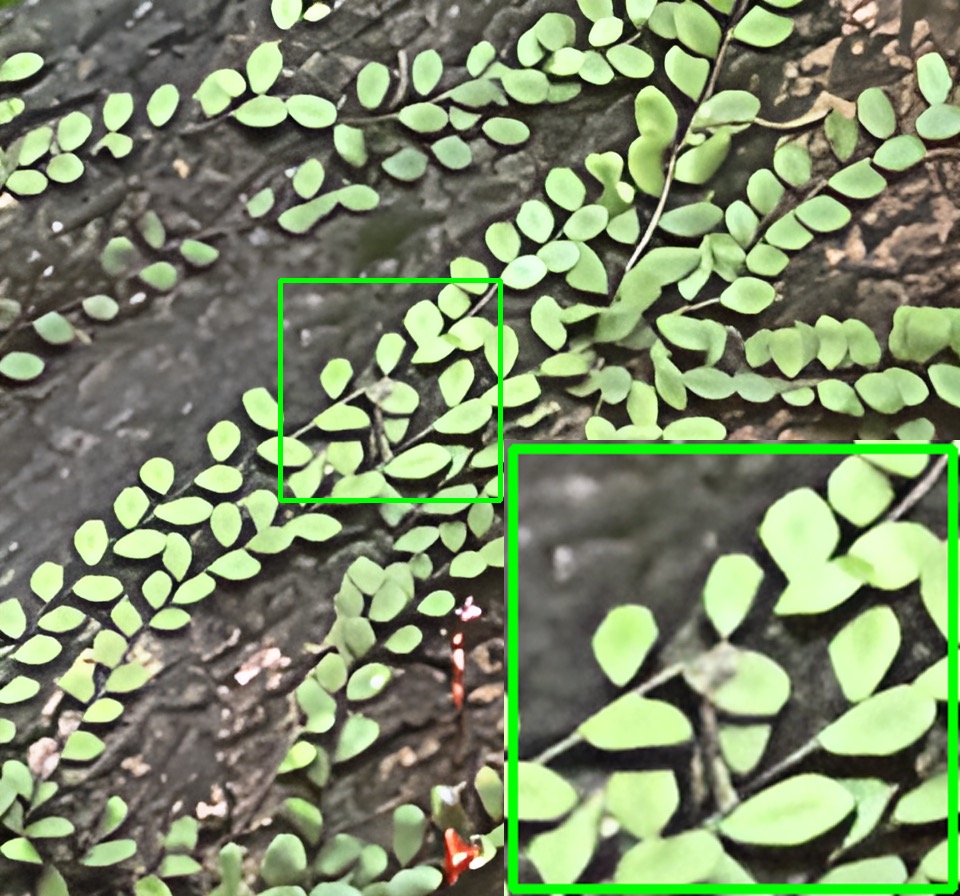}
    \includegraphics[width=\imgwidth]{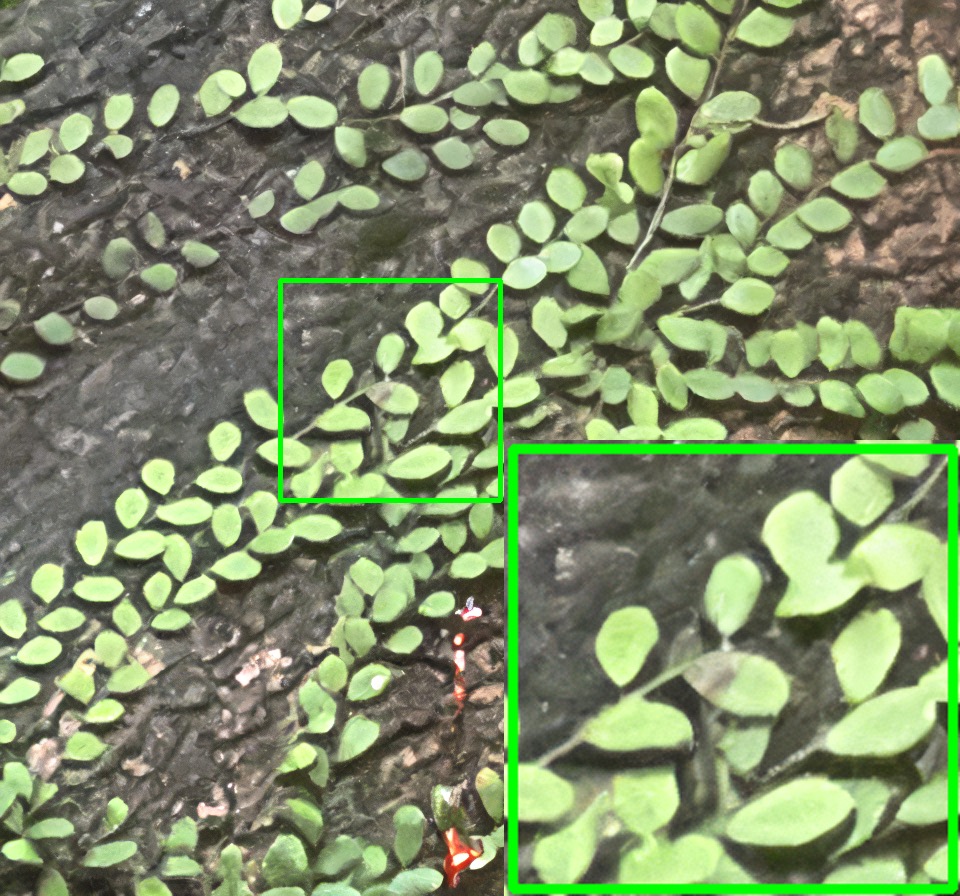}
    \includegraphics[width=\imgwidth]{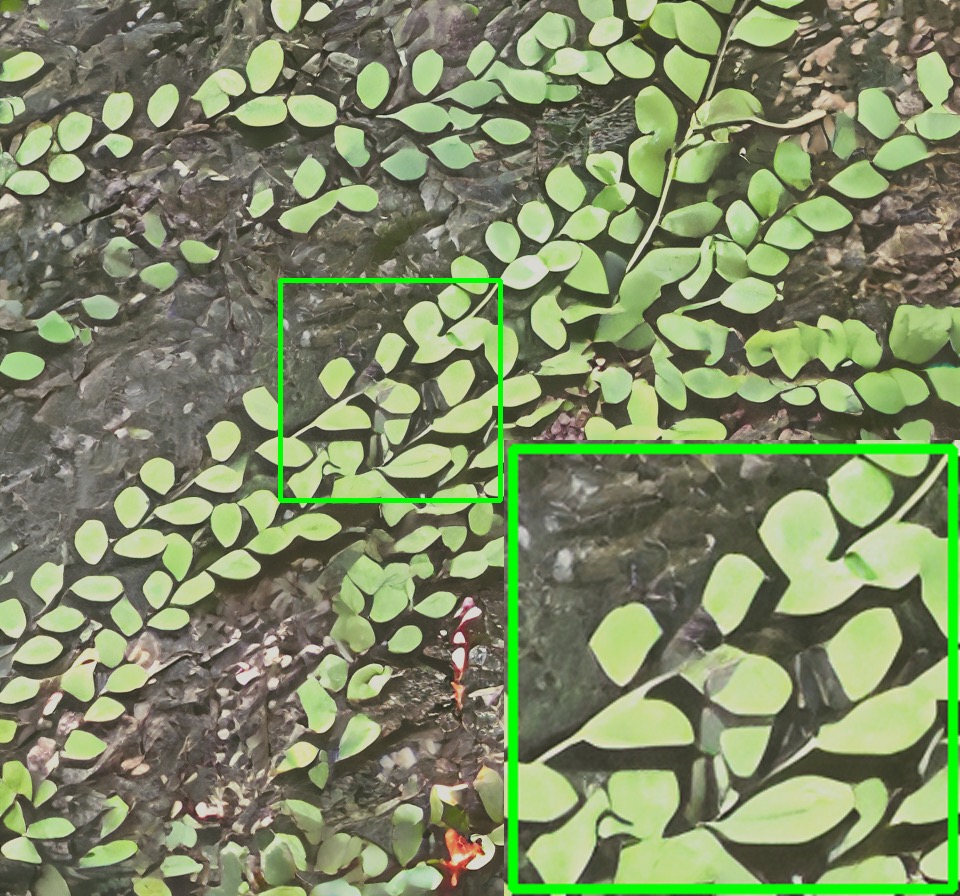}
    \\
    \includegraphics[width=\linewidth]{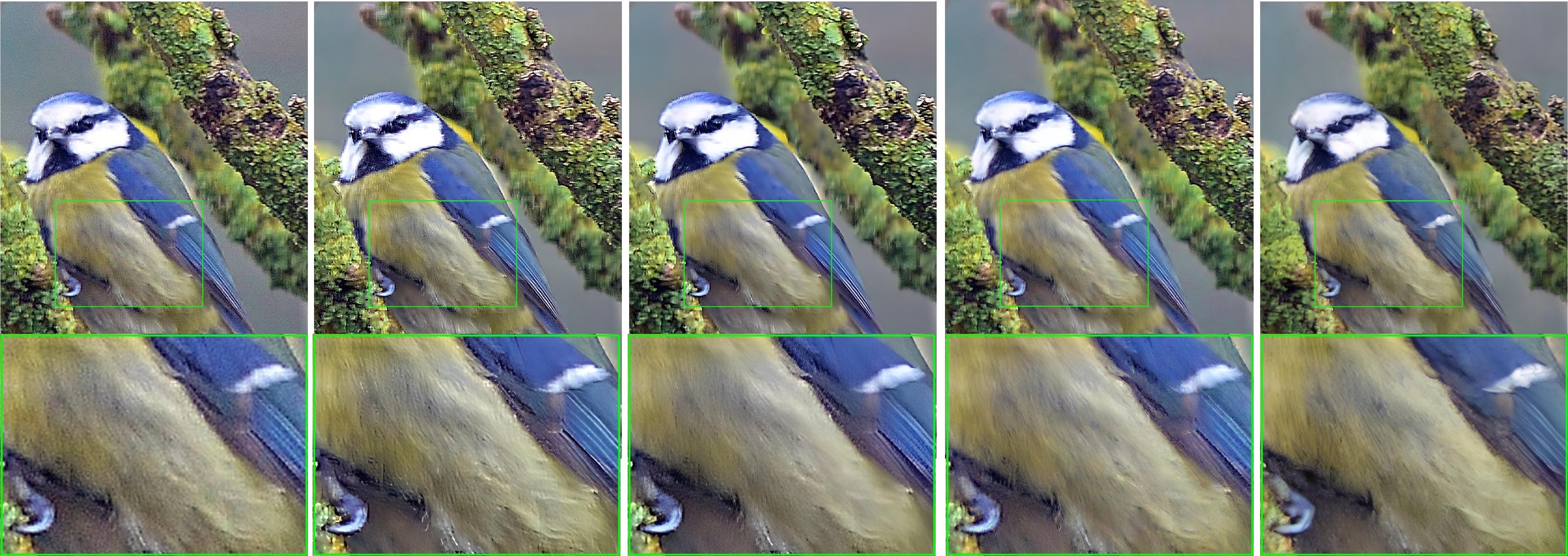}
    \\
    \makebox[\imgwidth]{\small (a) LQ ($\times4$)}
    \makebox[\imgwidth]{\small (b) FeMaSR}
    \makebox[\imgwidth]{\small (c) MM-RealSR}
    \makebox[\imgwidth]{\small (d) LDM-BSR}
    \makebox[\imgwidth]{\small (e) \textbf{ITER (Ours)}}
    \caption{Visual comparison between recent approaches and the proposed ITER on real LQ images. More examples are in supplementary material. \red{Please zoom in for best view.}}
    \label{fig:vis_comp}
\end{figure*}

\begin{table*}[t]
\centering
\caption{Quantitative comparison (NIQE $\downarrow$ and PI $\downarrow$) on real-world benchmarks. The best and second performance are marked in \red{red} and \textcolor{blue}{blue}. Results of BSRGAN and Real-ESRGAN are taken from \cite{wang2021real}, and others are tested with official codes.} \label{tab:real_benchmarks}
\resizebox{1.0\textwidth}{!}{
\renewcommand{\arraystretch}{1.2}
\begin{tabular}{|c|*{8}{cc|}}
\hline
\multirow{2}{*}{Datasets} & \multicolumn{2}{c|}{Bicubic} & \multicolumn{2}{c|}{\cellcolor{gan}BSRGAN} & \multicolumn{2}{c|}{\cellcolor{gan}Real-ESRGAN} & \multicolumn{2}{c|}{\cellcolor{gan}SwinIR-GAN} & \multicolumn{2}{c|}{\cellcolor{gan}FeMaSR} & \multicolumn{2}{c|}{\cellcolor{gan}MM-RealSR} & \multicolumn{2}{c|}{\cellcolor{dm}LDM-BSR} & \multicolumn{2}{c|}{\cellcolor{dm}\textbf{Ours}} \\ \cline{2-17}
 & NIQE & PI & NIQE & PI & NIQE & PI & NIQE & PI & NIQE & PI & NIQE & PI & NIQE & PI & NIQE & PI \\ \hline
RealSR & 6.24 & 8.16 & 5.74 & 4.51 & 4.83 & 4.54 & 4.76 & 4.65 & 4.74 & 4.51 & \second{4.69} & \second{4.50} & 5.56 & 4.75 & \best{4.67} & \best{4.47} \\ 
DRealSR & 6.58 & 8.58 & 6.14 & 4.78 & 4.98 & 4.77 & 4.71 & 4.74 & \second{4.20} & \second{4.30} & 4.82 & 4.76 & 5.14 & 4.46 & \best{4.15} & \best{4.27} \\ 
DPED-iphone & 6.01 & 7.48 & 5.99 & 4.55 & 5.44 & 5.02 & \second{4.95} & 4.78 & 5.11 & \second{4.36} & 5.56 & 5.36 & 5.89 & 4.61 & \best{4.84} & \best{4.23} \\ 
RealSRSet & 7.98 & 7.35 & 5.49 & 4.79 & 5.65 & 4.92 & 5.30 & 4.68 & \best{5.18} & \best{4.31} &  \second{5.25}  & \second{4.59} & 6.03 & 4.60 & 5.29 & 4.62 \\ \hline
\end{tabular}
}
\end{table*}

\section{Implementation Details}

\subsection{Datasets}

\paragraph{Training Dataset.} 
Our training dataset generation process follows that of Real-ESRGAN \cite{wang2021real}, in which we obtain HQ images sourced from DIV2K \cite{DIV2K}, Flickr2K \cite{lim2017enhanced}, and OutdoorSceneTraining \cite{wang2018sftgan}. These images are cropped into non-overlapping patches of size $256\times256$ to serve as HQ images. Meanwhile, the corresponding LQ images are produced using the second-order degradation model proposed in \cite{wang2021real}.

\paragraph{Testing Datasets.} 
We evaluate the performance of our model on multiple benchmarks that include real-world LQ images such as RealSR \cite{wang2021towards}, DRealSR \cite{wei2020cdc}, DPED-iphone \cite{ignatov2017dslr}, and RealSRSet \cite{zhang2021designing}. Additionally, we create a synthetic dataset using the DIV2K validation set to validate the effectiveness of different model configurations.

\subsection{Training and inference details.} 
ITER is composed of three networks, namely $E_l$, $\phi_r$, and $\phi_e$, trained with cross-entropy losses in \cref{eq:token_pred,eq:loss_refine,eq:loss_eva}. In theory, the optimal strategy comprises training $E_l$ foremost, succeeded by $\phi_e$ and $\phi_r$ sequentially. Nevertheless, we discovered that training them concurrently works well in practice, thereby leading to a significant reduction in overall training time. The prominent Adam optimizer \cite{kingma2014adam} is employed to optimize all three networks, with specific parameters of $lr=0.0001$, $\beta_1=0.9$, and $\beta_2=0.99$. Each batch contains 16 HQ images of dimensions $256\times256$, paired with their corresponding LQ images. All networks are implemented by PyTorch \cite{pytorch} and trained for 400k iterations with 4 Tesla V100 GPUs. More details are in supplementary material.

\section{Experiments} 

\begin{figure}[!t]
    \centering
    \newcommand{\imgwidth}{0.32\linewidth}
    \includegraphics[width=\linewidth]{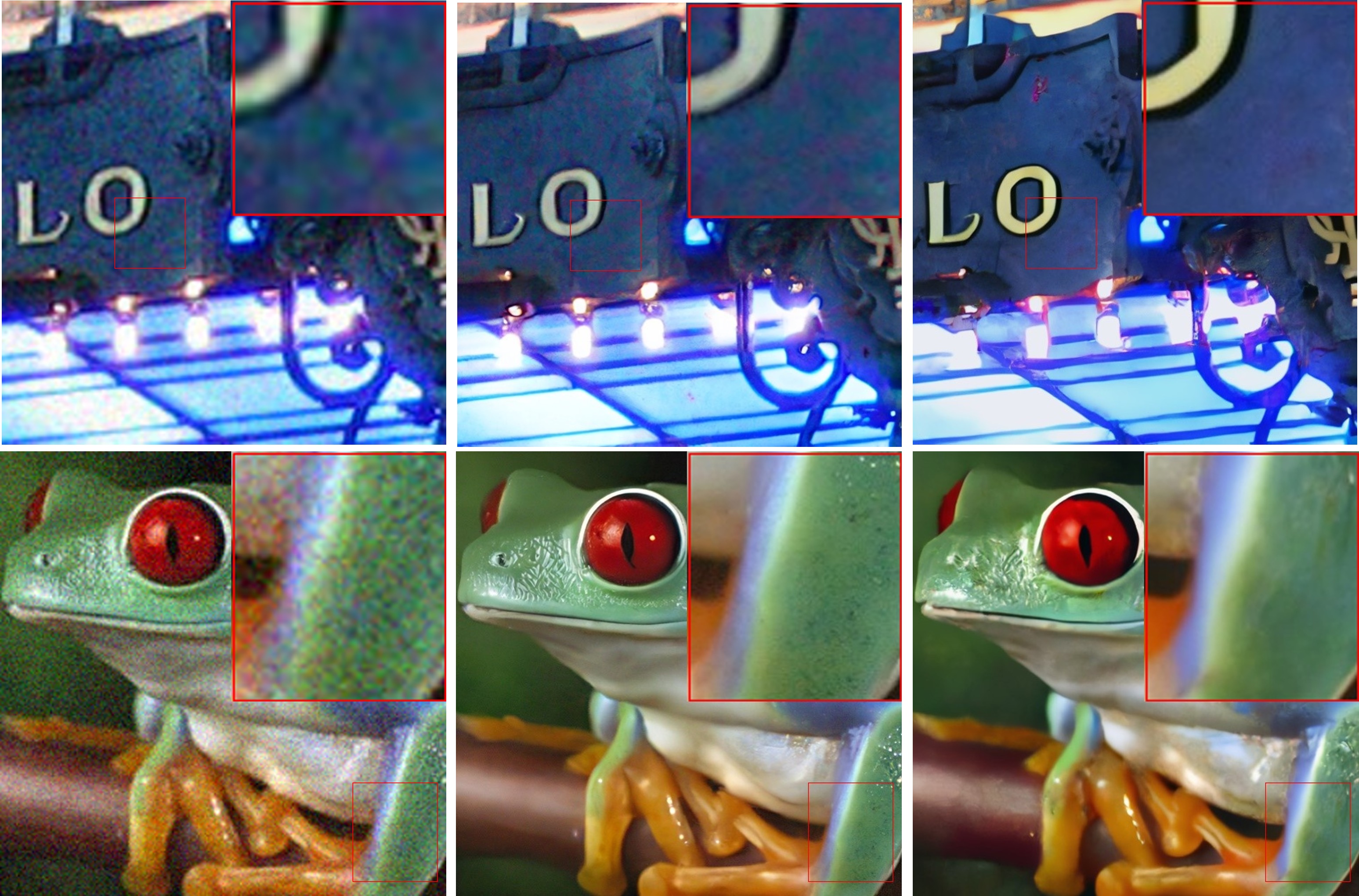}
    \makebox[\imgwidth]{\small (a) LQ input} 
    \makebox[\imgwidth]{\small (b) LDM-BSR} 
    \makebox[\imgwidth]{\small (c) \textbf{ITER (Ours)}}
    \caption{Problem of LDM-BSR without explicit distortion removal. (Zoom in for best view)}
    \label{fig:ldmbsr}
\end{figure}

\begin{figure}[!t]
    \centering
    \newcommand{\imgwidth}{0.32\linewidth}
    \includegraphics[width=\linewidth]{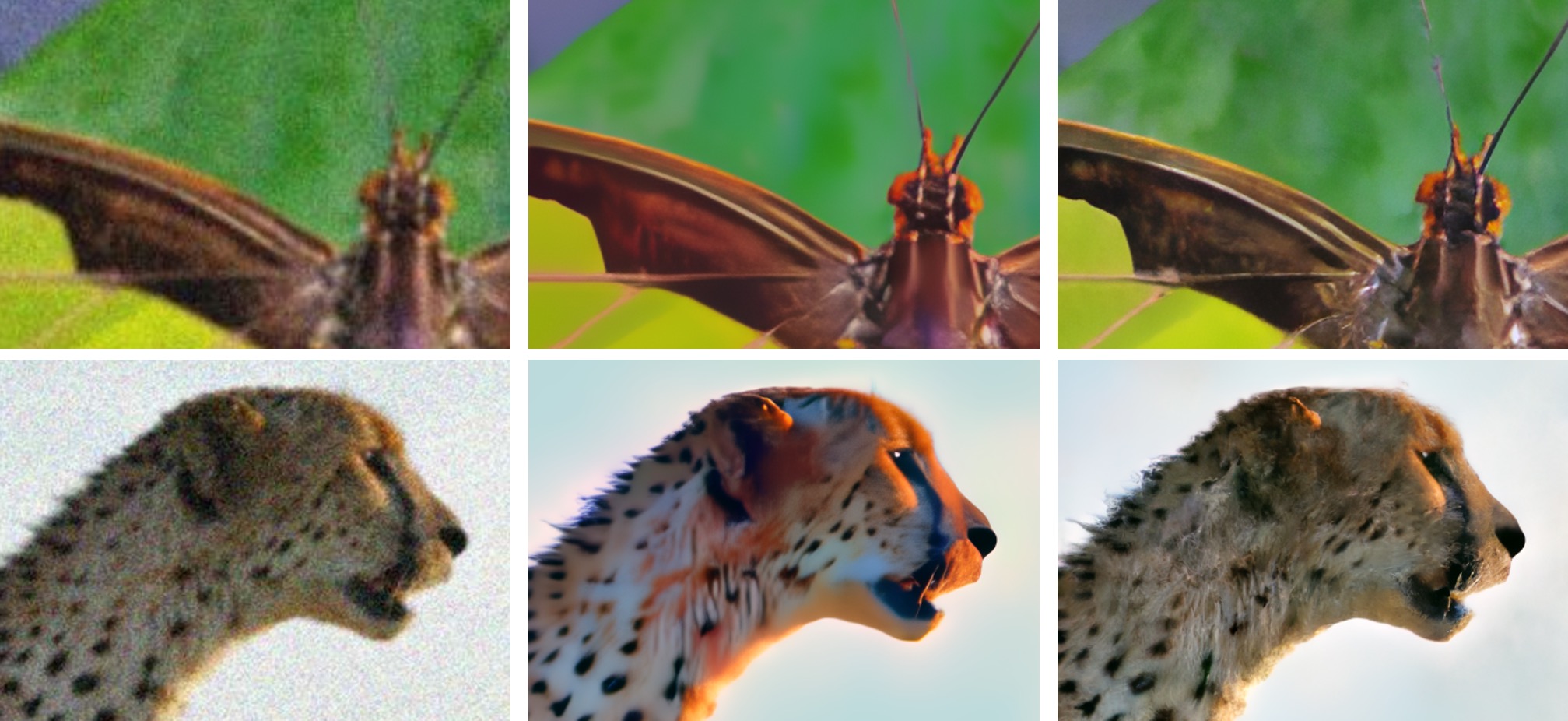}
    \\
    \makebox[\imgwidth]{(a) LQ inputs}
    \makebox[\imgwidth]{(b) w/o Refinement}
    \makebox[\imgwidth]{(c) w/ Refinement}
    \vspace{-1em}
    \caption{Comparison of results with and without iterative refinement. We can observe that the results only with distortion removal present overly smoothed textures and inconsistent color. After iterative refinement, the textures are enriched and the color is also corrected.}
    \label{fig:compare_woiter}
\end{figure}

\begin{figure*}[!t]
    \centering
    \begin{minipage}[c]{0.74\textwidth}
    \includegraphics[width=\linewidth]{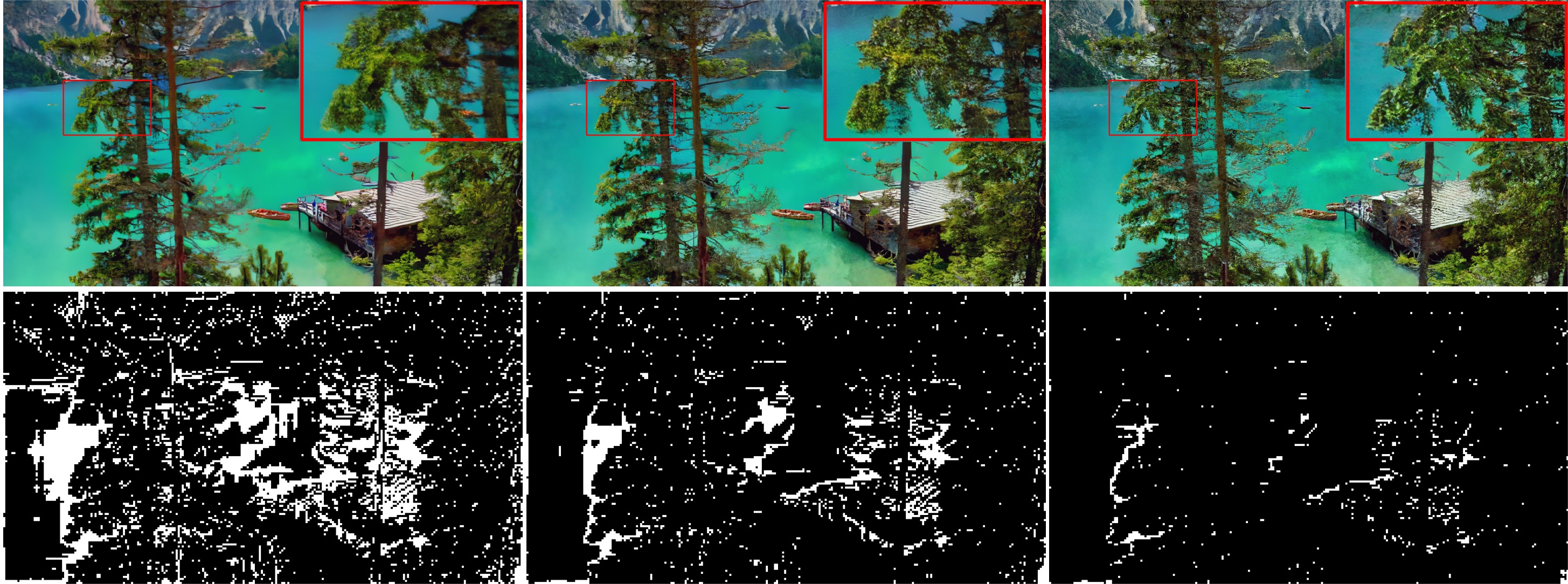} \\
    \makebox[0.32\linewidth]{\small $\alpha=0.4, T_s=3$}
    \makebox[0.32\linewidth]{\small $\alpha=0.5, T_s=4$}
    \makebox[0.32\linewidth]{\small $\alpha=0.6, T_s=6$}
    \caption{Visual examples of different threshold. Top: final results; bottom: masks at start time step. Bigger $\alpha$ leads to stronger texture effect because more refinement steps are conducted.}
    \label{fig:threshold_eg}
    \end{minipage}
    \begin{minipage}[c]{0.24\textwidth}
    \centering
    \includegraphics[width=1.01\linewidth]{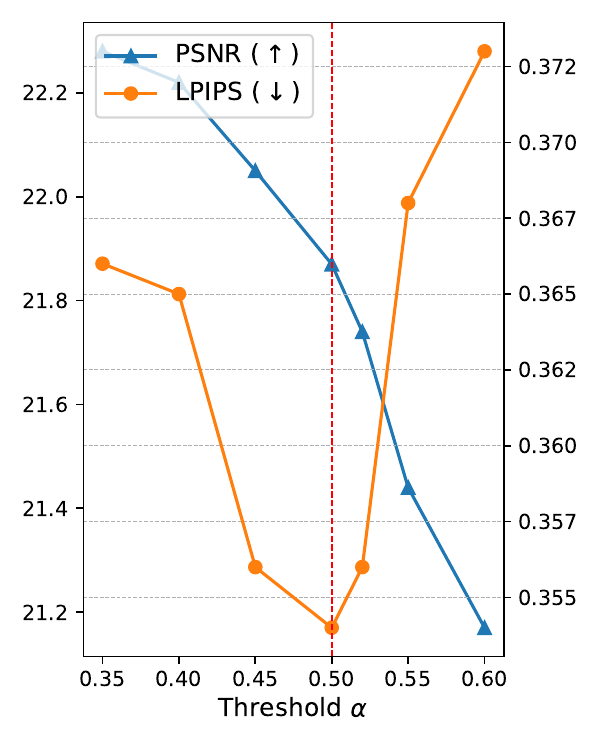}
    \caption{LPIPS/PSNR with different $\alpha$.}
    \label{fig:threshold}
    \end{minipage}
    \vspace{-1em}
\end{figure*}

\begin{figure}[!t]
    \centering
    \includegraphics[width=\linewidth]{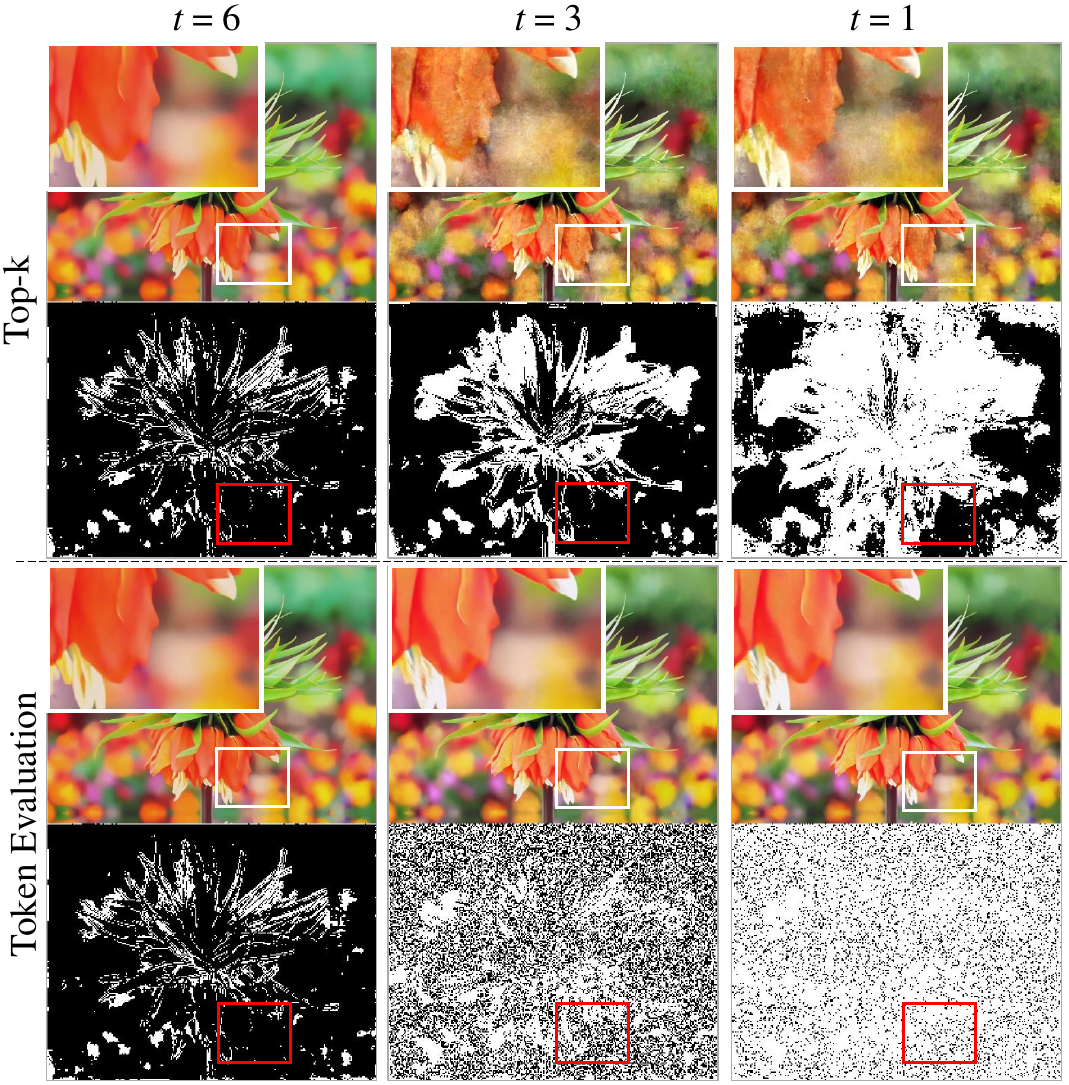}
    \caption{The top-k masking technique suffers from the local propagation problem, which is effectively avoided by the proposed token evaluation block.}
    \label{fig:topk_localtrap}
\end{figure}

\subsection{Comparison with other methods}

We perform a comprehensive comparison of ITER against several state-of-the-art GAN-based approaches, including BSRGAN \cite{zhang2021designing}, Real-ESRGAN \cite{wang2021towards}, SwinIR-GAN \cite{liang2021swinir}, FeMaSR \cite{chen2022femasr}, and MM-RealSR \cite{mou2022mmrealsr}. Specifically, BSRGAN, Real-ESRGAN, and MM-RealSR employ the RRDBNet backbone proposed by \cite{wang2018esrgan}, whereas SwinIR-GAN utilizes the Swin transformer architecture, and FeMaSR utilizes the VQGAN prior. Regarding diffusion-based models, we compare with the most popular work, LDM-BSR \cite{rombach2022latentdiffusion}, which operates in the latent feature space using the denoising diffusion models. The model is finetuned with the same dataset for fair comparison. SR3 \cite{saharia2022sr3} is not included in comparison due to the unavailability of public models.

We use two different no-reference metrics, namely NIQE \cite{2012niqe} and PI (perceptual index) \cite{blau2018pi}, to evaluate the performance of different approaches. NIQE is widely used in previous works involving RWSR, such as \cite{wang2021towards, zhang2021blind, mou2022mmrealsr}, while PI has been extensively used in recent low-level computer vision workshops, including the renowned NTIRE \cite{2019ntire, 2020ntire, 2021ntire} and AIM \cite{2019aim, 2020aim}. 

\paragraph{Comparison with GAN methods.} 
As demonstrated in \cref{tab:real_benchmarks}, our ITER yields the best performance in 3 out of 4 benchmarks as demonstrated, and the results in the last RealSRSet are also competitive. These results demonstrate the clear superiority of ITER over existing GAN-based methods. The visual examples depicted in \cref{fig:vis_comp} illustrate why ITER performs better. We can observe that the textures in the images generated by ITER look more natural and realistic. On the other hand, the results from other GAN-based approaches are either over-smoothed (first row in \cref{fig:vis_comp}) or over-sharpened (second row). GAN-based methods often encounter difficulties in generating realistic textures for different distortion levels. Moreover, they are generally harder to train and more likely to produce artifacts when not well-tuned. In conclusion, compared to GAN-based methods, our proposed ITER exhibits better performance and is more straightforward to train.

\paragraph{Comparison with LDM-BSR.}
As can be seen from \cref{tab:real_benchmarks}, it is evident that although LDM-BSR utilizes a diffusion-based model, its performance is worse than that of ITER. In \cref{fig:ldmbsr}, it is apparent why quantitative results of LDM-BSR are suboptimal for the RWSR task. Although LDM-BSR is capable of generating sharper edges for the blurry LQ inputs, it struggles with eliminating complex noise degradations in both examples. On the other hand, our proposed ITER does not face such challenges and can produce outputs with greater clarity while maintaining reasonably natural textures. This can be attributed to two main reasons. Firstly, LDM-BSR incorporates continuous diffusion models, while ITER relies on discrete representations. Prior studies \cite{zhou2022codeformer,chen2022femasr} have shown that a pre-trained discrete proxy space offers benefits for intricate distortions. Secondly, ITER explicitly filters out the distortions during the encoding of LQ images into token space before diffusion processing. As a result, ITER avoids generating additional textures similar to what can occur in LDM-BSR, as demonstrated in the second example.

\subsection{Ablation study and model analysis}

We performed a thorough analysis of various configurations of our model using a synthetic DIV2K validation test set. Firstly, we evaluated the effectiveness of refinement network in adding textures to the initial results $S_l$. Secondly, we assessed the necessity of the token evaluation block. Finally, we demonstrated how the token evaluation block can be exploited to manage the model preference toward removing distortions or generating textures. We utilized the PSNR metric to evaluate the quality of distortion removal and used the widely recognized perceptual metric LPIPS \cite{zhang2018perceptual} to measure the performance of texture generation. The incorporation of these two metrics allowed us to assess the extent to which the proposed ITER adjusts the visual effects of its outputs in accordance with the threshold value $\alpha$, as stated in \cref{eq:ms}.

\paragraph{Effectiveness of iterative refinement.} 
We first evaluate the effectiveness of the iterative refinement network for texture generation. As illustrated in \cref{fig:compare_woiter}, the results obtained without the iterative refinement stage exhibit an over-smoothed texture and inconsistency in color. This could be attributed to the inherent limitations of token classification when confronted with complex distortions present in diverse natural images. In contrast, the results with iterative refinement are more realistic. Noticeable enhancements in texture richness and color correction are observed. These observations provide compelling evidence that the iterative refinement network plays an crucial role in our framework. 

\paragraph{Necessity of token evaluation.} 
An alternative method to decide which tokens to retain or refine involves directly selecting the top-k tokens in $S_t$ with higher confidence, as implemented in MaskGIT \cite{chang2022maskgit}. However, our experimental findings indicate that the top-k mask selection is trapped with local propagation. This is due to the fact that under the greedy selection strategy, the refinement network $\phi_r$ tends to assign higher confidence to neighboring tokens of previous selections. As illustrated in \cref{fig:topk_localtrap}, the masks consistently expand around the previous step, resulting in some regions (indicated by black mask) being refined until the last step. This approach is unfavorable in the iterative texture generation process because it corrupts some good-looking regions with unnecessary refinement. Our hypothesis is that low-level vision tasks exhibit the locality property where neighboring features are naturally more correlated. Although the networks have large receptive fields with Swin transformer blocks, it still prefers to propagate information to neighbor features, resulting in higher confidence scores surrounding previous selections.

The use of the proposed token evaluation network $\phi_e$ allows the iterative refinement process to avoid the local propagation trap. As demonstrated in \cref{fig:topk_localtrap}, the masks are distributed more evenly, leading to more consistent results.

\paragraph{Balance restoration and generation.}
In \cref{fig:threshold_eg}, we have presented an example of the results with different threshold $\alpha$. It is evident from the results that a larger $\alpha$ will lead to the identification of fewer valid tokens, thereby necessitating more refinement steps, or in other words, a larger start time step $T_s$. Consequently, larger $\alpha$ create images with stronger textures. In \cref{fig:threshold}, we have provided quantitative results for the different $\alpha$ thresholds, where the effectiveness of each threshold can be seen in the score curves of LPIPS and PSNR. We have observed that smaller $\alpha$ produce enhanced PSNR scores, which is a clear indication of a better ability to eliminate distortion. As for texture generation performance, the optimal LPIPS score of $\alpha=0.5$ was achieved since both excessively strong and overly weak textures can negatively impact the perceptual quality. In practice, we can adjust $\alpha$ to obtain the desired results without having to modify the network, resulting in a more adaptable framework during inference than GAN-based techniques, which are unmodifiable once the training process is completed.

\section{Conclusion}

We presents a novel framework named ITER that utilizes iterative evaluation and refinement techniques for texture generation in real-world image super-resolution. Unlike GANs, which require painstake training, we incorporate discrete diffusion generative pipelines with token evaluation and refinement blocks for RWSR. This new approach simplifies training with just cross-entropy losses and allows for greater flexibility in balancing distortion removal and texture generation during inference. Furthermore, our ITER has demonstrated superior performance with $\leq8$ iterations, highlighting the vast potential of discrete diffusion models in RWSR.

%%%%%%%%% REFERENCES
% \bibliographystyle{ACM-Reference-Format}
\bibliography{egbib}

\begin{thebibliography}{68}
\providecommand{\natexlab}[1]{#1}

\bibitem[{Agustsson and Timofte(2017)}]{DIV2K}
Agustsson, E.; and Timofte, R. 2017.
\newblock {NTIRE} 2017 Challenge on Single Image Super-Resolution: Dataset and Study.
\newblock In \emph{CVPRW}.

\bibitem[{Anwar, Khan, and Barnes(2020)}]{anwar2020deepsurvey}
Anwar, S.; Khan, S.; and Barnes, N. 2020.
\newblock A deep journey into super-resolution: A survey.
\newblock \emph{ACM Computing Surveys (CSUR)}, 53(3): 1--34.

\bibitem[{Blau et~al.(2018)Blau, Mechrez, Timofte, Michaeli, and Zelnik-Manor}]{blau2018pi}
Blau, Y.; Mechrez, R.; Timofte, R.; Michaeli, T.; and Zelnik-Manor, L. 2018.
\newblock The 2018 PIRM challenge on perceptual image super-resolution.
\newblock In \emph{ECCVW}, 0--0.

\bibitem[{Bond-Taylor et~al.(2022)Bond-Taylor, Hessey, Sasaki, Breckon, and Willcocks}]{bond2021unleashing}
Bond-Taylor, S.; Hessey, P.; Sasaki, H.; Breckon, T.~P.; and Willcocks, C.~G. 2022.
\newblock Unleashing Transformers: Parallel Token Prediction with Discrete Absorbing Diffusion for Fast High-Resolution Image Generation from Vector-Quantized Codes.
\newblock In \emph{ECCV}.

\bibitem[{Brock, Donahue, and Simonyan(2019)}]{brock2018biggan}
Brock, A.; Donahue, J.; and Simonyan, K. 2019.
\newblock Large Scale {GAN} Training for High Fidelity Natural Image Synthesis.
\newblock In \emph{ICLR}.

\bibitem[{Cai et~al.(2019)}]{2019ntire}
Cai, J.; et~al. 2019.
\newblock NTIRE 2019 Challenge on Real Image Super-Resolution: Methods and Results.
\newblock \emph{CVPRW}.

\bibitem[{Chan et~al.(2021)Chan, Wang, Xu, Gu, and Loy}]{chan2021glean}
Chan, K.~C.; Wang, X.; Xu, X.; Gu, J.; and Loy, C.~C. 2021.
\newblock {GLEAN}: Generative latent bank for large-factor image super-resolution.
\newblock In \emph{CVPR}, 14245--14254.

\bibitem[{Chang et~al.(2022)Chang, Zhang, Jiang, Liu, and Freeman}]{chang2022maskgit}
Chang, H.; Zhang, H.; Jiang, L.; Liu, C.; and Freeman, W.~T. 2022.
\newblock {MaskGIT}: Masked Generative Image Transformer.
\newblock In \emph{CVPR}.

\bibitem[{Chen et~al.(2020)Chen, Gong, Wang, Li, and Wong}]{ChenSPARNet}
Chen, C.; Gong, D.; Wang, H.; Li, Z.; and Wong, K.-Y.~K. 2020.
\newblock Learning Spatial Attention for Face Super-Resolution.
\newblock In \emph{IEEE TIP}.

\bibitem[{Chen and Mo(2022)}]{pyiqa}
Chen, C.; and Mo, J. 2022.
\newblock {IQA-PyTorch}: PyTorch Toolbox for Image Quality Assessment.
\newblock [Online]. Available: \url{https://github.com/chaofengc/IQA-PyTorch}.

\bibitem[{Chen et~al.(2022)Chen, Shi, Qin, Li, Han, Yang, and Guo}]{chen2022femasr}
Chen, C.; Shi, X.; Qin, Y.; Li, X.; Han, X.; Yang, T.; and Guo, S. 2022.
\newblock Real-World Blind Super-Resolution via Feature Matching with Implicit High-Resolution Priors.
\newblock In \emph{ACM MM}.

\bibitem[{Chen et~al.(2021)Chen, Wang, Guo, Xu, Deng, Liu, Ma, Xu, Xu, and Gao}]{chen2020IPT}
Chen, H.; Wang, Y.; Guo, T.; Xu, C.; Deng, Y.; Liu, Z.; Ma, S.; Xu, C.; Xu, C.; and Gao, W. 2021.
\newblock Pre-Trained Image Processing Transformer.
\newblock In \emph{CVPR}.

\bibitem[{Chen et~al.(2023)Chen, Wang, Zhou, Qiao, and Dong}]{chen2023hat}
Chen, X.; Wang, X.; Zhou, J.; Qiao, Y.; and Dong, C. 2023.
\newblock Activating More Pixels in Image Super-Resolution Transformer.
\newblock In \emph{Proceedings of the IEEE/CVF Conference on Computer Vision and Pattern Recognition (CVPR)}, 22367--22377.

\bibitem[{Cui et~al.(2019)Cui, Jia, Lin, Song, and Belongie}]{cui2019class}
Cui, Y.; Jia, M.; Lin, T.-Y.; Song, Y.; and Belongie, S. 2019.
\newblock Class-balanced loss based on effective number of samples.
\newblock In \emph{Proceedings of the IEEE/CVF conference on computer vision and pattern recognition}, 9268--9277.

\bibitem[{Esser, Rombach, and Ommer(2021)}]{esser2021taming}
Esser, P.; Rombach, R.; and Ommer, B. 2021.
\newblock Taming transformers for high-resolution image synthesis.
\newblock In \emph{CVPR}, 12873--12883.

\bibitem[{Fritsche, Gu, and Timofte(2019)}]{fritsche2019frequency}
Fritsche, M.; Gu, S.; and Timofte, R. 2019.
\newblock Frequency separation for real-world super-resolution.
\newblock In \emph{ICCVW}, 3599--3608.

\bibitem[{Gao et~al.(2023)Gao, Liu, Zeng, Xu, Li, Luo, Liu, Zhen, and Zhang}]{gao2023implicit}
Gao, S.; Liu, X.; Zeng, B.; Xu, S.; Li, Y.; Luo, X.; Liu, J.; Zhen, X.; and Zhang, B. 2023.
\newblock Implicit diffusion models for continuous super-resolution.
\newblock In \emph{Proceedings of the IEEE/CVF Conference on Computer Vision and Pattern Recognition}, 10021--10030.

\bibitem[{Gu et~al.(2021)}]{2021ntire}
Gu, J.; et~al. 2021.
\newblock NTIRE 2021 Challenge on Perceptual Image Quality Assessment.
\newblock \emph{CVPRW}.

\bibitem[{Gu et~al.(2022)Gu, Chen, Bao, Wen, Zhang, Chen, Yuan, and Guo}]{gu2021vqdm}
Gu, S.; Chen, D.; Bao, J.; Wen, F.; Zhang, B.; Chen, D.; Yuan, L.; and Guo, B. 2022.
\newblock Vector Quantized Diffusion Model for Text-to-Image Synthesis.
\newblock \emph{CVPR}.

\bibitem[{He et~al.(2016)He, Zhang, Ren, and Sun}]{he2016deep}
He, K.; Zhang, X.; Ren, S.; and Sun, J. 2016.
\newblock Deep residual learning for image recognition.
\newblock In \emph{CVPR}, 770--778.

\bibitem[{Ignatov et~al.(2017)Ignatov, Kobyshev, Timofte, Vanhoey, and Van~Gool}]{ignatov2017dslr}
Ignatov, A.; Kobyshev, N.; Timofte, R.; Vanhoey, K.; and Van~Gool, L. 2017.
\newblock {DSLR}-quality photos on mobile devices with deep convolutional networks.
\newblock In \emph{ICCV}, 3277--3285.

\bibitem[{Ignatov et~al.(2019)}]{2019aim}
Ignatov, A.; et~al. 2019.
\newblock AIM 2019 Challenge on RAW to RGB Mapping: Methods and Results.
\newblock \emph{ICCVW}.

\bibitem[{Ignatov et~al.(2020)}]{2020aim}
Ignatov, A.; et~al. 2020.
\newblock AIM 2020 Challenge on Learned Image Signal Processing Pipeline.
\newblock \emph{ECCVW}, 152–170.

\bibitem[{Ji et~al.(2020)Ji, Cao, Tai, Wang, Li, and Huang}]{ji2020realsr}
Ji, X.; Cao, Y.; Tai, Y.; Wang, C.; Li, J.; and Huang, F. 2020.
\newblock Real-world super-resolution via kernel estimation and noise injection.
\newblock In \emph{CVPRW}, 466--467.

\bibitem[{Karras et~al.(2020)Karras, Laine, Aittala, Hellsten, Lehtinen, and Aila}]{karras2020analyzing}
Karras, T.; Laine, S.; Aittala, M.; Hellsten, J.; Lehtinen, J.; and Aila, T. 2020.
\newblock Analyzing and improving the image quality of stylegan.
\newblock In \emph{CVPR}, 8110--8119.

\bibitem[{Kim, Lee, and Lee(2016{\natexlab{a}})}]{kim2016accurate}
Kim, J.; Lee, J.~K.; and Lee, K.~M. 2016{\natexlab{a}}.
\newblock Accurate image super-resolution using very deep convolutional networks.
\newblock In \emph{CVPR}, 1646--1654.

\bibitem[{Kim, Lee, and Lee(2016{\natexlab{b}})}]{kim2016deeply}
Kim, J.; Lee, J.~K.; and Lee, K.~M. 2016{\natexlab{b}}.
\newblock Deeply-recursive convolutional network for image super-resolution.
\newblock In \emph{CVPR}, 1637--1645.

\bibitem[{Kingma and Ba(2014)}]{kingma2014adam}
Kingma, D.~P.; and Ba, J. 2014.
\newblock Adam: A method for stochastic optimization.
\newblock \emph{arXiv preprint arXiv:1412.6980}.

\bibitem[{Ledig et~al.(2017)Ledig, Theis, Husz{\'a}r, Caballero, Cunningham, Acosta, Aitken, Tejani, Totz, Wang et~al.}]{ledig2017srgan}
Ledig, C.; Theis, L.; Husz{\'a}r, F.; Caballero, J.; Cunningham, A.; Acosta, A.; Aitken, A.; Tejani, A.; Totz, J.; Wang, Z.; et~al. 2017.
\newblock Photo-realistic single image super-resolution using a generative adversarial network.
\newblock In \emph{CVPR}, 4681--4690.

\bibitem[{Lezama et~al.(2022)Lezama, Chang, Jiang, and Essa}]{lezama2022tokencritic}
Lezama, J.; Chang, H.; Jiang, L.; and Essa, I. 2022.
\newblock Improved masked image generation with token-critic.
\newblock \emph{ECCV}.

\bibitem[{Li et~al.(2022)Li, Chen, Lin, Zuo, and Zhang}]{Li_2022_ReDegNet}
Li, X.; Chen, C.; Lin, X.; Zuo, W.; and Zhang, L. 2022.
\newblock From Face to Natural Image: Learning Real Degradation for Blind Image Super-Resolution.
\newblock In \emph{ECCV}.

\bibitem[{Liang et~al.(2021)Liang, Cao, Sun, Zhang, Van~Gool, and Timofte}]{liang2021swinir}
Liang, J.; Cao, J.; Sun, G.; Zhang, K.; Van~Gool, L.; and Timofte, R. 2021.
\newblock SwinIR: Image Restoration Using Swin Transformer.
\newblock In \emph{ICCVW}.

\bibitem[{Liang, Zeng, and Zhang(2022)}]{jie2022DASR}
Liang, J.; Zeng, H.; and Zhang, L. 2022.
\newblock Efficient and Degradation-Adaptive Network for Real-World Image Super-Resolution.
\newblock In \emph{ECCV}.

\bibitem[{Lim et~al.(2017)Lim, Son, Kim, Nah, and Mu~Lee}]{lim2017enhanced}
Lim, B.; Son, S.; Kim, H.; Nah, S.; and Mu~Lee, K. 2017.
\newblock Enhanced deep residual networks for single image super-resolution.
\newblock In \emph{CVPRW}, 136--144.

\bibitem[{Liu et~al.(2022)Liu, Liu, Gu, Qiao, and Dong}]{liu2022blindsrsurvey}
Liu, A.; Liu, Y.; Gu, J.; Qiao, Y.; and Dong, C. 2022.
\newblock Blind image super-resolution: A survey and beyond.
\newblock \emph{IEEE TPAMI}.

\bibitem[{Liu et~al.(2023)Liu, Wei, Wu, Zuo, and Zhang}]{liu2023survey}
Liu, M.; Wei, Y.; Wu, X.; Zuo, W.; and Zhang, L. 2023.
\newblock Survey on leveraging pre-trained generative adversarial networks for image editing and restoration.
\newblock \emph{Science China Information Sciences}, 66(5): 1--28.

\bibitem[{Liu et~al.(2021)Liu, Lin, Cao, Hu, Wei, Zhang, Lin, and Guo}]{liu2021swin}
Liu, Z.; Lin, Y.; Cao, Y.; Hu, H.; Wei, Y.; Zhang, Z.; Lin, S.; and Guo, B. 2021.
\newblock Swin transformer: Hierarchical vision transformer using shifted windows.
\newblock \emph{ICCV}.

\bibitem[{Maeda(2020)}]{maeda2020unpaired}
Maeda, S. 2020.
\newblock Unpaired image super-resolution using pseudo-supervision.
\newblock In \emph{CVPR}, 291--300.

\bibitem[{Mei, Fan, and Zhou(2021)}]{Mei_2021_CVPR_NLSA}
Mei, Y.; Fan, Y.; and Zhou, Y. 2021.
\newblock Image Super-Resolution With Non-Local Sparse Attention.
\newblock In \emph{CVPR}, 3517--3526.

\bibitem[{Mittal, Soundararajan, and Bovik(2012)}]{2012niqe}
Mittal, A.; Soundararajan, R.; and Bovik, A.~C. 2012.
\newblock Making a “completely blind” image quality analyzer.
\newblock \emph{IEEE Sign. Process. Letters}, 20(3): 209--212.

\bibitem[{Mou et~al.(2022)Mou, Wu, Wang, Dong, Zhang, and Shan}]{mou2022mmrealsr}
Mou, C.; Wu, Y.; Wang, X.; Dong, C.; Zhang, J.; and Shan, Y. 2022.
\newblock {MM-RealSR}: Metric Learning based Interactive Modulation for Real-World Super-Resolution.
\newblock \emph{ECCV}.

\bibitem[{Niu et~al.(2020)Niu, Wen, Ren, Zhang, Yang, Wang, Zhang, Cao, and Shen}]{niu2020HAN}
Niu, B.; Wen, W.; Ren, W.; Zhang, X.; Yang, L.; Wang, S.; Zhang, K.; Cao, X.; and Shen, H. 2020.
\newblock Single image super-resolution via a holistic attention network.
\newblock In \emph{ECCV}, 191--207. Springer.

\bibitem[{Pan et~al.(2020)Pan, Zhan, Dai, Lin, Loy, and Luo}]{pan2020exploiting}
Pan, X.; Zhan, X.; Dai, B.; Lin, D.; Loy, C.~C.; and Luo, P. 2020.
\newblock Exploiting deep generative prior for versatile image restoration and manipulation.
\newblock In \emph{ECCV}, 262--277. Springer.

\bibitem[{Paszke et~al.(2019)Paszke, Gross, Massa, Lerer, Bradbury, Chanan, Killeen, Lin, Gimelshein, Antiga, Desmaison, Kopf, Yang, DeVito, Raison, Tejani, Chilamkurthy, Steiner, Fang, Bai, and Chintala}]{pytorch}
Paszke, A.; Gross, S.; Massa, F.; Lerer, A.; Bradbury, J.; Chanan, G.; Killeen, T.; Lin, Z.; Gimelshein, N.; Antiga, L.; Desmaison, A.; Kopf, A.; Yang, E.; DeVito, Z.; Raison, M.; Tejani, A.; Chilamkurthy, S.; Steiner, B.; Fang, L.; Bai, J.; and Chintala, S. 2019.
\newblock PyTorch: An Imperative Style, High-Performance Deep Learning Library.
\newblock In \emph{NeurIPS}, volume~32, 8026--8037.

\bibitem[{Rombach et~al.(2022)Rombach, Blattmann, Lorenz, Esser, and Ommer}]{rombach2022latentdiffusion}
Rombach, R.; Blattmann, A.; Lorenz, D.; Esser, P.; and Ommer, B. 2022.
\newblock High-resolution image synthesis with latent diffusion models.
\newblock In \emph{CVPR}, 10684--10695.

\bibitem[{Saharia et~al.(2022)Saharia, Ho, Chan, Salimans, Fleet, and Norouzi}]{saharia2022sr3}
Saharia, C.; Ho, J.; Chan, W.; Salimans, T.; Fleet, D.~J.; and Norouzi, M. 2022.
\newblock Image super-resolution via iterative refinement.
\newblock \emph{IEEE TPAMI}.

\bibitem[{Wan et~al.(2020)Wan, Zhang, Chen, Zhang, Chen, Liao, and Wen}]{wan2020bringing}
Wan, Z.; Zhang, B.; Chen, D.; Zhang, P.; Chen, D.; Liao, J.; and Wen, F. 2020.
\newblock Bringing old photos back to life.
\newblock In \emph{CVPR}, 2747--2757.

\bibitem[{Wang et~al.(2023)Wang, Yue, Zhou, Chan, and Loy}]{wang2023stablesr}
Wang, J.; Yue, Z.; Zhou, S.; Chan, K.~C.; and Loy, C.~C. 2023.
\newblock Exploiting Diffusion Prior for Real-World Image Super-Resolution.
\newblock In \emph{arXiv preprint arXiv:2305.07015}.

\bibitem[{Wang et~al.(2021{\natexlab{a}})Wang, Wang, Dong, Xu, Yang, An, and Guo}]{wang2021unsupervised}
Wang, L.; Wang, Y.; Dong, X.; Xu, Q.; Yang, J.; An, W.; and Guo, Y. 2021{\natexlab{a}}.
\newblock Unsupervised Degradation Representation Learning for Blind Super-Resolution.
\newblock In \emph{CVPR}, 10581--10590.

\bibitem[{Wang et~al.(2021{\natexlab{b}})Wang, Li, Zhang, and Shan}]{wang2021towards}
Wang, X.; Li, Y.; Zhang, H.; and Shan, Y. 2021{\natexlab{b}}.
\newblock Towards Real-World Blind Face Restoration with Generative Facial Prior.
\newblock In \emph{CVPR}, 9168--9178.

\bibitem[{Wang et~al.(2021{\natexlab{c}})Wang, Xie, Dong, and Shan}]{wang2021real}
Wang, X.; Xie, L.; Dong, C.; and Shan, Y. 2021{\natexlab{c}}.
\newblock Real-ESRGAN: Training Real-World Blind Super-Resolution with Pure Synthetic Data.
\newblock \emph{ICCVW}.

\bibitem[{Wang et~al.(2018{\natexlab{a}})Wang, Yu, Dong, and Loy}]{wang2018sftgan}
Wang, X.; Yu, K.; Dong, C.; and Loy, C.~C. 2018{\natexlab{a}}.
\newblock Recovering realistic texture in image super-resolution by deep spatial feature transform.
\newblock In \emph{CVPR}.

\bibitem[{Wang et~al.(2018{\natexlab{b}})Wang, Yu, Wu, Gu, Liu, Dong, Qiao, and Change~Loy}]{wang2018esrgan}
Wang, X.; Yu, K.; Wu, S.; Gu, J.; Liu, Y.; Dong, C.; Qiao, Y.; and Change~Loy, C. 2018{\natexlab{b}}.
\newblock Esrgan: Enhanced super-resolution generative adversarial networks.
\newblock In \emph{ECCVW}, 0--0.

\bibitem[{Wei et~al.(2020)Wei, Xie, Lu, Zhan, Ye, Zuo, and Lin}]{wei2020cdc}
Wei, P.; Xie, Z.; Lu, H.; Zhan, Z.; Ye, Q.; Zuo, W.; and Lin, L. 2020.
\newblock Component divide-and-conquer for real-world image super-resolution.
\newblock In \emph{ECCV}, 101--117. Springer.

\bibitem[{Wei et~al.(2021)Wei, Gu, Li, Timofte, Jin, and Song}]{wei2021unsupervised}
Wei, Y.; Gu, S.; Li, Y.; Timofte, R.; Jin, L.; and Song, H. 2021.
\newblock Unsupervised real-world image super resolution via domain-distance aware training.
\newblock In \emph{CVPR}, 13385--13394.

\bibitem[{Yang et~al.(2021)Yang, Ren, Xie, and Zhang}]{yang2021gan}
Yang, T.; Ren, P.; Xie, X.; and Zhang, L. 2021.
\newblock GAN Prior Embedded Network for Blind Face Restoration in the Wild.
\newblock In \emph{CVPR}, 672--681.

\bibitem[{Zamir et~al.(2022)Zamir, Arora, Khan, Hayat, Khan, and Yang}]{Zamir2021Restormer}
Zamir, S.~W.; Arora, A.; Khan, S.; Hayat, M.; Khan, F.~S.; and Yang, M.-H. 2022.
\newblock Restormer: Efficient Transformer for High-Resolution Image Restoration.
\newblock In \emph{CVPR}.

\bibitem[{Zhang et~al.(2021{\natexlab{a}})Zhang, Lu, Zhan, and Yu}]{zhang2021blind}
Zhang, J.; Lu, S.; Zhan, F.; and Yu, Y. 2021{\natexlab{a}}.
\newblock Blind Image Super-Resolution via Contrastive Representation Learning.
\newblock \emph{arXiv preprint arXiv:2107.00708}.

\bibitem[{Zhang et~al.(2021{\natexlab{b}})Zhang, Liang, Van~Gool, and Timofte}]{zhang2021designing}
Zhang, K.; Liang, J.; Van~Gool, L.; and Timofte, R. 2021{\natexlab{b}}.
\newblock Designing a practical degradation model for deep blind image super-resolution.
\newblock \emph{ICCV}.

\bibitem[{Zhang et~al.(2020)}]{2020ntire}
Zhang, K.; et~al. 2020.
\newblock NTIRE 2020 Challenge on Perceptual Extreme Super-Resolution: Methods and Results.
\newblock \emph{CVPRW}.

\bibitem[{Zhang et~al.(2018{\natexlab{a}})Zhang, Isola, Efros, Shechtman, and Wang}]{zhang2018perceptual}
Zhang, R.; Isola, P.; Efros, A.~A.; Shechtman, E.; and Wang, O. 2018{\natexlab{a}}.
\newblock The Unreasonable Effectiveness of Deep Features as a Perceptual Metric.
\newblock In \emph{CVPR}.

\bibitem[{Zhang et~al.(2019{\natexlab{a}})Zhang, Liu, Dong, and Qiao}]{zhang2019ranksrgan}
Zhang, W.; Liu, Y.; Dong, C.; and Qiao, Y. 2019{\natexlab{a}}.
\newblock Ranksrgan: Generative adversarial networks with ranker for image super-resolution.
\newblock In \emph{CVPR}, 3096--3105.

\bibitem[{Zhang et~al.(2022)Zhang, Zeng, Guo, and Zhang}]{zhang2022ELAN}
Zhang, X.; Zeng, H.; Guo, S.; and Zhang, L. 2022.
\newblock Efficient Long-Range Attention Network for Image Super-resolution.
\newblock In \emph{ECCV}.

\bibitem[{Zhang et~al.(2018{\natexlab{b}})Zhang, Li, Li, Wang, Zhong, and Fu}]{zhang2018image}
Zhang, Y.; Li, K.; Li, K.; Wang, L.; Zhong, B.; and Fu, Y. 2018{\natexlab{b}}.
\newblock Image super-resolution using very deep residual channel attention networks.
\newblock In \emph{ECCV}, 286--301.

\bibitem[{Zhang et~al.(2019{\natexlab{b}})Zhang, Li, Li, Zhong, and Fu}]{zhang2019rnan}
Zhang, Y.; Li, K.; Li, K.; Zhong, B.; and Fu, Y. 2019{\natexlab{b}}.
\newblock Residual Non-local Attention Networks for Image Restoration.
\newblock In \emph{ICLR}.

\bibitem[{Zhang et~al.(2018{\natexlab{c}})Zhang, Tian, Kong, Zhong, and Fu}]{zhang2018residual}
Zhang, Y.; Tian, Y.; Kong, Y.; Zhong, B.; and Fu, Y. 2018{\natexlab{c}}.
\newblock Residual dense network for image super-resolution.
\newblock In \emph{CVPR}, 2472--2481.

\bibitem[{Zhou et~al.(2022)Zhou, Chan, Li, and Loy}]{zhou2022codeformer}
Zhou, S.; Chan, K.~C.; Li, C.; and Loy, C.~C. 2022.
\newblock Towards Robust Blind Face Restoration with Codebook Lookup TransFormer.
\newblock In \emph{NeurIPS}.

\bibitem[{Zhou et~al.(2020)Zhou, Zhang, Zuo, and Loy}]{zhou2020cross}
Zhou, S.; Zhang, J.; Zuo, W.; and Loy, C.~C. 2020.
\newblock Cross-Scale Internal Graph Neural Network for Image Super-Resolution.
\newblock In \emph{NeurIPS}.

\end{thebibliography}

\appendix

\section{Network and Training Details}

\subsection{Network Architectures}

\begin{figure*}[h]
    \centering
    \includegraphics[width=\linewidth]{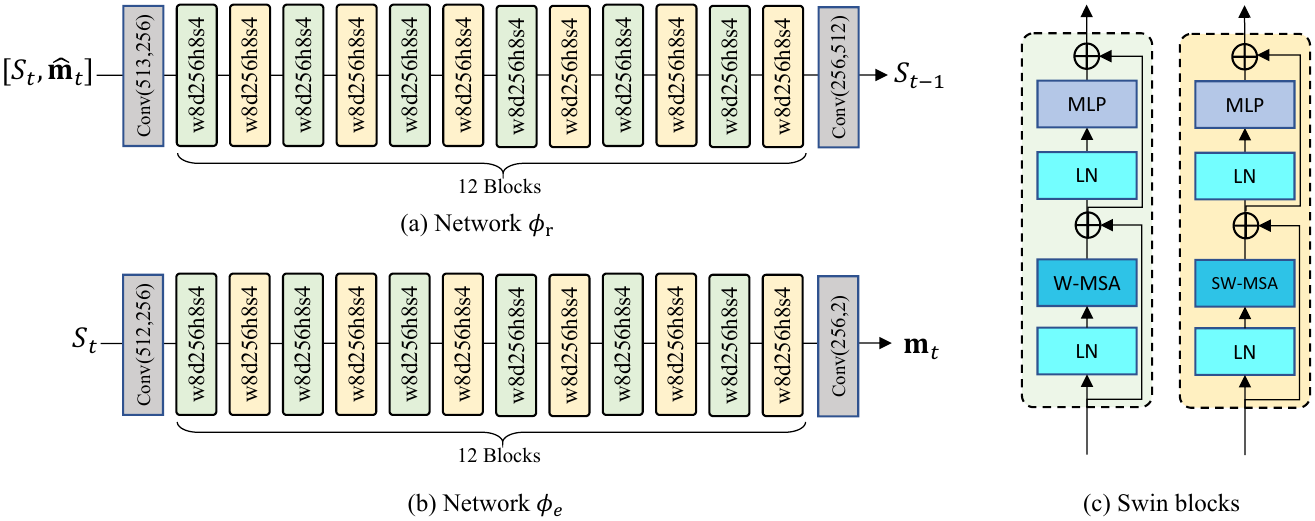}
    \caption{Detailed network architectures of $\phi_e$ and $\phi_r$. ``w8d256h8s4'' refers to: window size $8\times8$, feature dimension $256$, number of heads $8$, MLP scale ratio $4$. ``Conv(M, N)'' refers to convolution layer with $1\times1$ kernel, $M$ in channels and $N$ out channels.}
    \label{fig:network_archs}
\end{figure*}

As shown in \cref{fig:network_archs}, we use 12 Swin transformer blocks with window attention (W-MSA) and shifted window attention (SW-MSA) alternatively for token evaluation block $\phi_e$ and token refinement block $\phi_r$ respectively. The inputs $S_t$ are one-hot embeddings of image token indexes, and $\hat{\mathbf{m}}_t$ is the binary evaluation mask of size $1\times m\times n$, where $m=H/f, n = W/f$ and $H\times W$ is the size of HQ image. As for the distortion removal network $E_l$, we use a similar architecture as \cite{chen2022femasr}, except that we use the same 12 Swin blocks instead of RSTB blocks of SwinIR \cite{liang2021swinir}. 

\subsection{Training of Swin-VQGAN}

Similar as original VQGAN \cite{rombach2022latentdiffusion}, we use the same training losses as below:
\begin{gather*}
    \loss_{pix} = \|I_{rec} - I_h\|_1, \\ \loss_{per} =  \|\Psi(I_{rec}) - \Psi(I_h)\|_2^2, \\ \loss_{ssim} = 1 - \text{SSIM}(I_{rec}, I_h) \\
    \loss_{vq} = \|\text{sg}(Z_h) - Z_c \|_2^2 + \beta \|Z_h - \text{sg}(Z_c)\|_2^2 
\end{gather*}
where $I_{rec}$ is the reconstructed image, $\Psi$ is the LPIPS based perception function, SSIM is the differentiable SSIM function\footnote{Implemented with \texttt{IQA-PyTorch} \cite{pyiqa}}, ``sg'' is the stop gradient operation and $\beta=0.25$ as \cite{esser2021taming}. The gradient commitment operation is applied to copy gradient from decoder $D_H$ to encoder $E_H$ during training because vector quantization operation is non-differentiable. We use the hinge version GAN loss same as \cite{esser2021taming}.

The network is trained on 4 Tesla V100 GPUs with a batch size of 32. We empirically found that a smaller batch size decreases the reconstruction performance. The model is trained for 400k iterations and takes about 3 days.

\subsection{Running Time}

\begin{table*}[h]
    \centering
    \caption{Comparison of inference time with different methods.}
    \renewcommand{\arraystretch}{1.5}
    \begin{tabular}{c|cccc}
        \hline
         Model &  RRDBNet & SwinIR & LDM-BSR & ITER (ours, 8 iterations) \\ \hline
         Inference Time (s) & 0.06 & 0.21 & 4.2 & 1.7 \\ \hline 
    \end{tabular}
    \label{tab:supp_time}
\end{table*}

\Cref{tab:supp_time} shows the inference time comparison with different methods. The shape of input is $128\times128$ and upsampled by $\times4$ to get outputs of shape $512\times512$. All models are tested on a single Tesla V100 GPU and the time is averaged over 10 runs. 

It is expected that models with RRDB backbone based on pure convolution layers run faster than others. Although the running time of ITER is about 8 times of SwinIR, it is still much faster than LDM-BSR which requires 100 iterations. To further improve the efficiency of ITER, we can replace the slow Swin blocks in $\phi_e$ and $\phi_r$ with U-Net like LDM-BSR. It may decrease quantitative performance, but is likely to get similar qualitative results.

\subsection{More implementation details}

\paragraph{Metric Calculation.} For consistency in quantitative results, we calculate all metrics, \ie, NIQE, PSNR, and LPIPS with the open-source toolbox \texttt{IQA-PyTorch} \cite{pyiqa}.

\paragraph{Class Balanced Loss for $\phi_e$.} When training the network $\phi_e$ with \cref{eq:loss_eva}, we found that the labels in $\hat{\mathbf{m}}_l$ are quite imbalanced. This is because the distortion removal $E_l$ is not able to exactly restore the correct ground truth tokens $S_h$ for input $I_l$, which results in much more zeros than ones in $\hat{\mathbf{m}}_l$.
\begin{align}
    \loss_e = - \hat{\mt} \log \bigl( \phi_e(S_t) \bigr) - \mathbf{\hat{m}}_l \log \bigl( \phi_e(S_l) \bigr), \label{eq:loss_eva} 
\end{align}
This makes the learning of $\phi_e$ quite difficult with naive cross-entropy loss. We found that the simple class-balanced cross-entropy loss \cite{cui2019class} helps a lot, which can be formulated as below:
\begin{equation}
    \loss_e = - \hat{\mt} \log \bigl( \phi_e(S_t) \bigr) - \frac{1 - \beta}{1 - \beta^{n_y}}\mathbf{\hat{m}}_l \log \bigl( \phi_e(S_l) \bigr), \label{eq:loss_eva_balance} 
\end{equation}
where $n_y$ is the number of tokens for $y=0$ or $y=1$ in each batch, and $\beta=0.9999$ as suggested in \cite{cui2019class}. The class balanced loss re-weight the losses of ones and zeros according to their numbers, and works well to train $\phi_e$.    

We also tried to apply such loss to $\phi_r$, but it does not bring improvement. We suppose that this is because there are only $512$ token classes and each class has enough number of samples. Therefore, the class imbalance is not much of a problem.

\begin{figure*}[t]
    \centering
    \begin{subfigure}[t]{0.7\linewidth}
        \includegraphics[width=.99\linewidth]{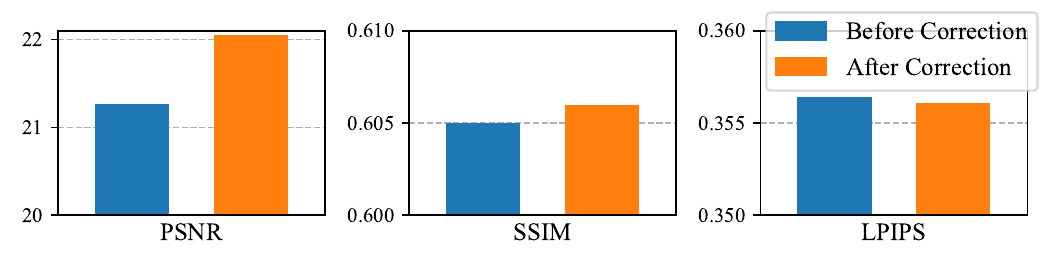}
        \caption{Influence of color correction to different metrics.} \label{fig:ttcc_quant}
    \end{subfigure}
    \\
    \begin{subfigure}[t]{0.99\linewidth}
    \includegraphics[width=.99\linewidth]{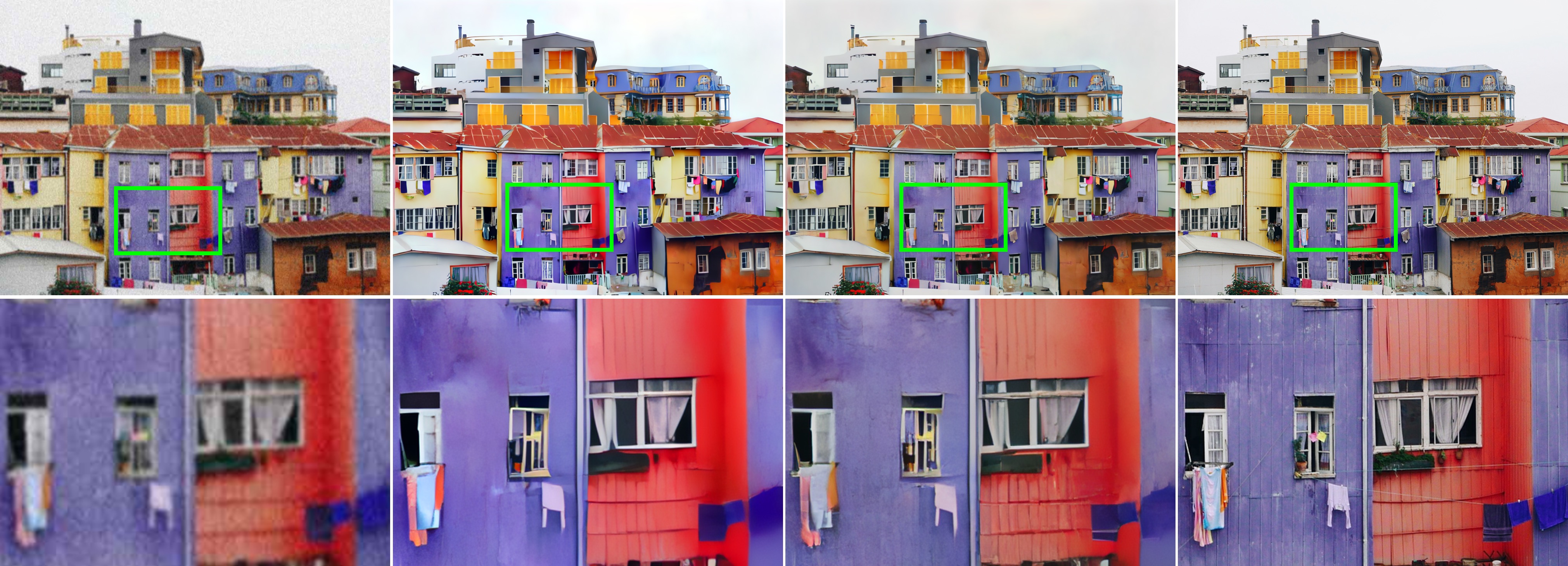}
    \newcommand{\imgwidth}{0.24\linewidth}
    \makebox[\imgwidth]{\small LQ input}
    \makebox[\imgwidth]{\small Result Before Correction}
    \makebox[\imgwidth]{\small Result After Correction}
    \makebox[\imgwidth]{\small Ground Truth HQ}
    \caption{Visual example for color correction.} \label{fig:ttcc_vis}
    \end{subfigure}
    \caption{Illustration for test-time color correction.}
    \label{fig:ttcc}
\end{figure*}

\paragraph{Test-Time Color Correction.} Although token space restoration is more robust, we found there exists slight color shift because ITER has no pixel space constraint like previous works \cite{chen2022femasr,zhou2022codeformer}. To solve this problem, we propose a simple test-time color correction to align the RGB distribution between LQ inputs and SR results as below:
\begin{equation}
    \hat{I}_{sr} = \frac{I_{sr} - \mu(I_{sr})}{\sigma(I_{sr})} \cdot \sigma(I_{lr}) + \mu(I_{lr}), \label{eq:ttcc}
\end{equation}
where $\mu(\cdot)$ and $\sigma(\cdot)$ are mean and standard deviation. This is based on the fact that RWSR usually does not contain color changes, and the global color distribution of LQ inputs often stays unchanged. \Cref{fig:ttcc} demonstrates the results of color correction. From \cref{fig:ttcc_quant}, we can observe that the PSNR improves a lot after correction while SSIM and LPIPS remain almost unchanged. This is expected because SSIM and LPIPS are more sensitive to texture quality. \Cref{fig:ttcc_vis} shows an example from the synthetic DIV2K validation set, it can be observed that the colors of results after correction is more close to ground truth.

\section{More Results and Analysis}

\subsection{Comparison Before and After Refinement}

\Cref{fig:compare_woiter} shows more examples proving the effectiveness and necessity of iterative token refinement. We can observe that simple distortion removal based on code prediction has two main problems: color changes and over-smooth. Note that the results are already calibrated with the \cref{eq:ttcc}. This indicates that the color problem is intrinsic to token prediction. With the proposed token refinement, we can largely resolve such problem. On the other hand, simple distortion removal generates over-smoothed results. With the proposed token refinement, our ITER is able to generate plausible and realistic textures.  

\begin{figure*}[h]
    \centering
    \newcommand{\imgwidth}{0.33\linewidth}
    \includegraphics[width=\imgwidth]{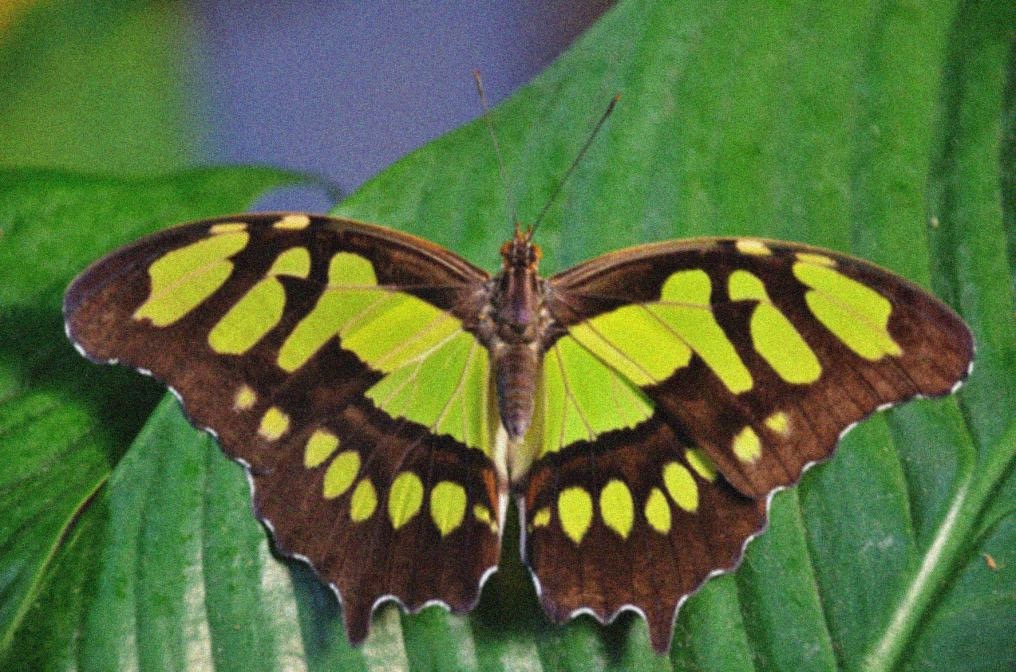}
    \includegraphics[width=\imgwidth]{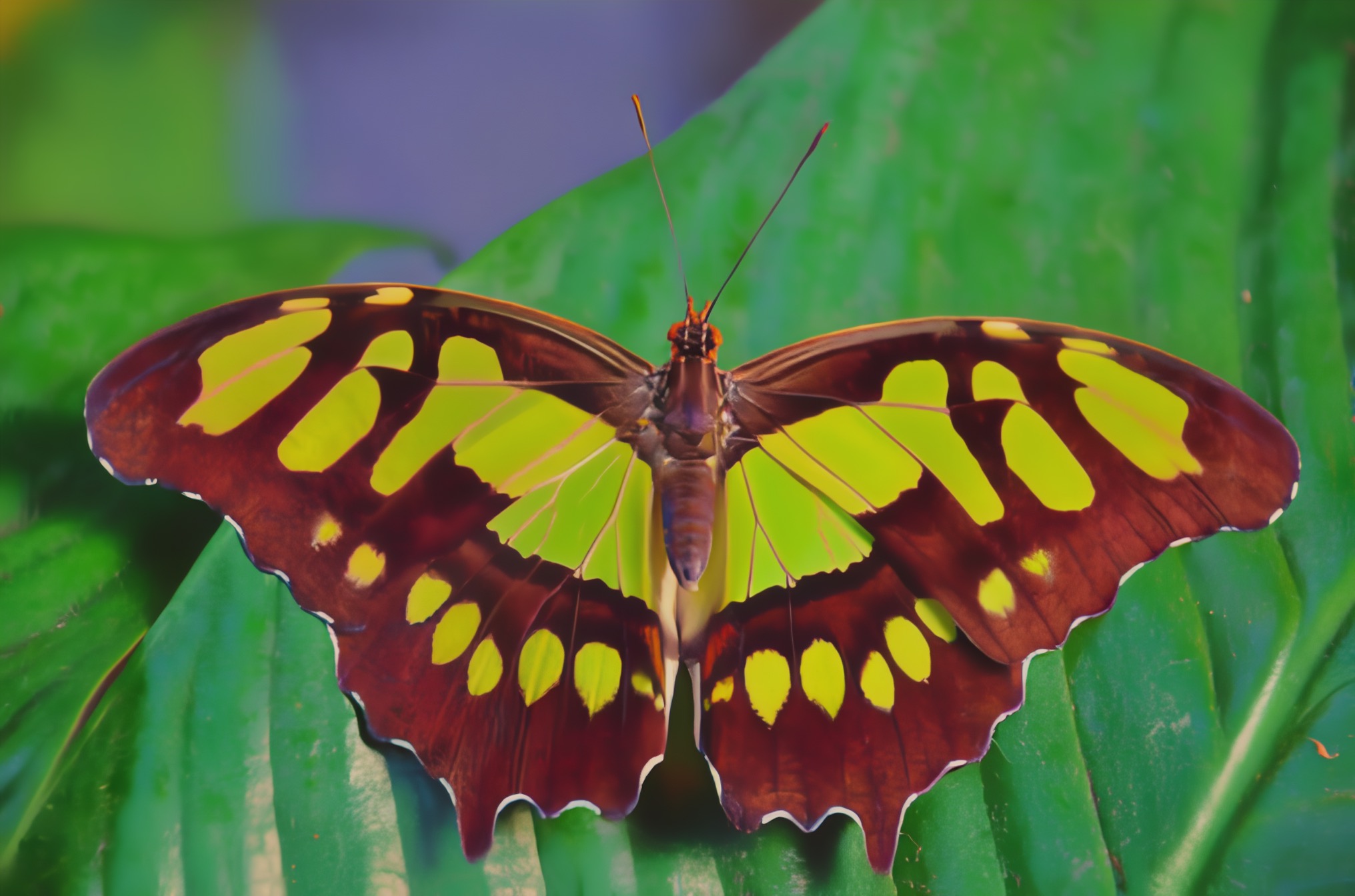}
    \includegraphics[width=\imgwidth]{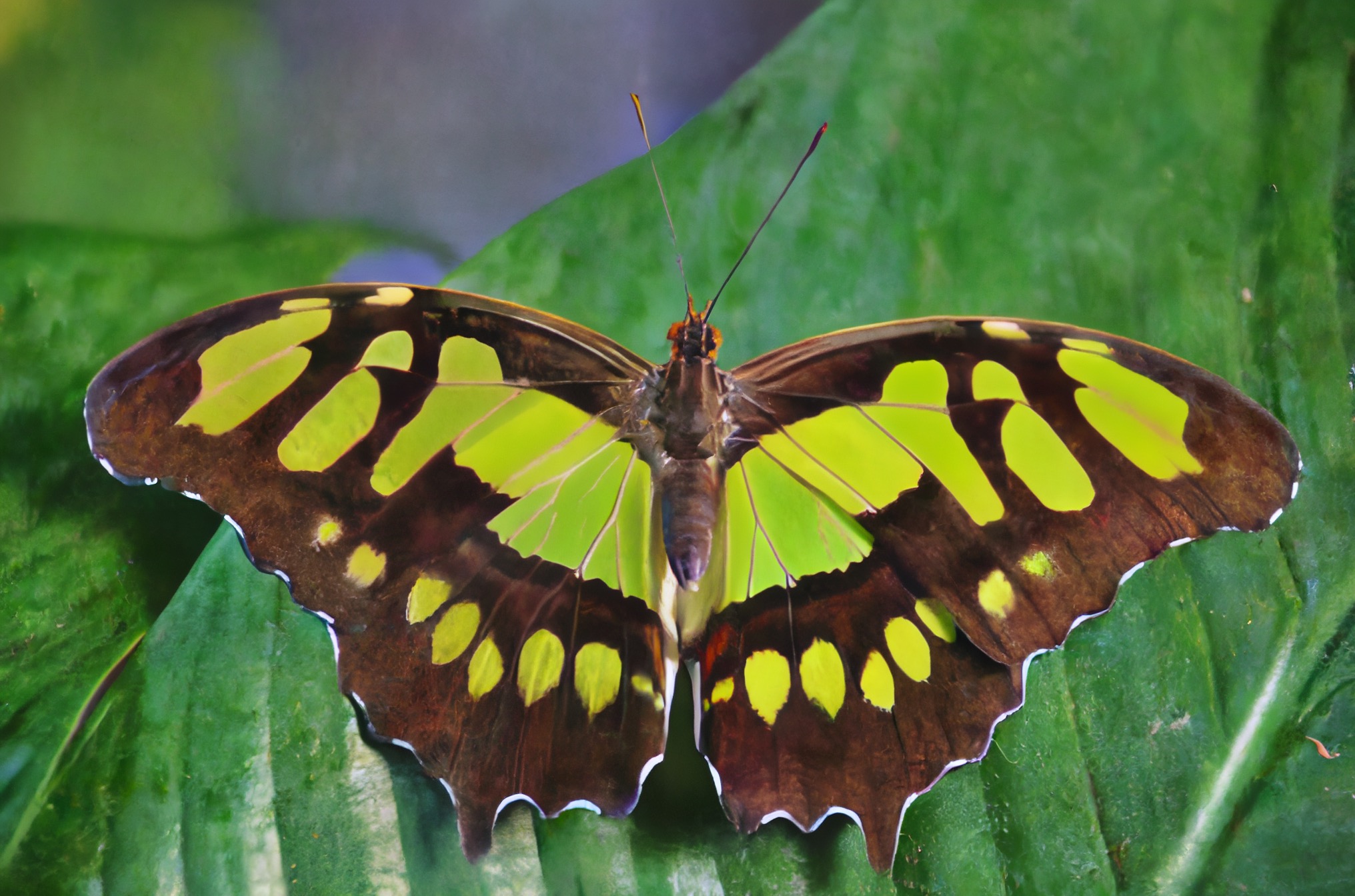}
    \includegraphics[width=\imgwidth]{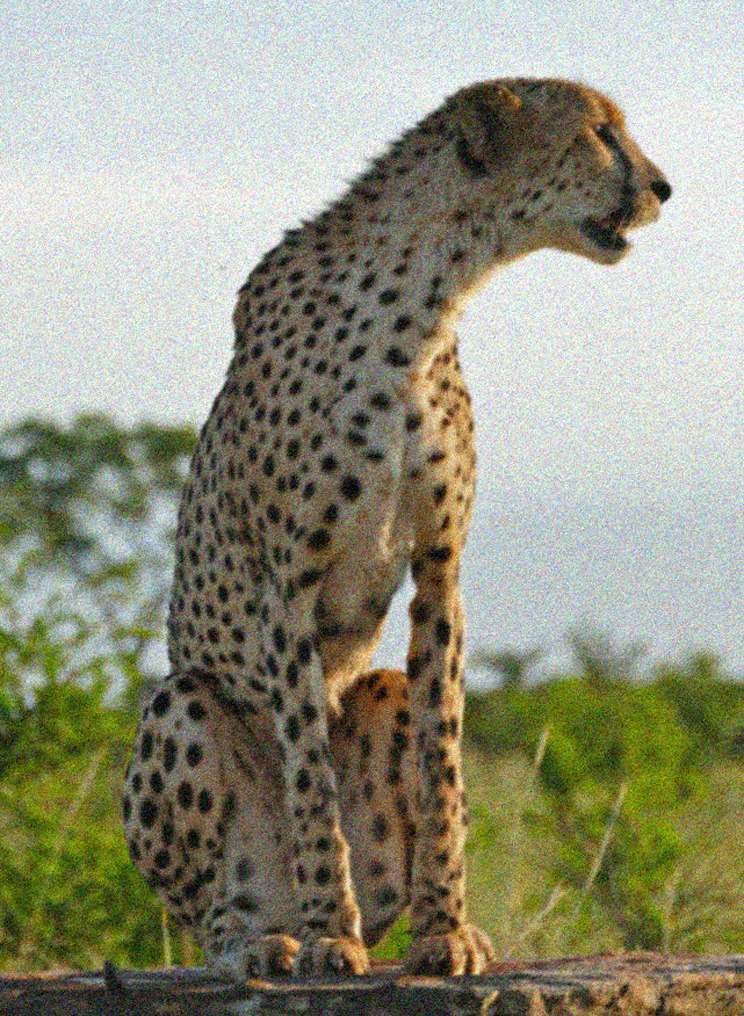}
    \includegraphics[width=\imgwidth]{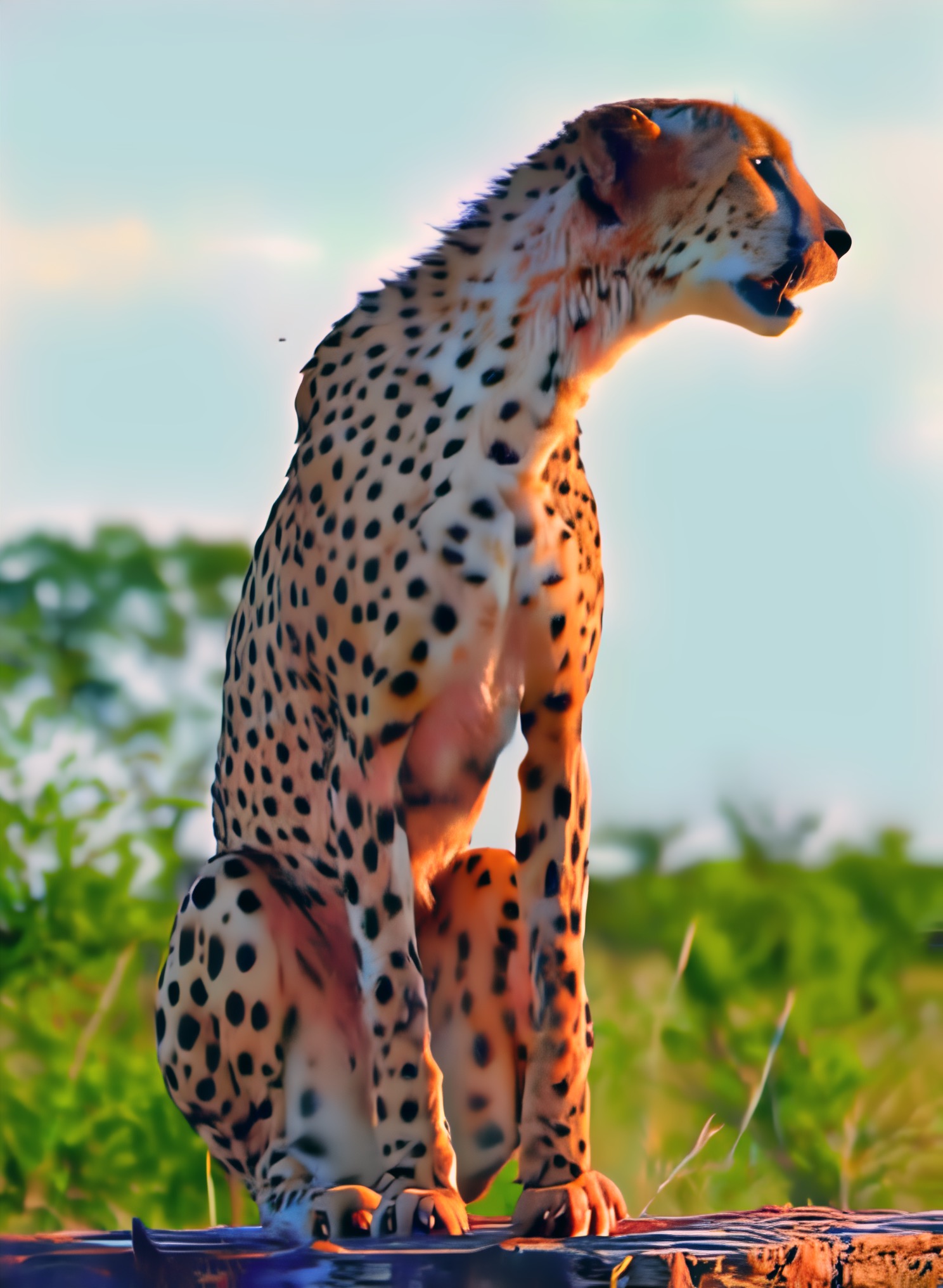}
    \includegraphics[width=\imgwidth]{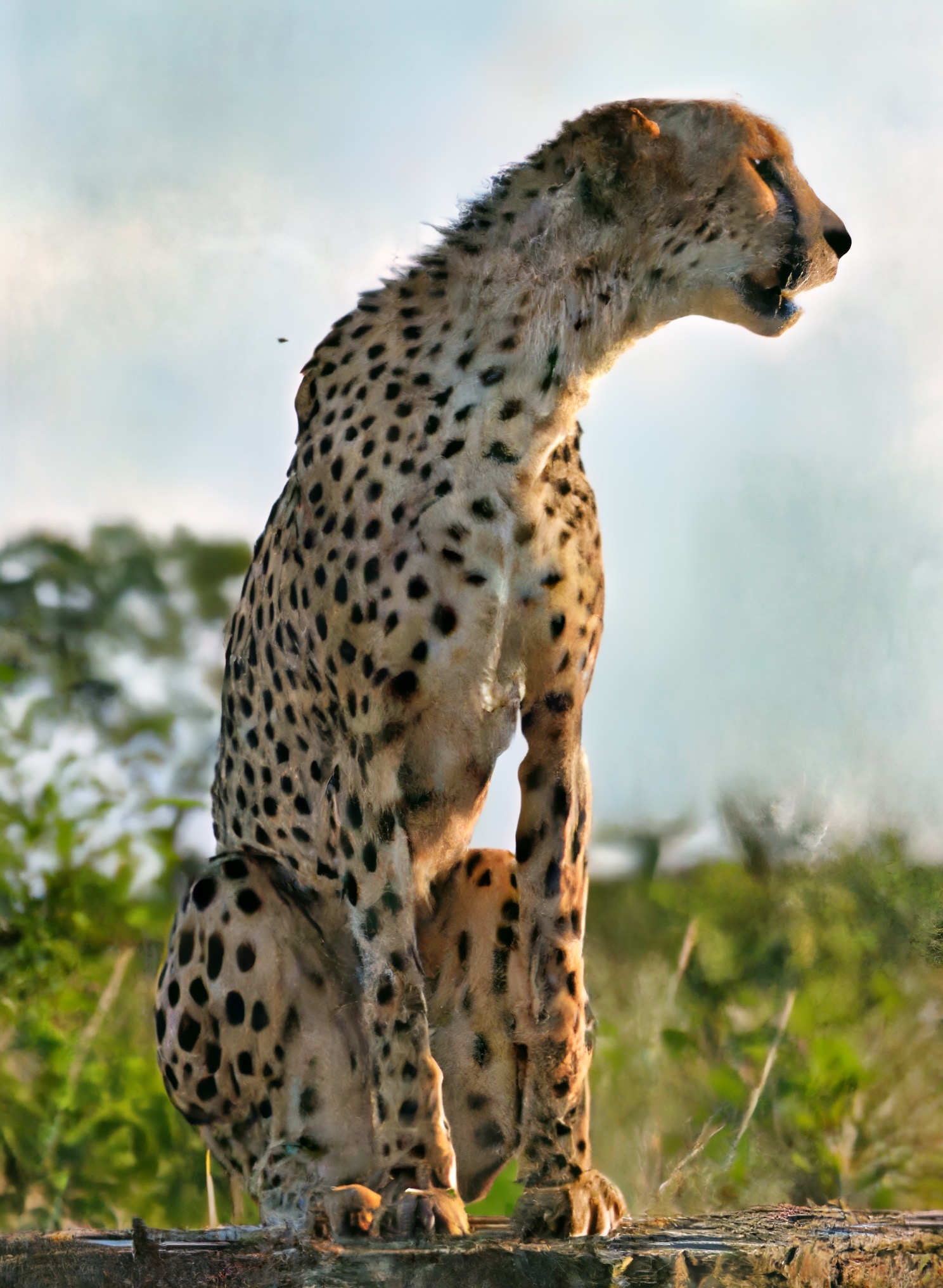}
    \makebox[\imgwidth]{(a) LQ inputs}
    \makebox[\imgwidth]{(b) Before Refinement}
    \makebox[\imgwidth]{(c) After Refinement}
    \caption{Comparison of results before and after iterative refinement. We can observe that the distortion removal based on code prediction generates results with severe color problem and over-smoothed details. After iterative refinement, the color is corrected and the textures are enriched. (Zoom in for best view)}
    \label{fig:compare_woiter}
\end{figure*}

\subsection{Additional Results with Different Threshold $\alpha$}

\begin{figure*}[h]
    \centering
    \newcommand{\imgwidth}{0.33\linewidth}
    \includegraphics[width=\imgwidth]{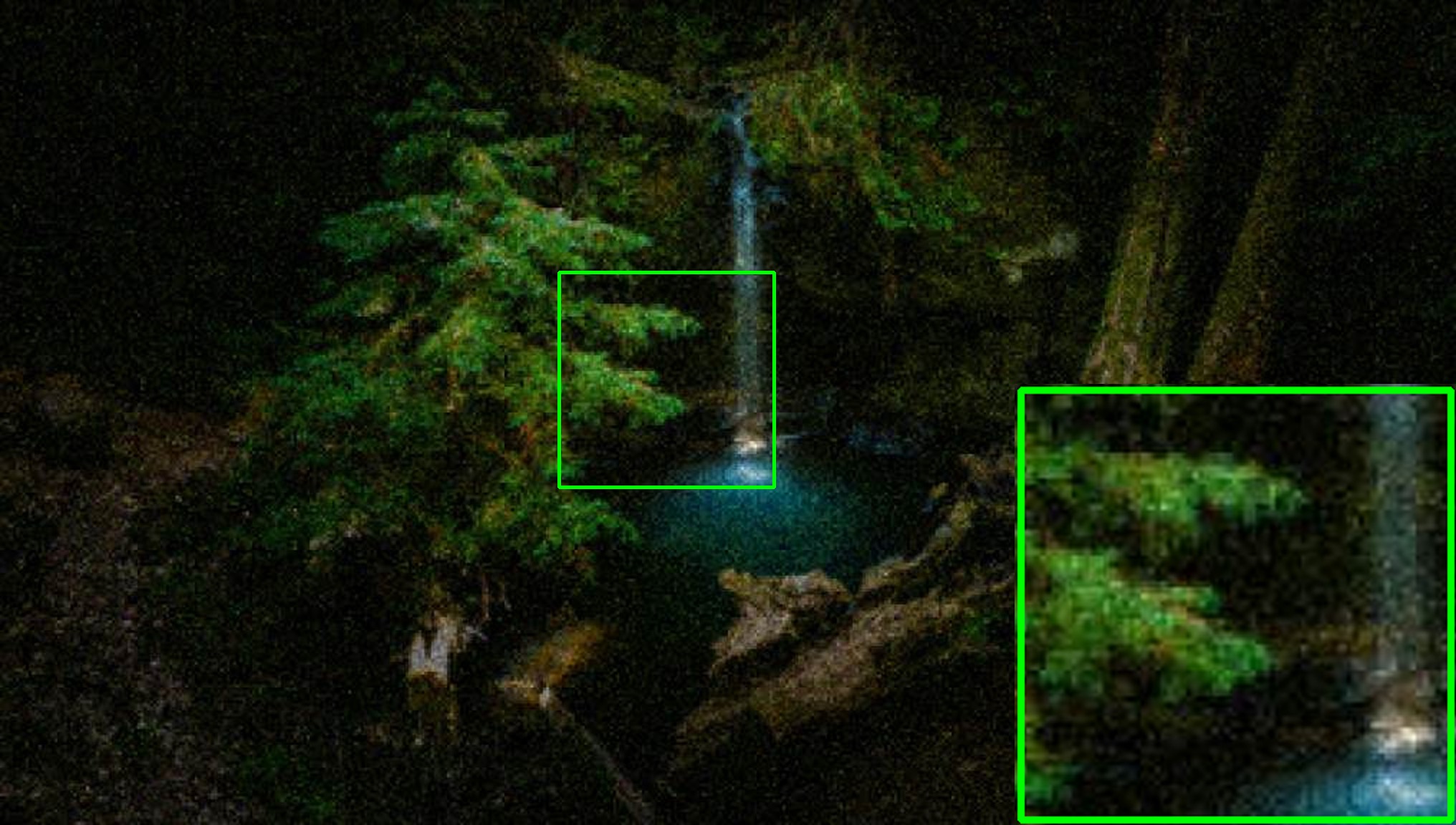}
    \includegraphics[width=\imgwidth]{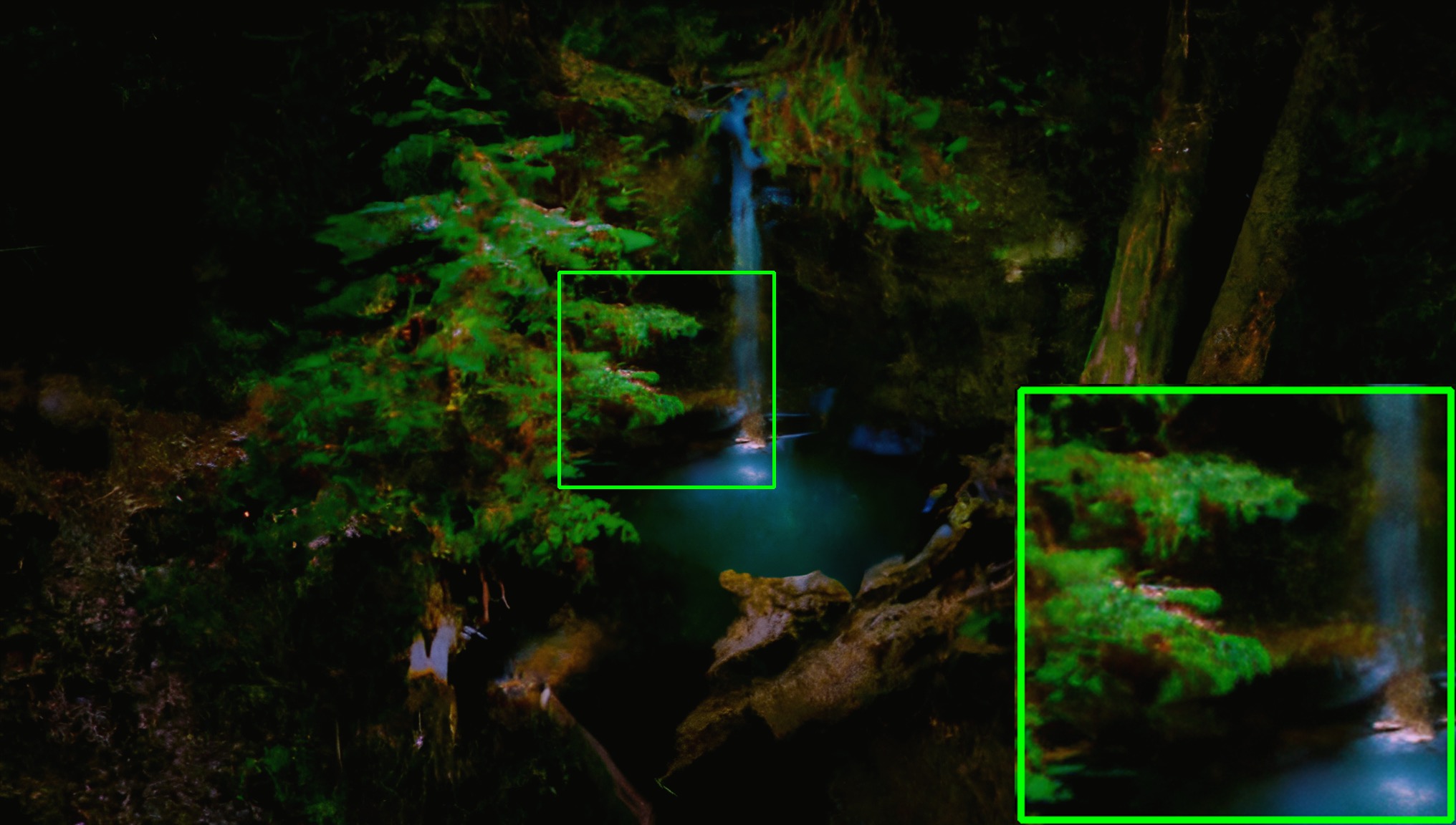}
    \includegraphics[width=\imgwidth]{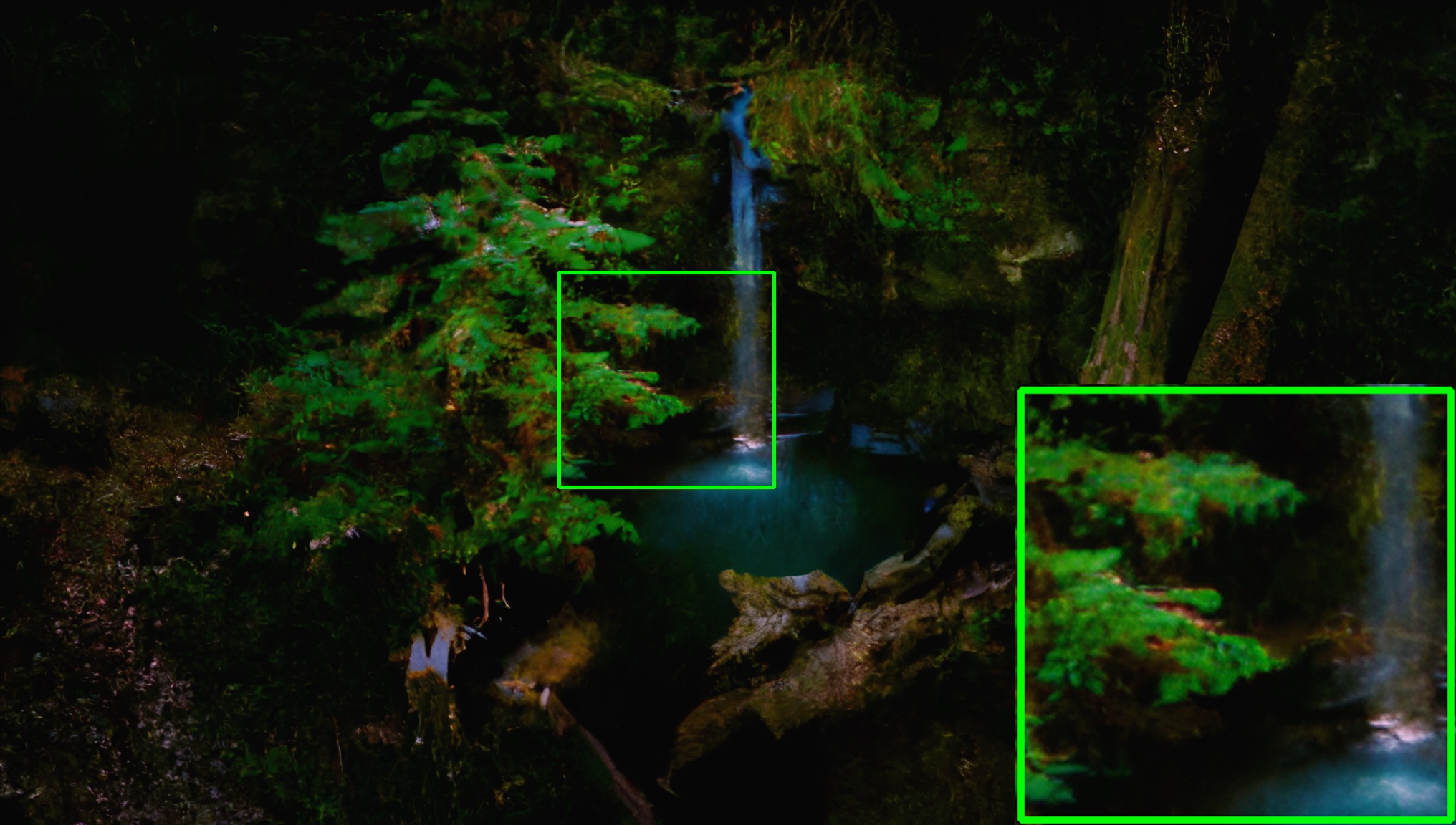}
    \\
    \makebox[\imgwidth]{(a) LQ input}
    \makebox[\imgwidth]{(b) $\alpha=0.35$}
    \makebox[\imgwidth]{(c) $\alpha=0.40$} 
    \\
    \includegraphics[width=\imgwidth]{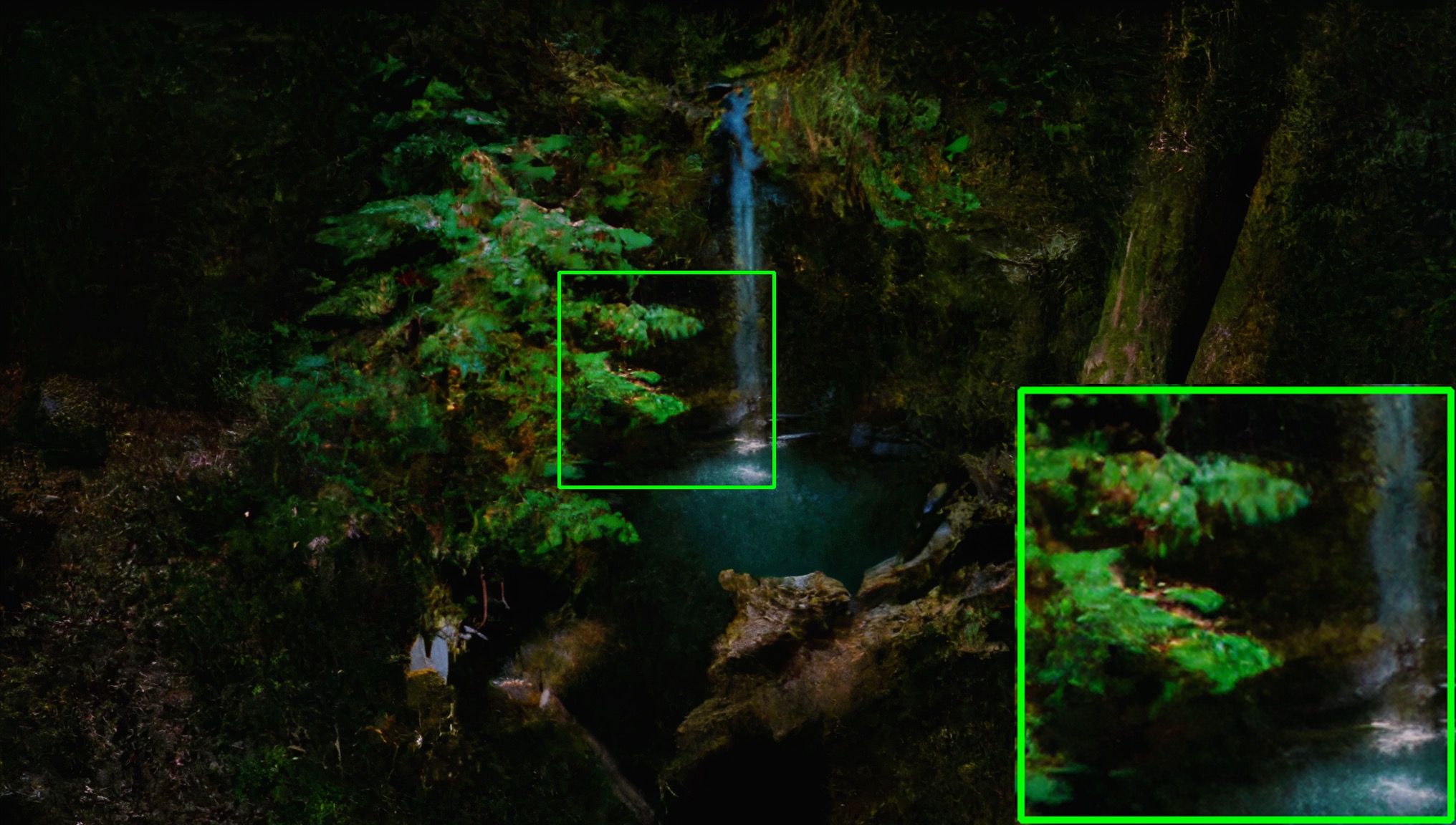}
    \includegraphics[width=\imgwidth]{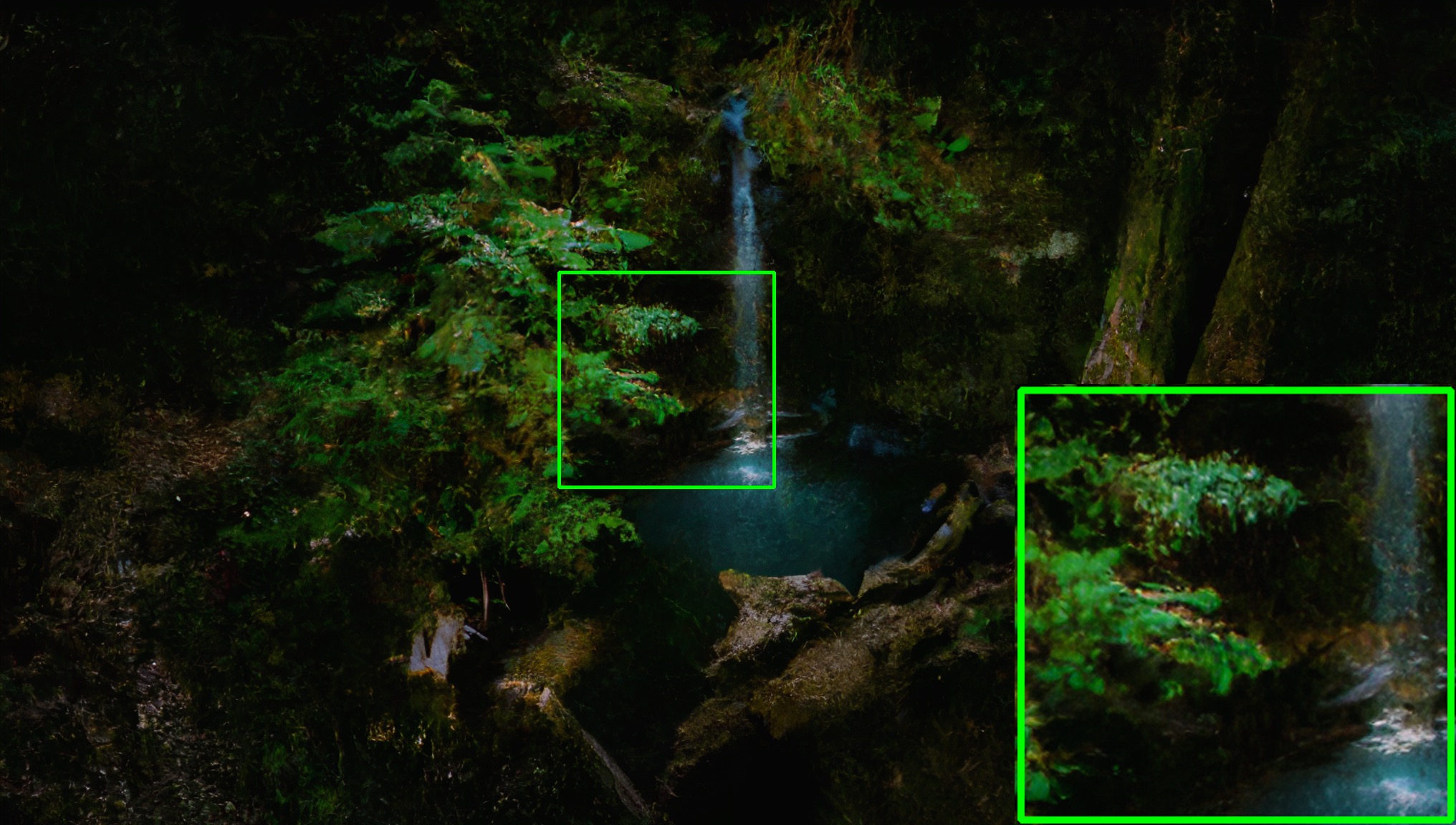}
    \includegraphics[width=\imgwidth]{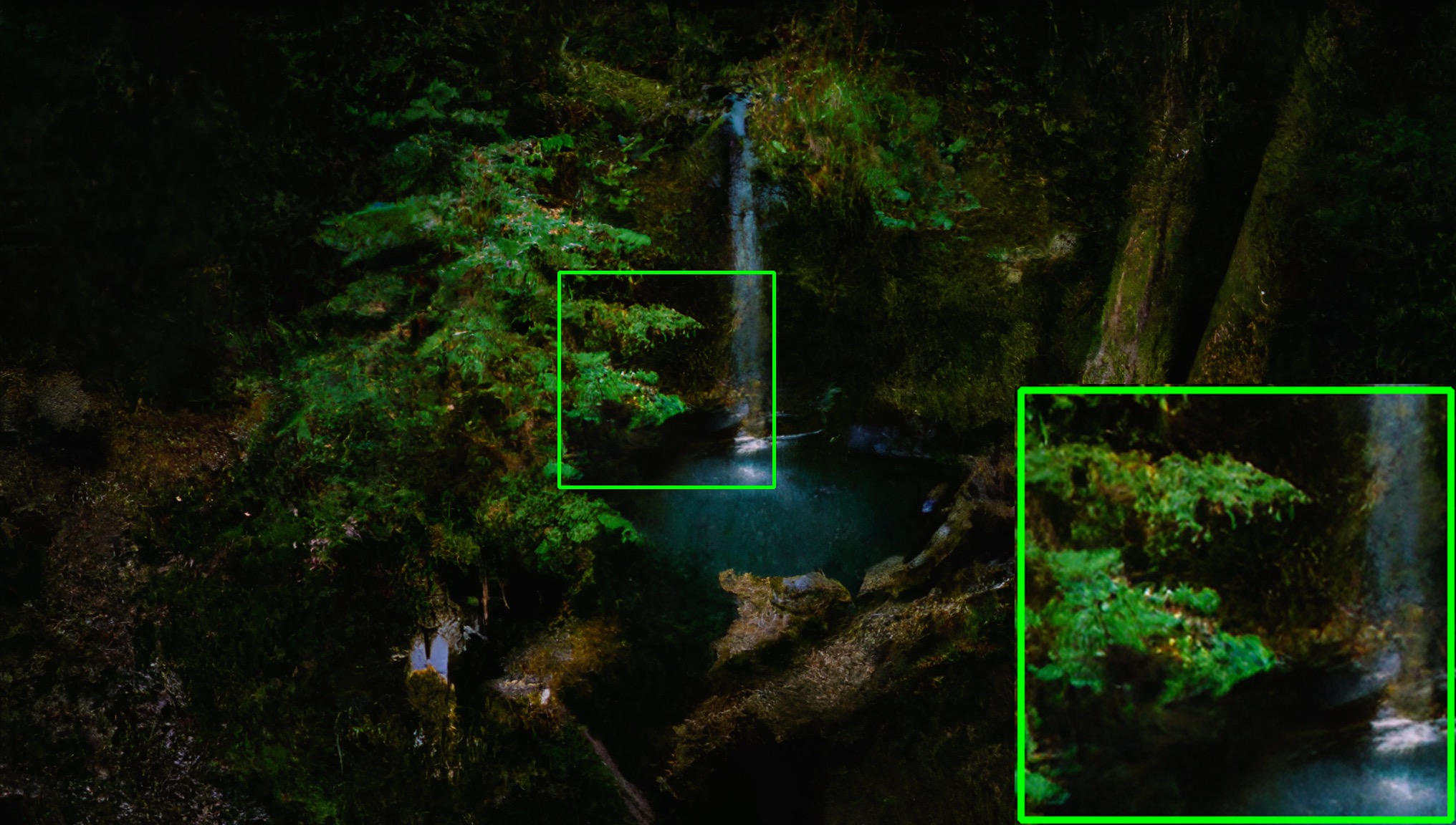}
    \\
    \makebox[\imgwidth]{(d) $\alpha=0.45$}
    \makebox[\imgwidth]{(e) $\alpha=0.50$}
    \makebox[\imgwidth]{(f) $\alpha=0.55$}
    \\
    \includegraphics[width=\imgwidth]{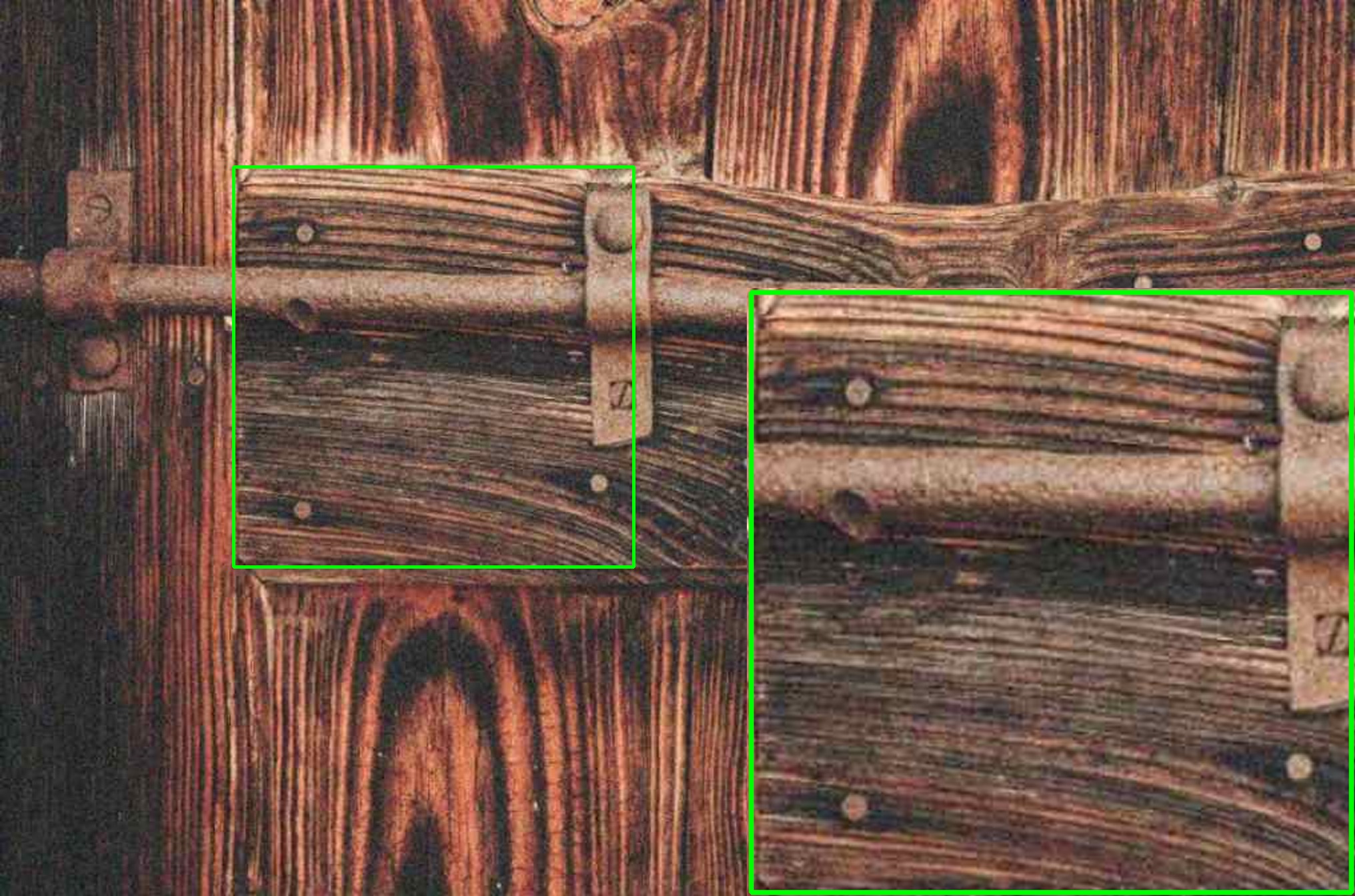}
    \includegraphics[width=\imgwidth]{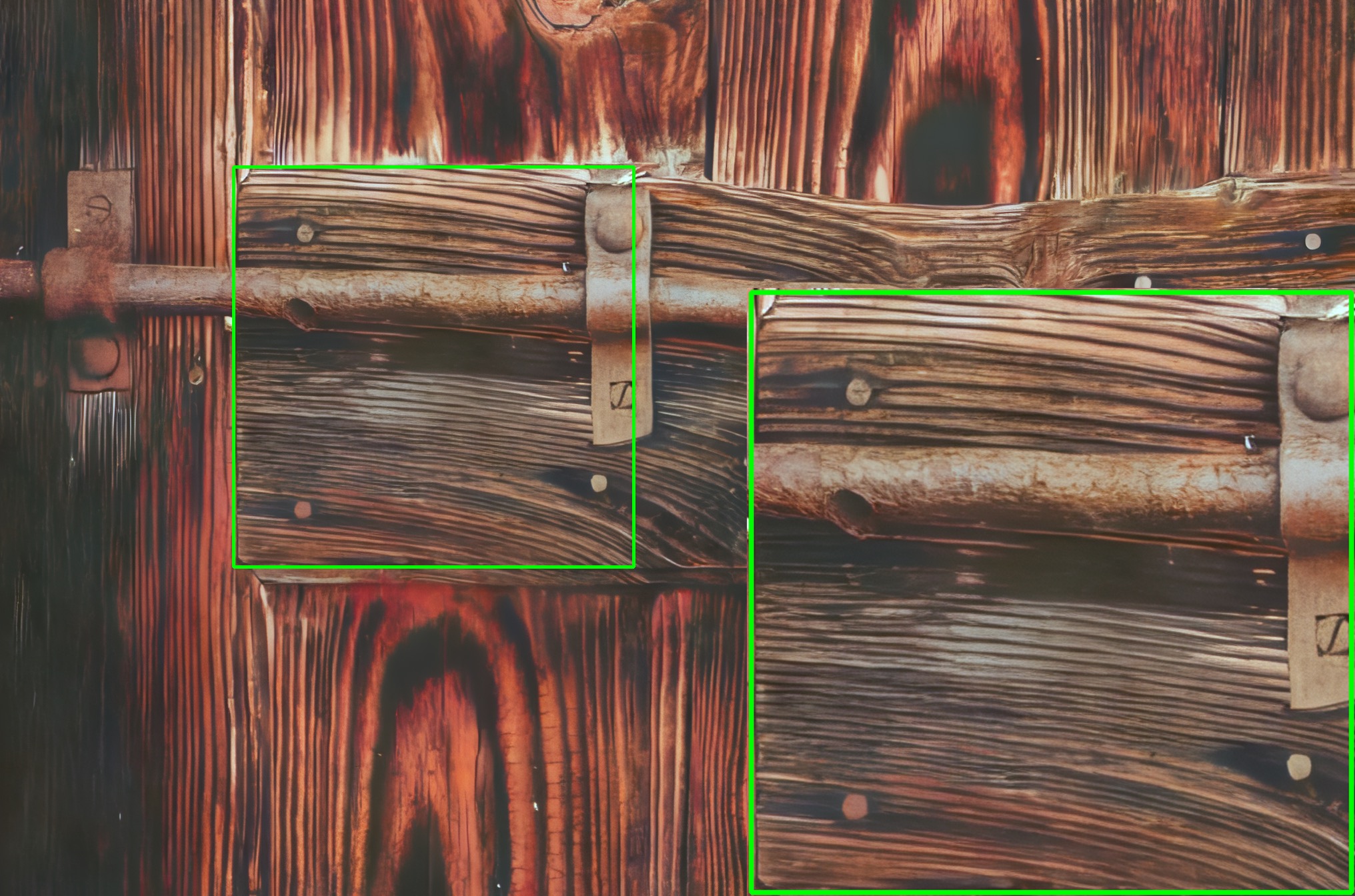}
    \includegraphics[width=\imgwidth]{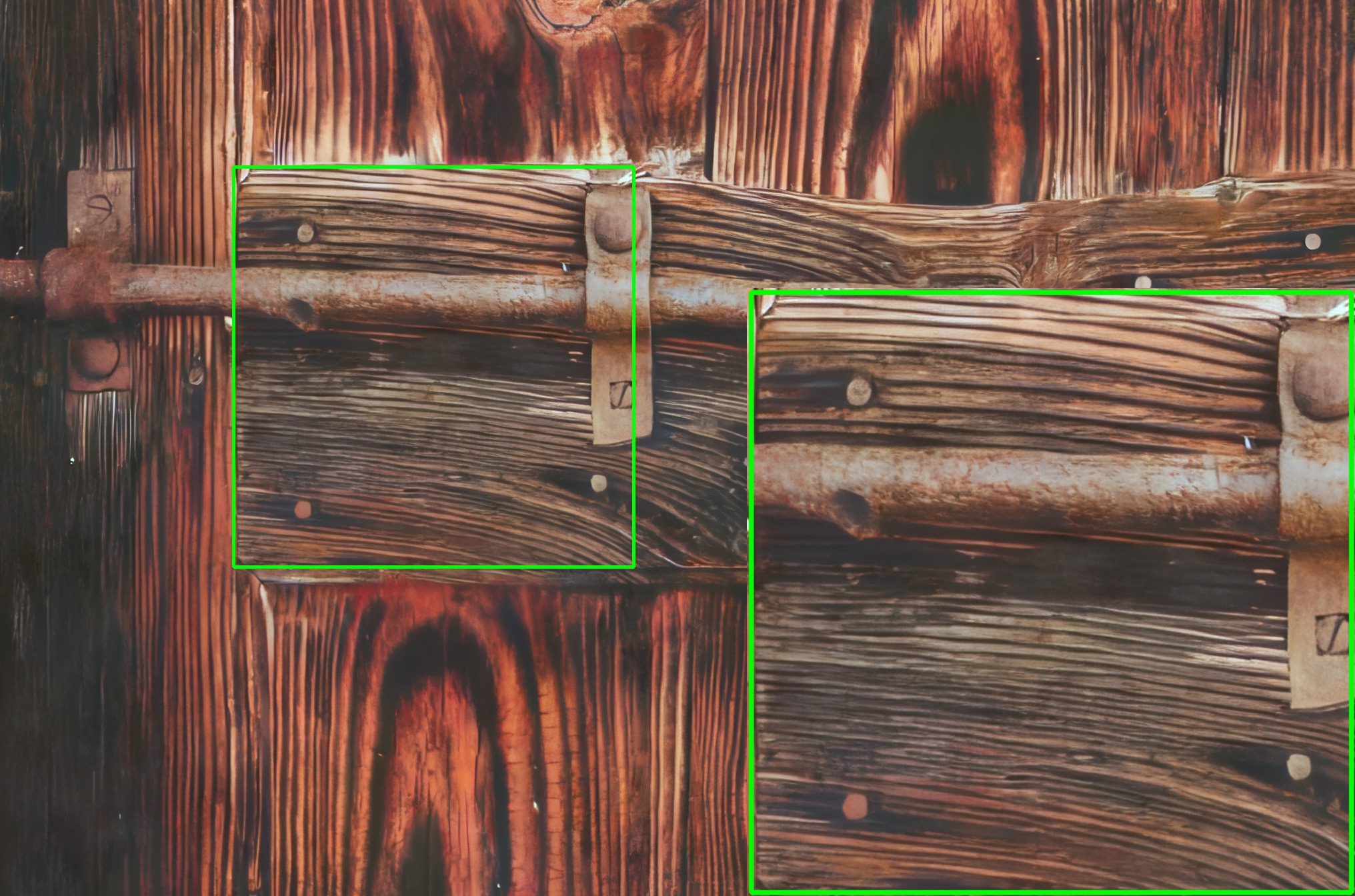}
    \\
    \makebox[\imgwidth]{(a) LQ input}
    \makebox[\imgwidth]{(b) $\alpha=0.35$}
    \makebox[\imgwidth]{(c) $\alpha=0.40$} 
    \\
    \includegraphics[width=\imgwidth]{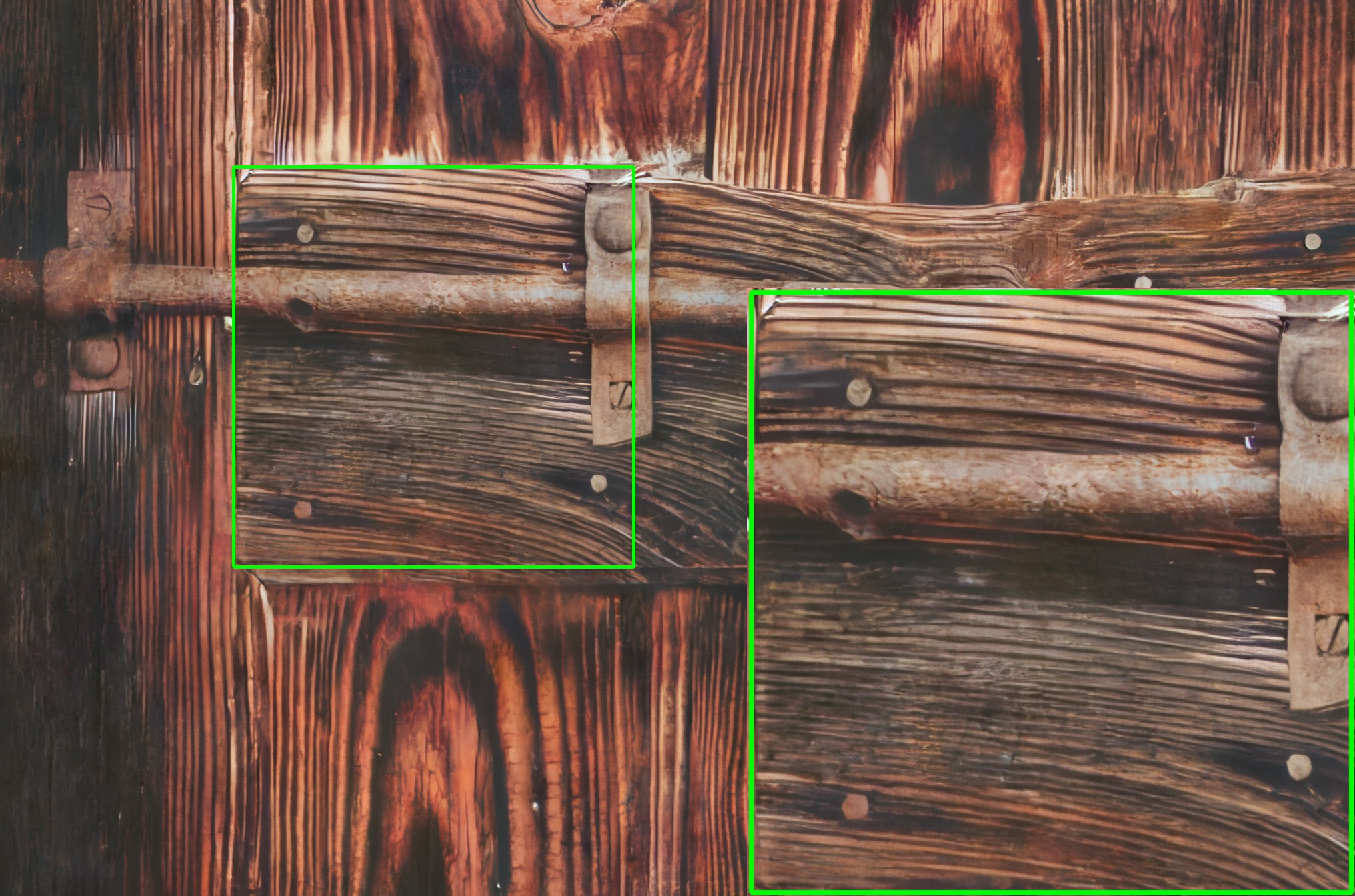}
    \includegraphics[width=\imgwidth]{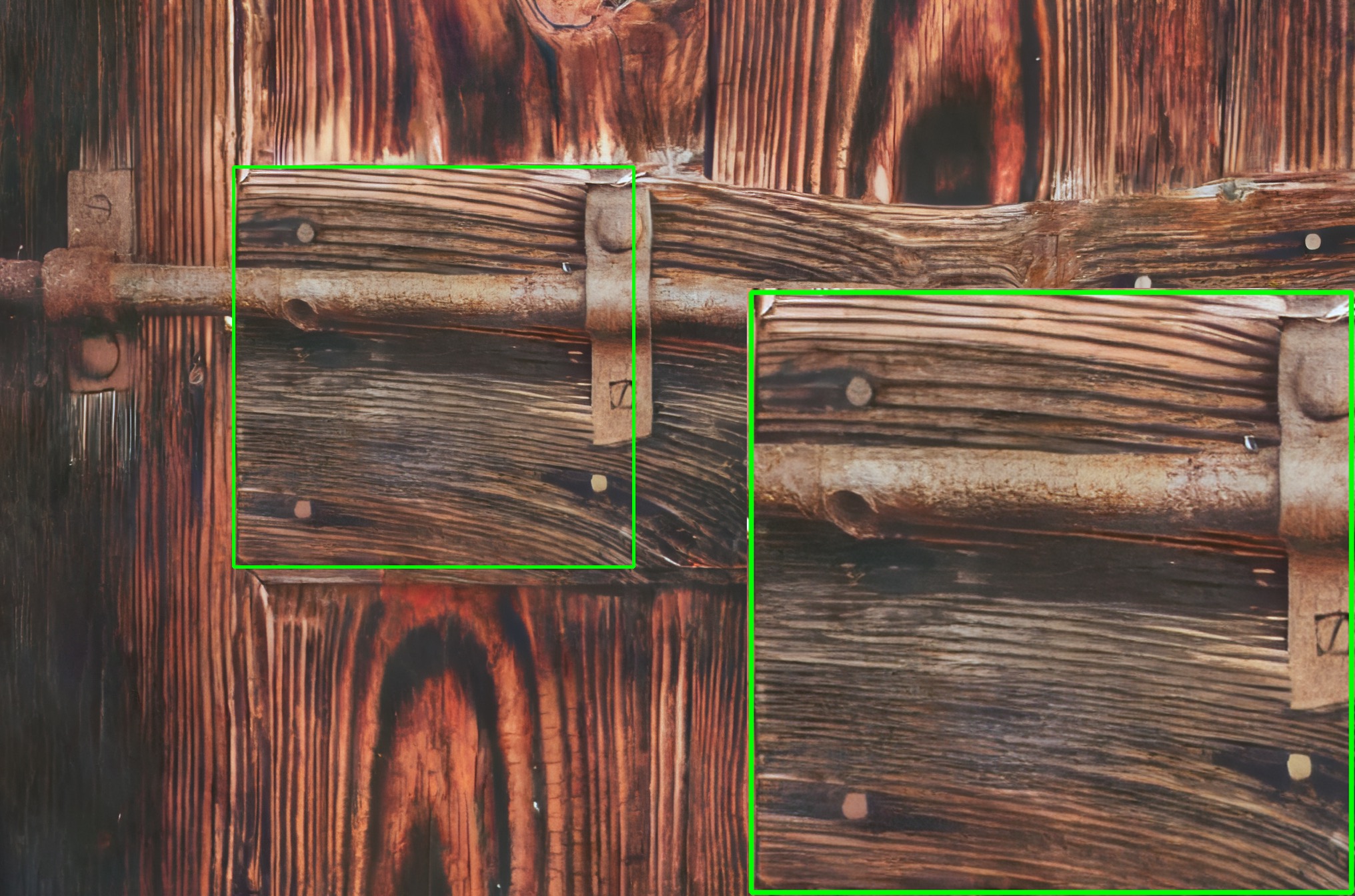}
    \includegraphics[width=\imgwidth]{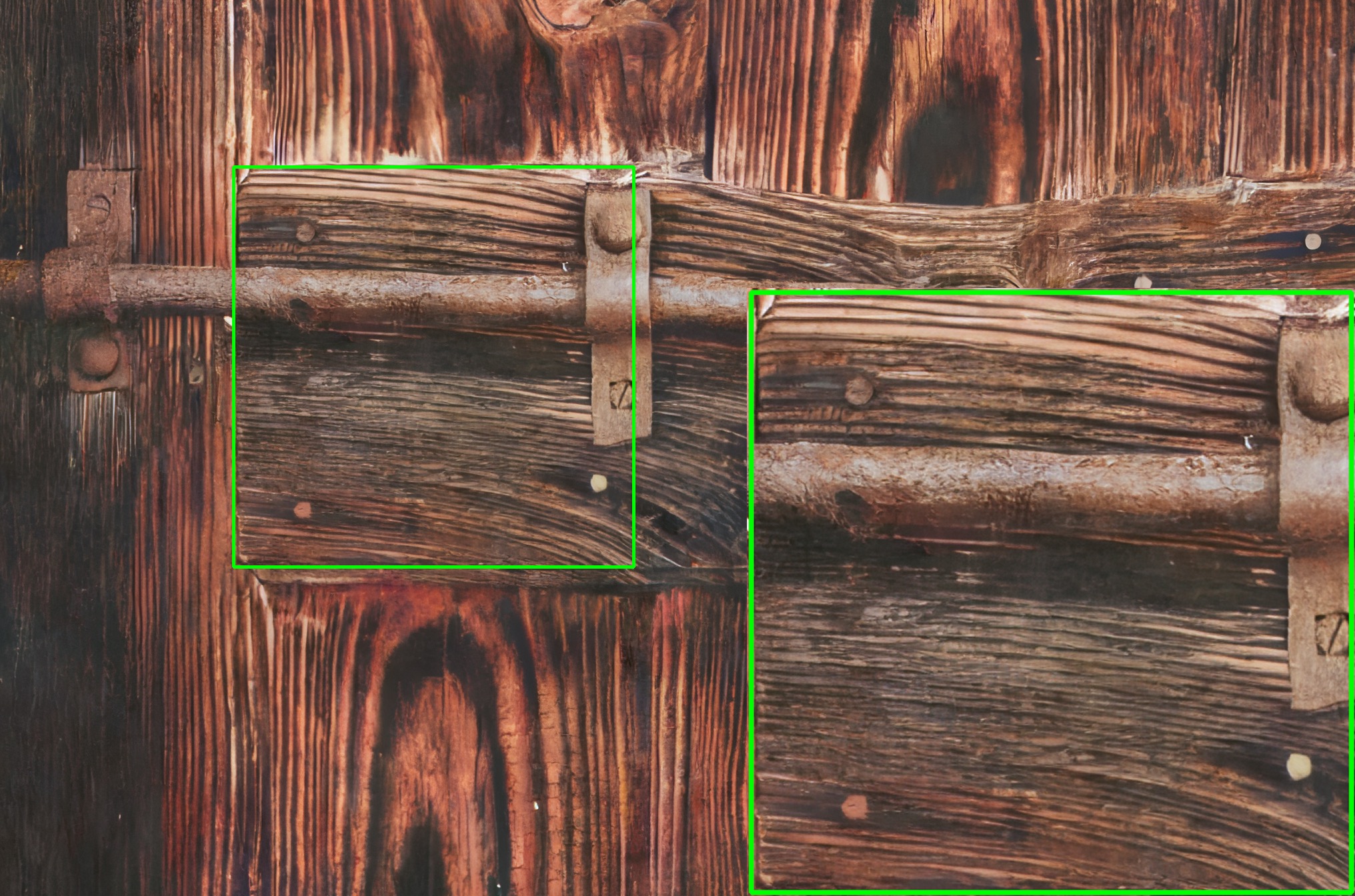}
    \\
    \makebox[\imgwidth]{(d) $\alpha=0.45$}
    \makebox[\imgwidth]{(e) $\alpha=0.50$}
    \makebox[\imgwidth]{(f) $\alpha=0.55$}
    \\
    \caption{Additional results with threshold $\alpha \in \{0.35, 0.40, 0.45, 0.50, 0.55\}$. (Zoom in for best view)}
    \label{fig:supp_threshold}
\end{figure*}

We present more examples with different threshold $\alpha$ in \cref{fig:supp_threshold}. It can be observed that by increasing $\alpha$ from $0.35$ to $0.55$, we can gradually increase the texture strength in the final results.

\subsection{Comparison with LDM-BSR}

Examples in \cref{fig:ldmbsr} illustrate why the quantitative results of LDM-BSR are not satisfactory on RWSR in Tab. 1 of the main paper. We can observe that although LDM-BSR is able to generate sharper edges for the blurry LQ inputs, it has difficulties to eliminate other complex distortions. Because of explicit distortion removal module, our proposed ITER does not have such problem. 

\begin{figure*}[h]
    \centering
    \newcommand{\imgwidth}{0.32\linewidth}
    \includegraphics[width=\linewidth]{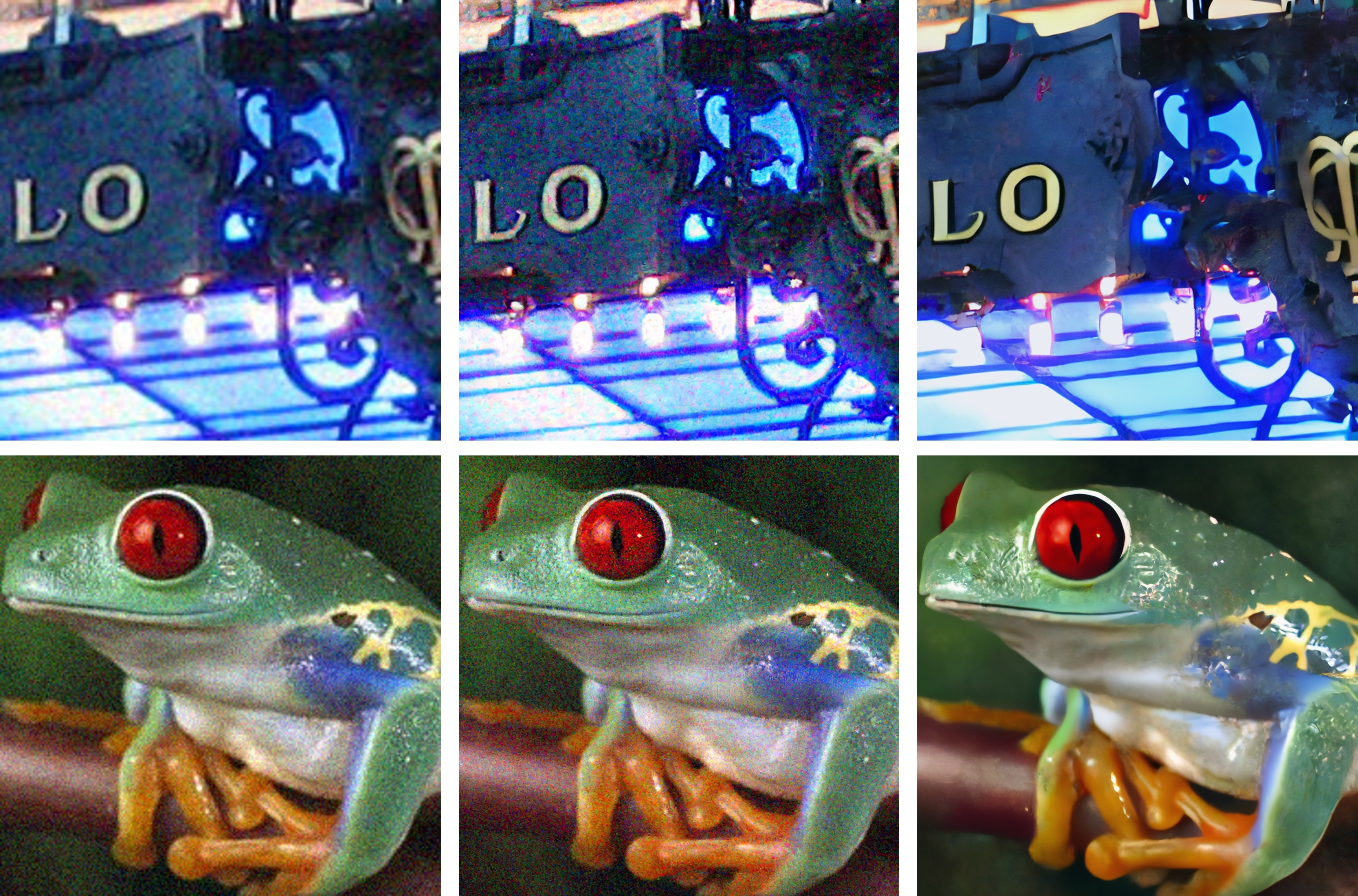}
    \makebox[\imgwidth]{\small (a) LQ input} 
    \makebox[\imgwidth]{\small (b) LDM-BSR} 
    \makebox[\imgwidth]{\small (c) \textbf{ITER (Ours)}}
    \caption{Problem of LDM-BSR without explicit distortion removal. Examples are from RealSRSet \cite{zhang2021designing}. (Zoom in for best view)}
    \label{fig:ldmbsr}
\end{figure*}

\subsection{Additional Results on Real-World Benchmarks}

\begin{figure*}[h]
    \centering
    \newcommand{\imgwidth}{0.33\linewidth}
    \includegraphics[width=\imgwidth]{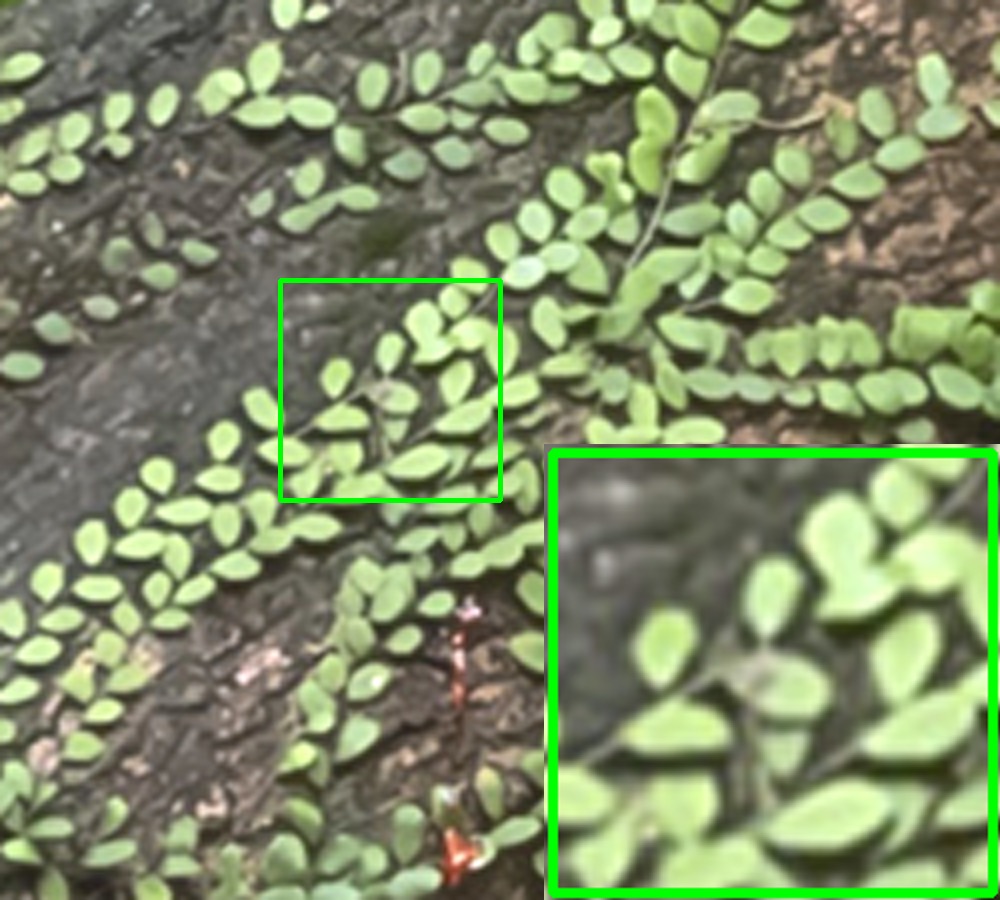}
    \includegraphics[width=\imgwidth]{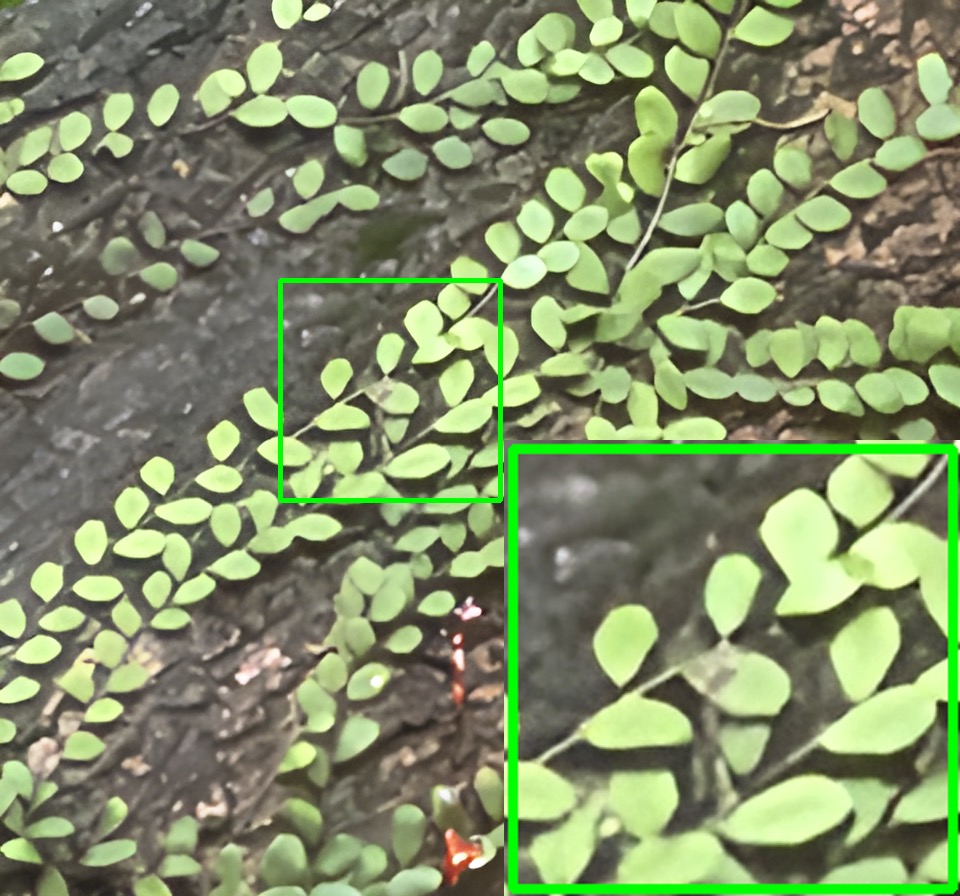}
    \includegraphics[width=\imgwidth]{compare_fema_Nikon_014_LR4.jpg}
    \\
    \makebox[\imgwidth]{(a) Real-world LQ input}
    \makebox[\imgwidth]{(b) Real-ESRGAN \cite{wang2021real}}
    \makebox[\imgwidth]{(c) FeMaSR \cite{chen2022femasr}}
    \\
    \includegraphics[width=\imgwidth]{compare_ldmbsr_Nikon_014_LR4.jpg}
    \includegraphics[width=\imgwidth]{compare_mmrealsr_Nikon_014_LR4.jpg}
    \includegraphics[width=\imgwidth]{compare_iter_Nikon_014_LR4.jpg}
    \\
    \makebox[\imgwidth]{(d) LDM-BSR \cite{rombach2022latentdiffusion}}
    \makebox[\imgwidth]{(e) MM-RealSR \cite{mou2022mmrealsr}}
    \makebox[\imgwidth]{(f) \textbf{ITER (Ours)} }
    \\ \vspace{1em} 
    \includegraphics[width=\imgwidth]{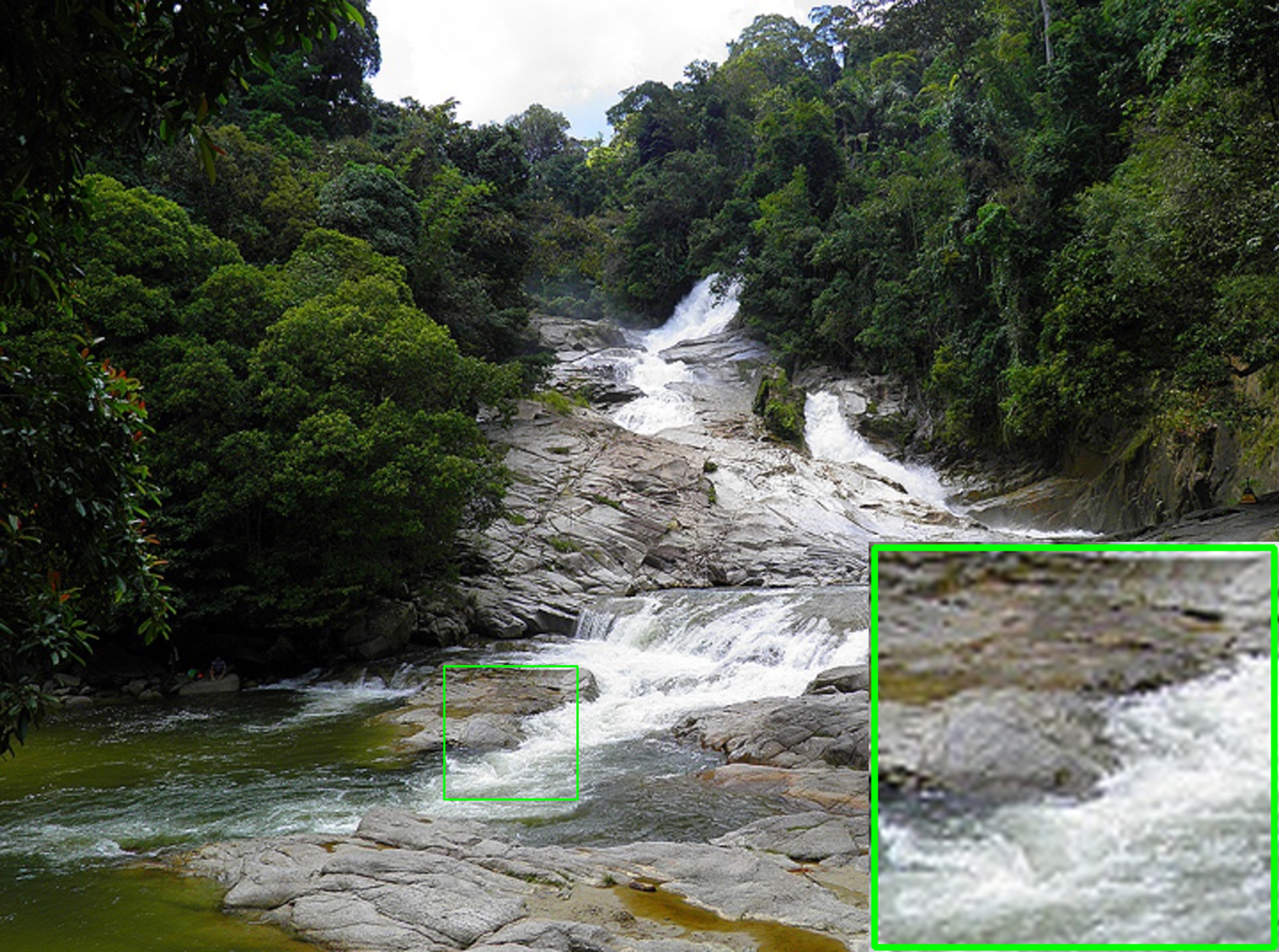}
    \includegraphics[width=\imgwidth]{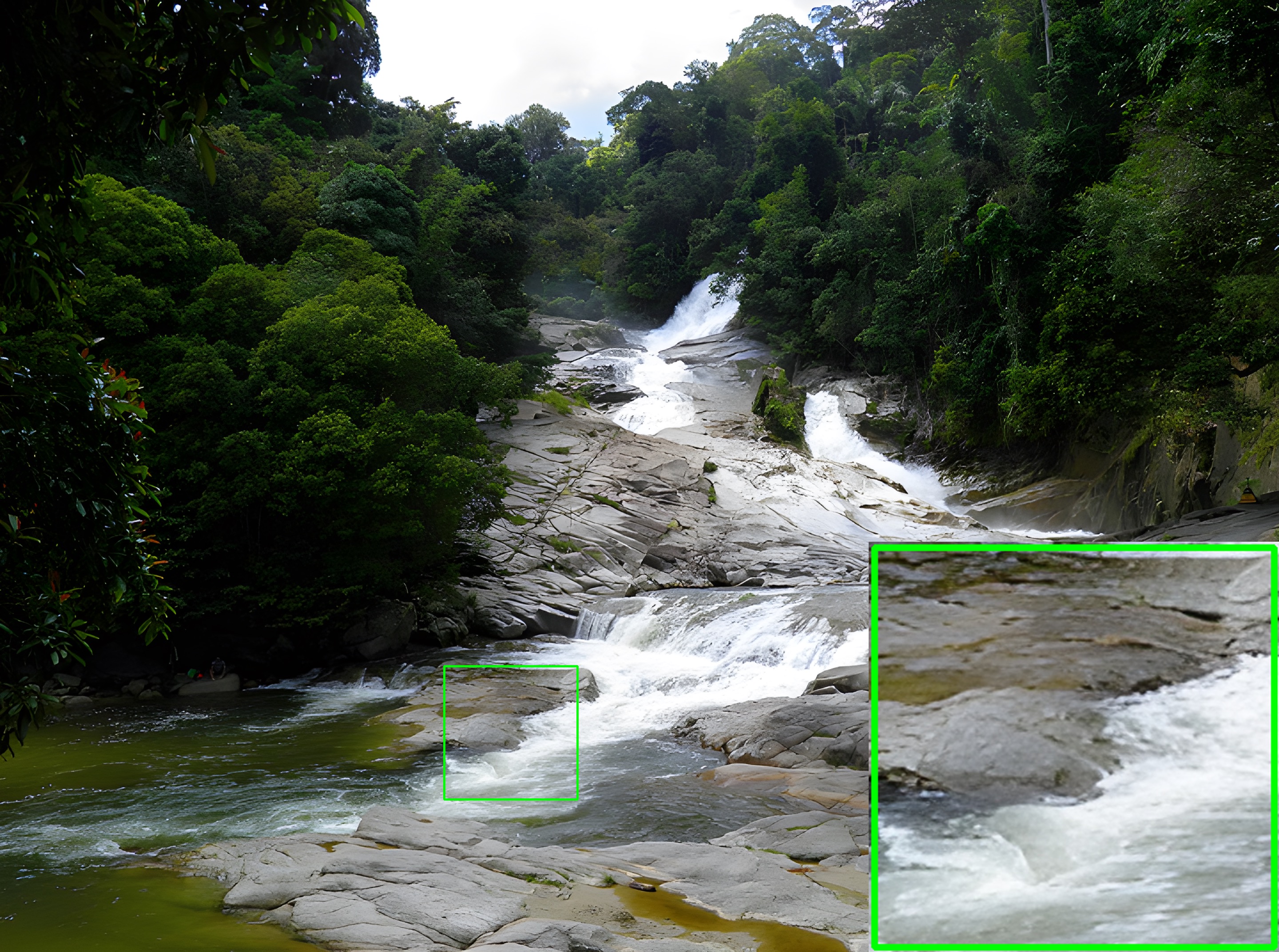}
    \includegraphics[width=\imgwidth]{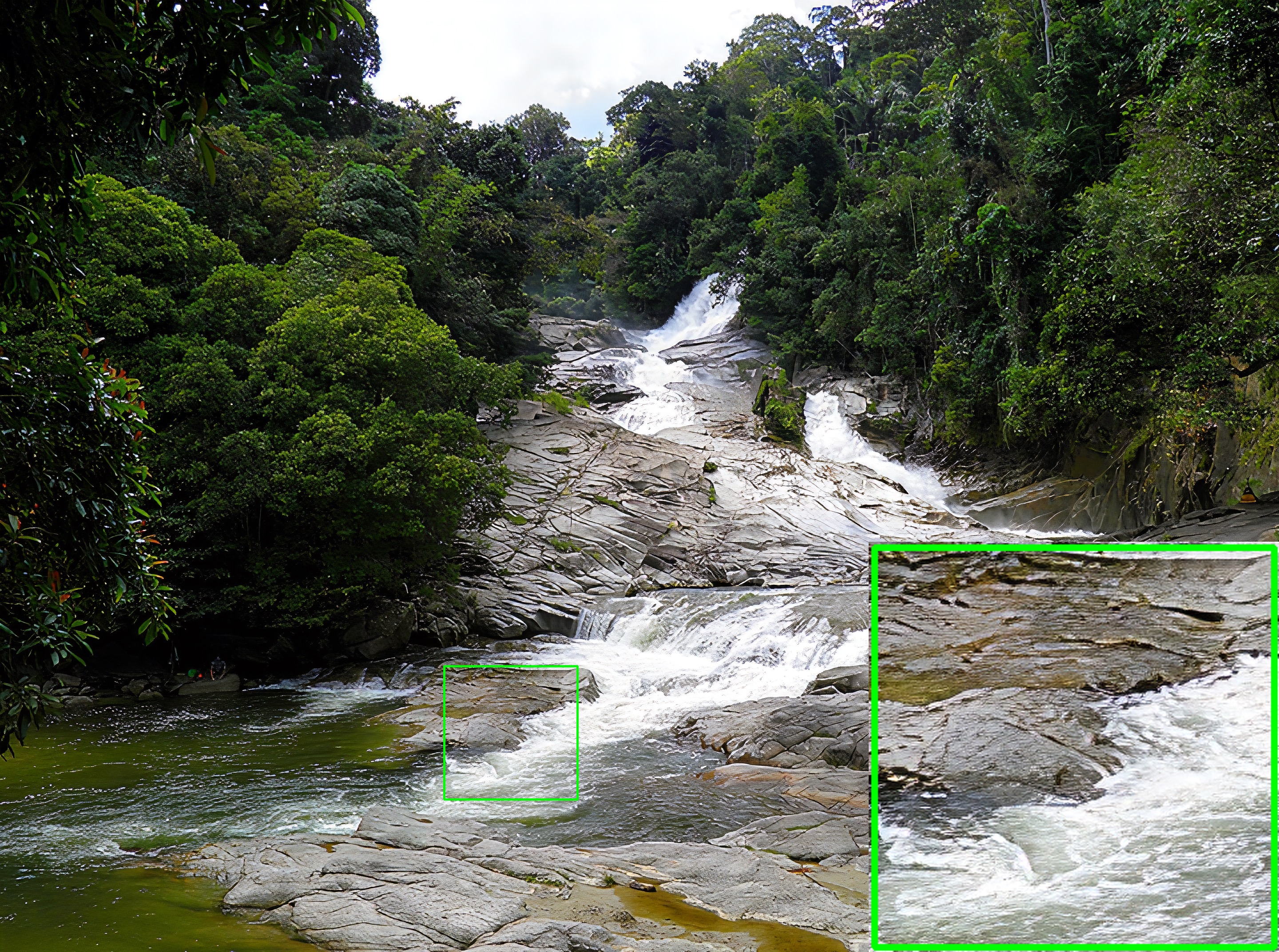}
    \\
    \makebox[\imgwidth]{(a) Real-world LQ input}
    \makebox[\imgwidth]{(b) Real-ESRGAN \cite{wang2021real}}
    \makebox[\imgwidth]{(c) FeMaSR \cite{chen2022femasr}}
    \\
    \includegraphics[width=\imgwidth]{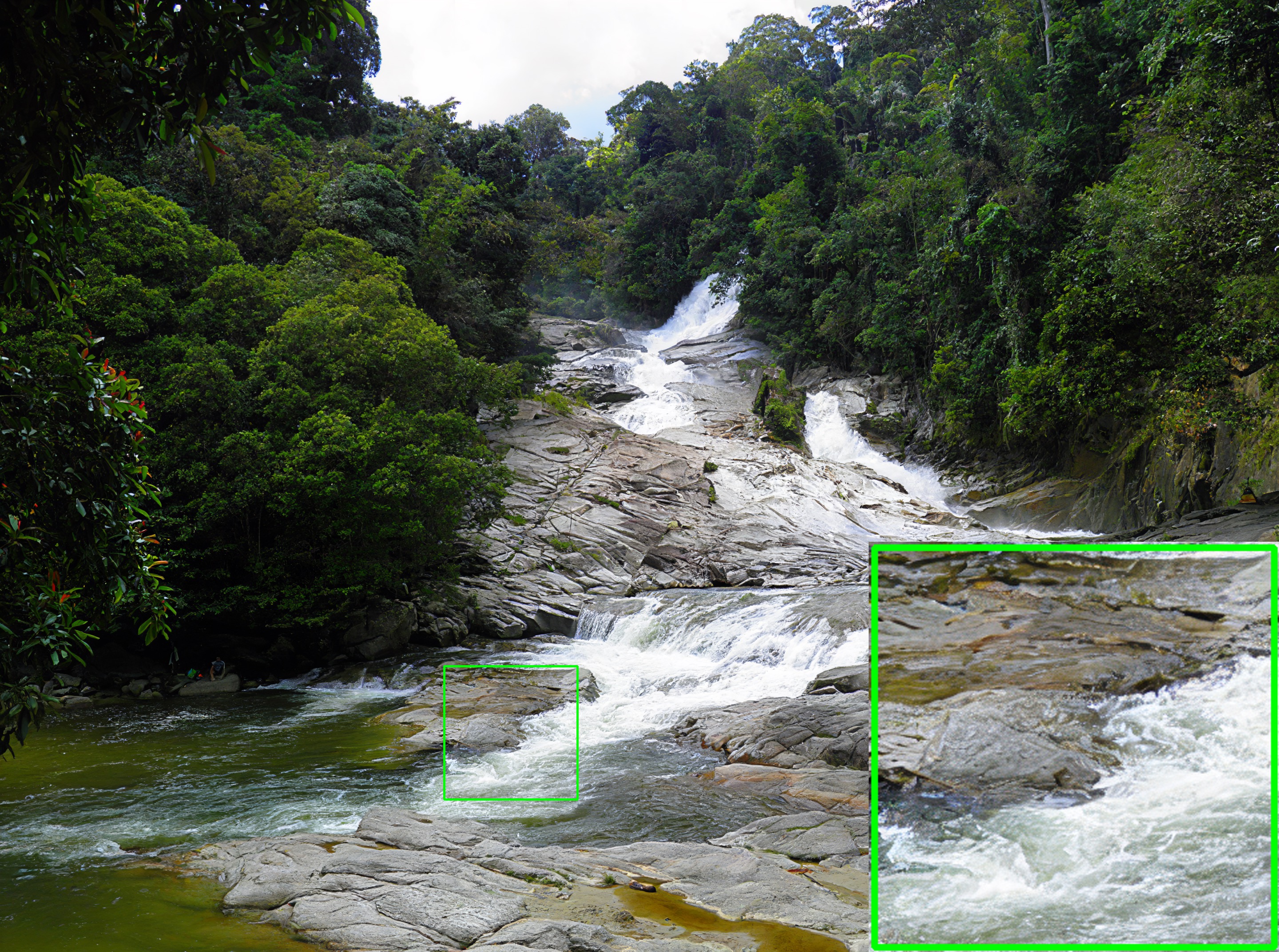}
    \includegraphics[width=\imgwidth]{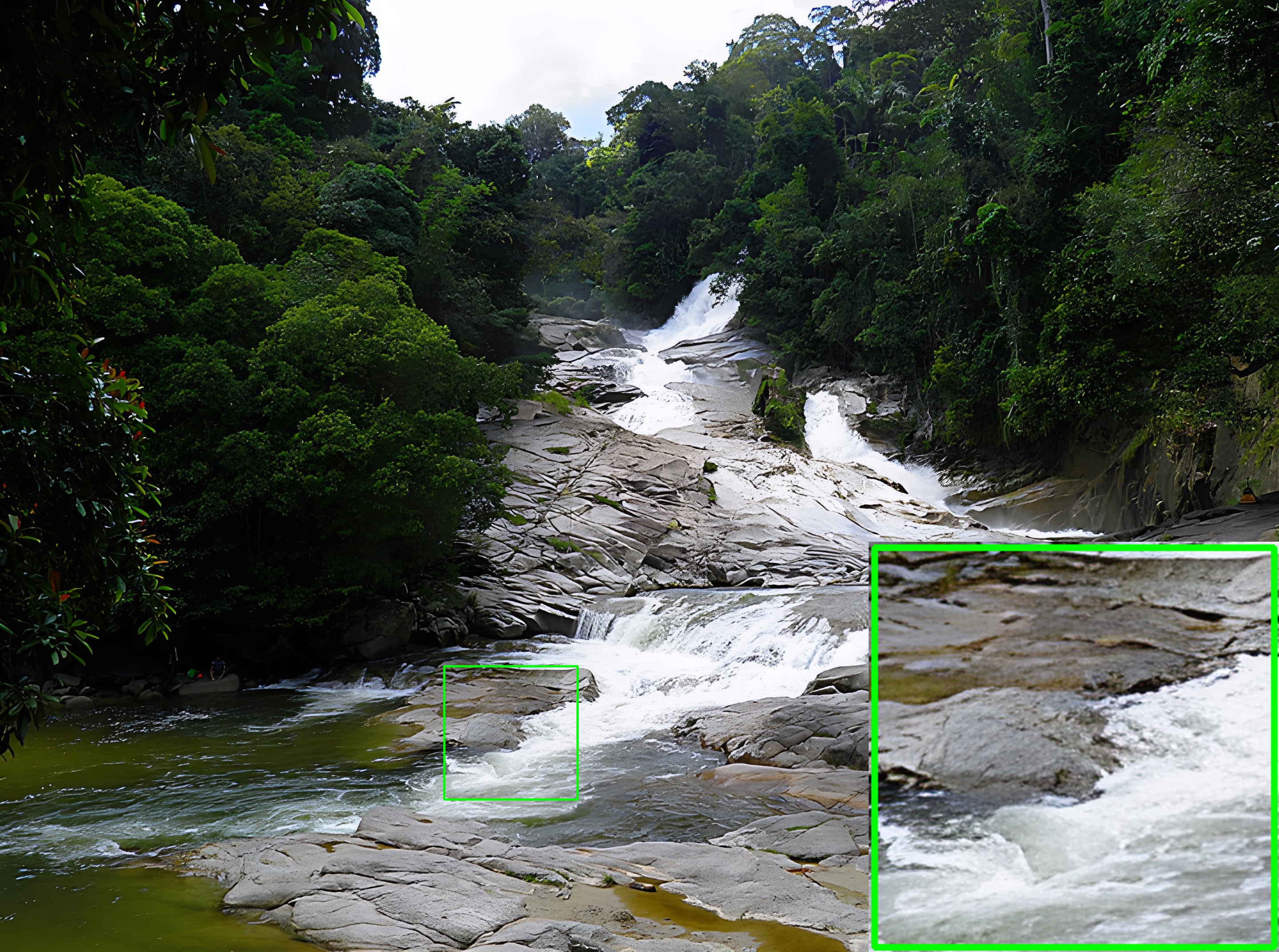}
    \includegraphics[width=\imgwidth]{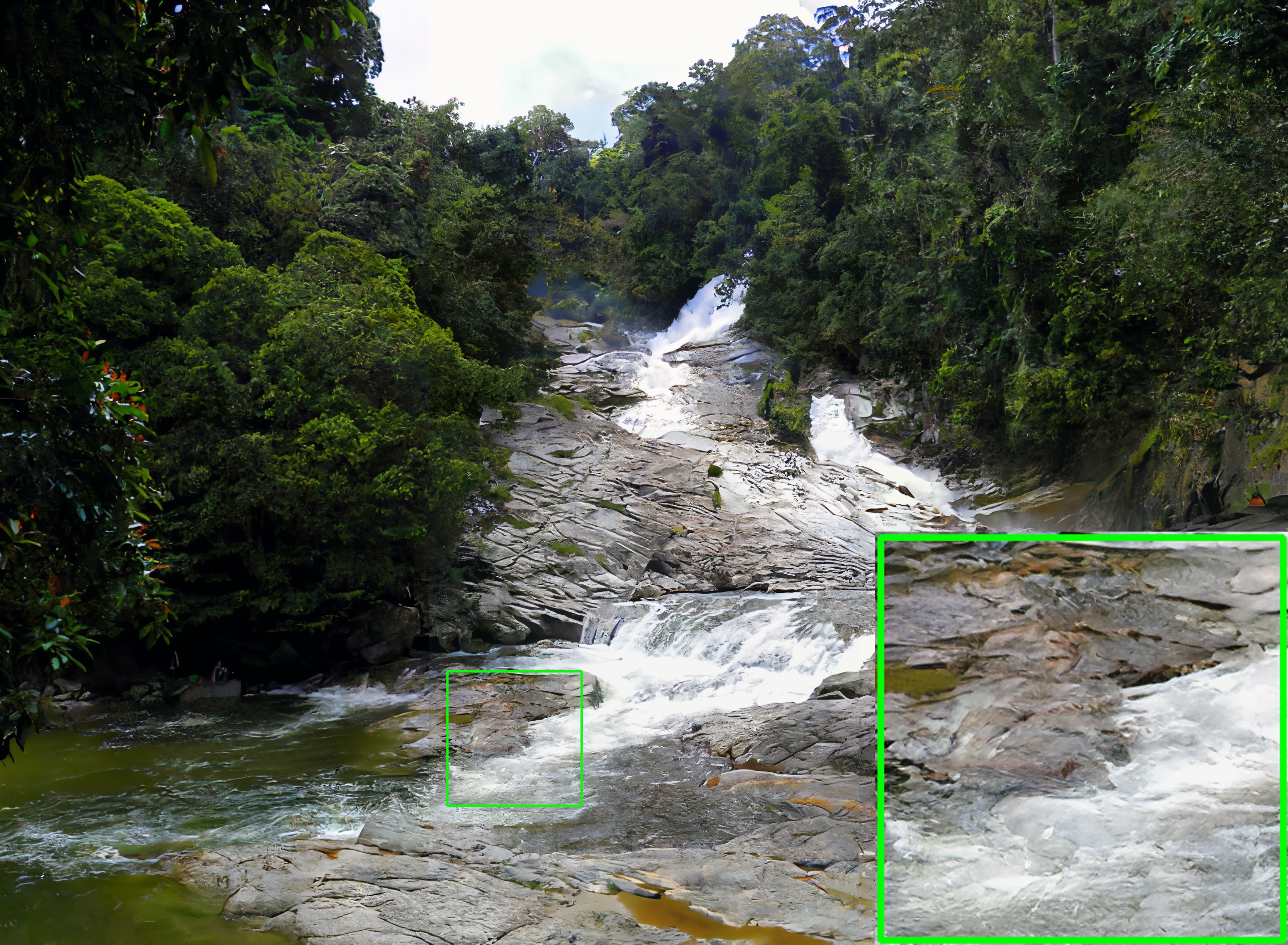}
    \\
    \makebox[\imgwidth]{(d) LDM-BSR  \cite{rombach2022latentdiffusion}}
    \makebox[\imgwidth]{(e) MM-RealSR \cite{mou2022mmrealsr}}
    \makebox[\imgwidth]{(f) \textbf{ITER (Ours)} }
    \caption{Additional results from real-world benchmarks.}
    \label{fig:supp_vis1}
\end{figure*}

\begin{figure*}[h]
    \centering
    \newcommand{\imgwidth}{0.33\linewidth}
    \includegraphics[width=\imgwidth]{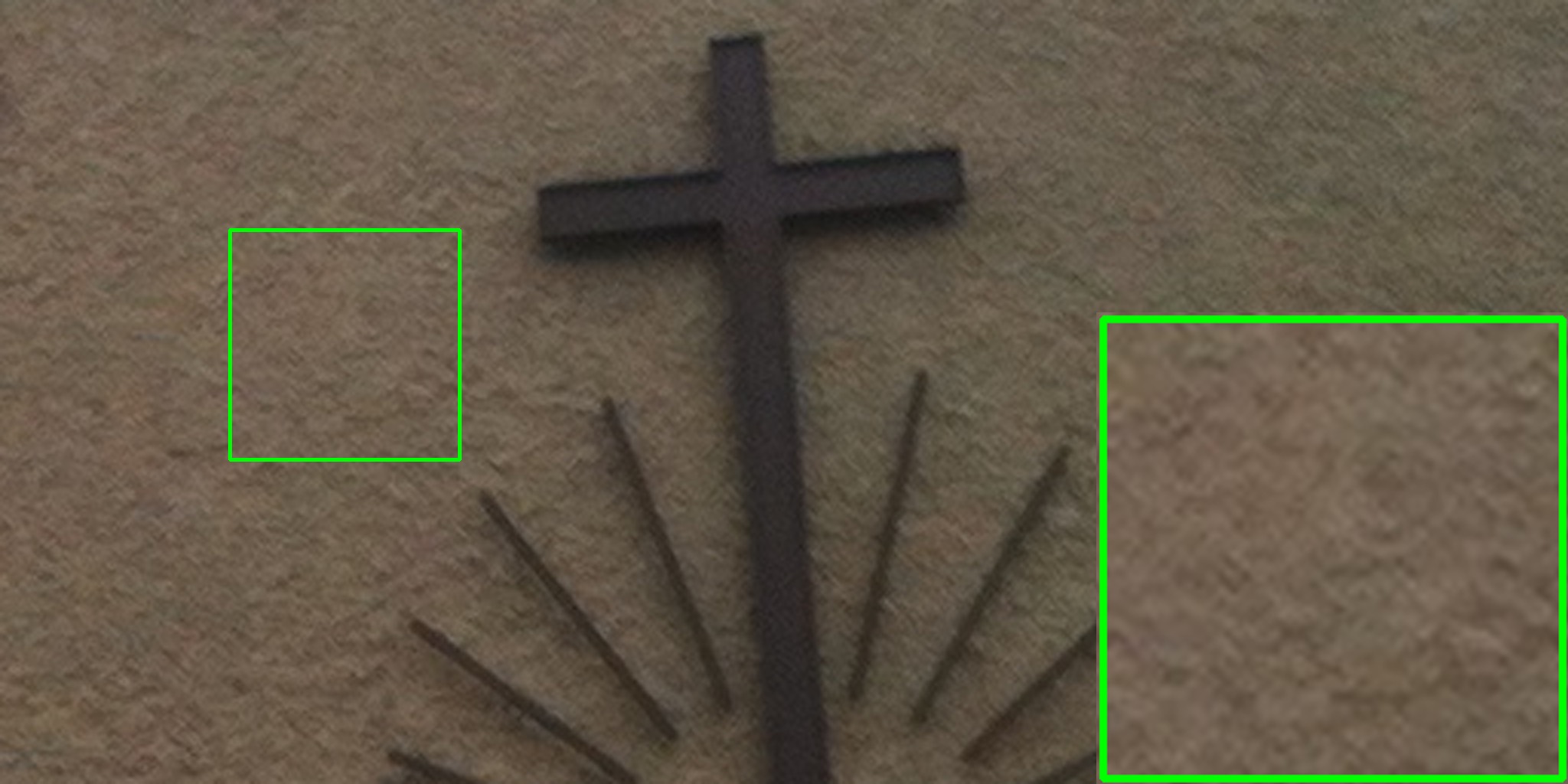}
    \includegraphics[width=\imgwidth]{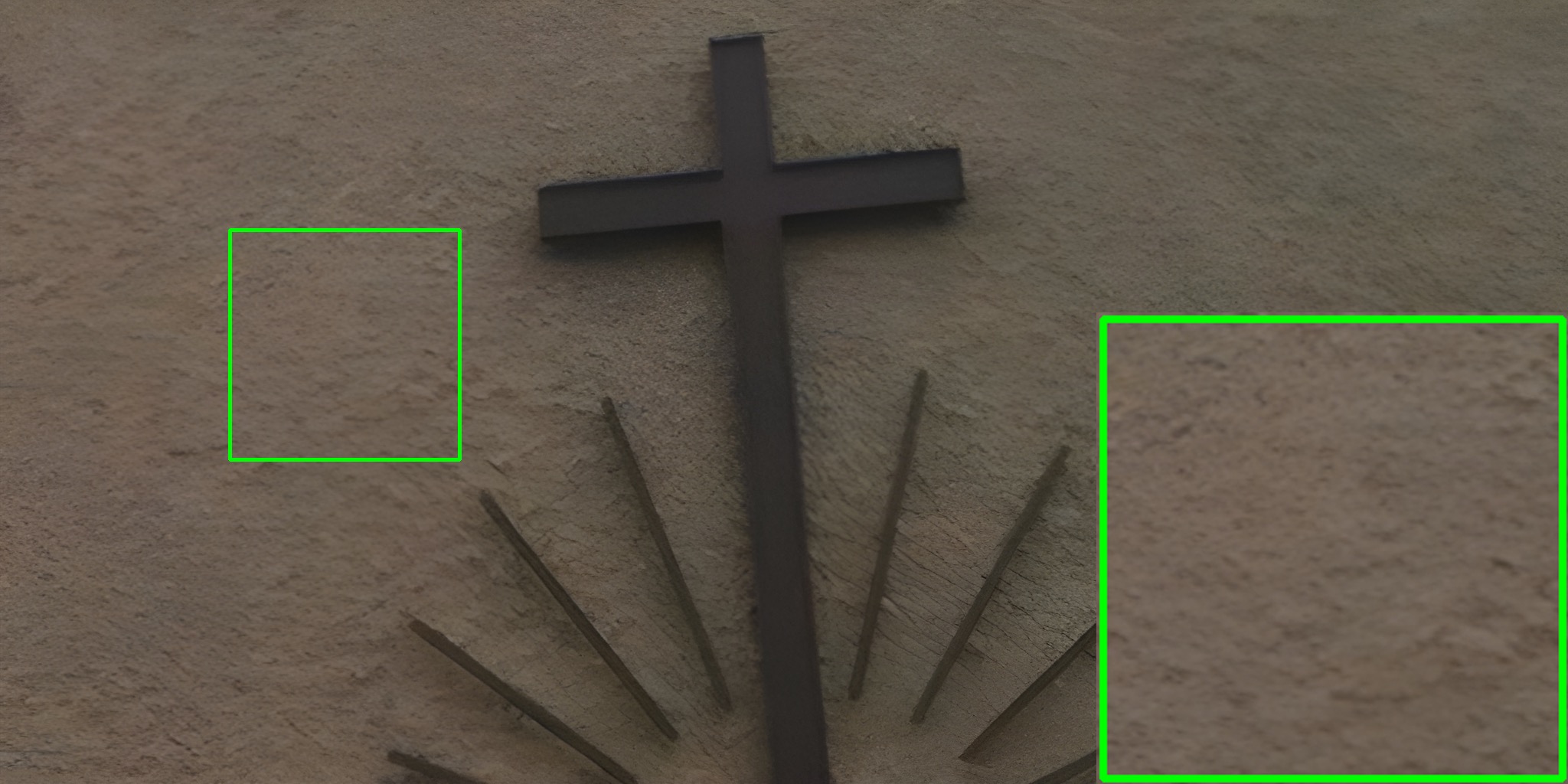}
    \includegraphics[width=\imgwidth]{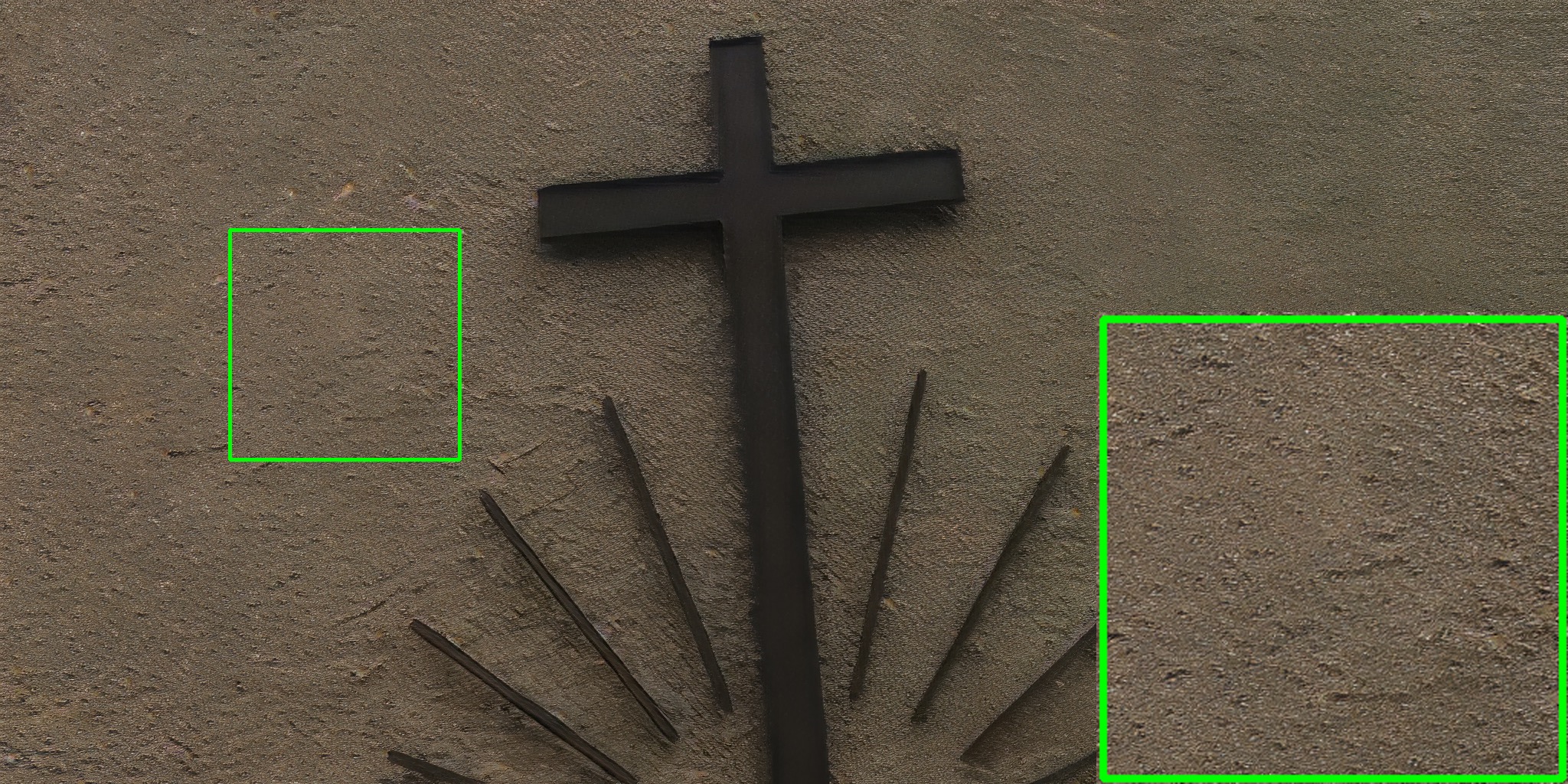}
    \\
    \makebox[\imgwidth]{(a) Real-world LQ input }
    \makebox[\imgwidth]{(b) Real-ESRGAN \cite{wang2021real}}
    \makebox[\imgwidth]{(c) FeMaSR \cite{chen2022femasr}}
    \\
    \includegraphics[width=\imgwidth]{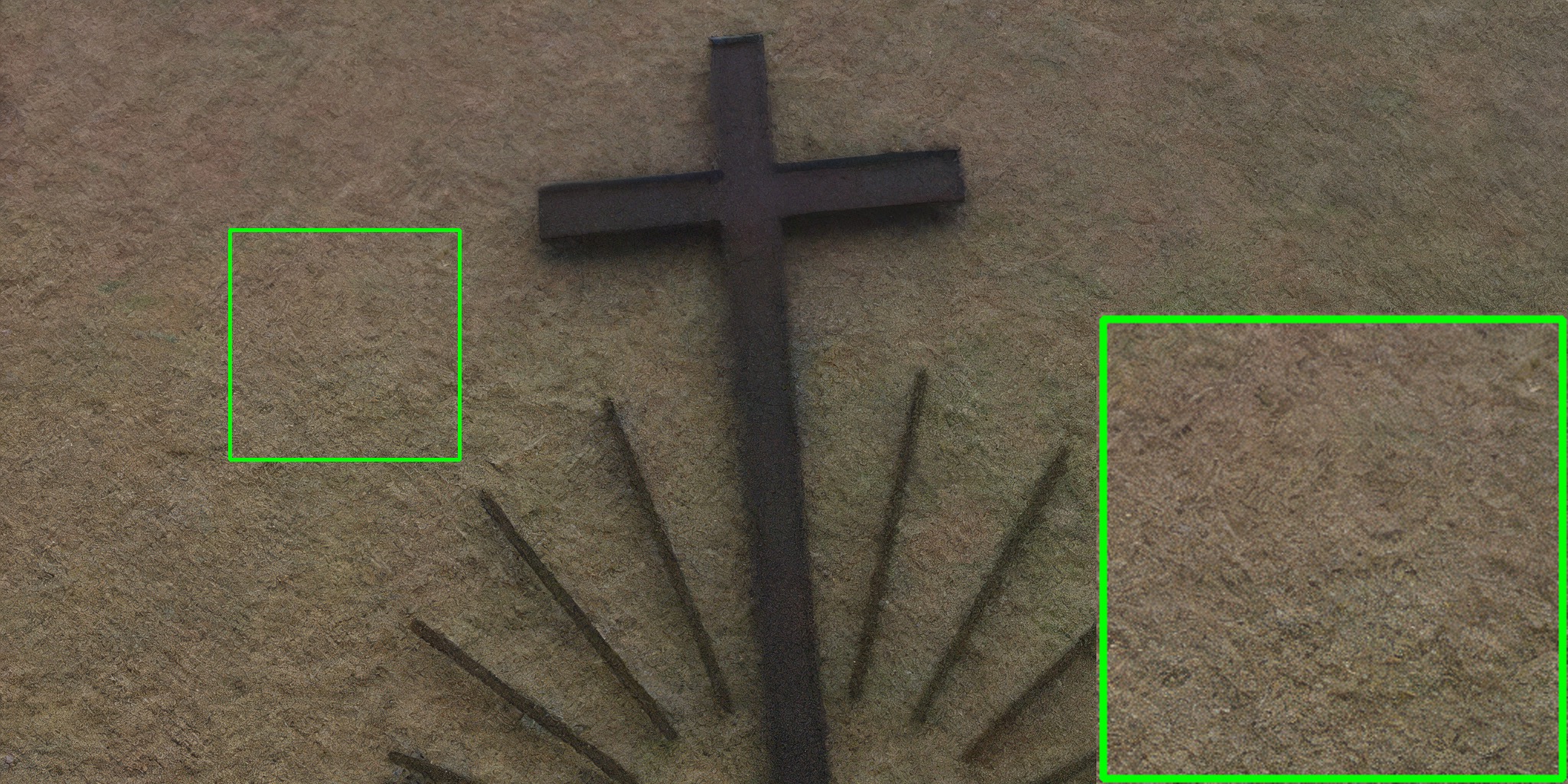}
    \includegraphics[width=\imgwidth]{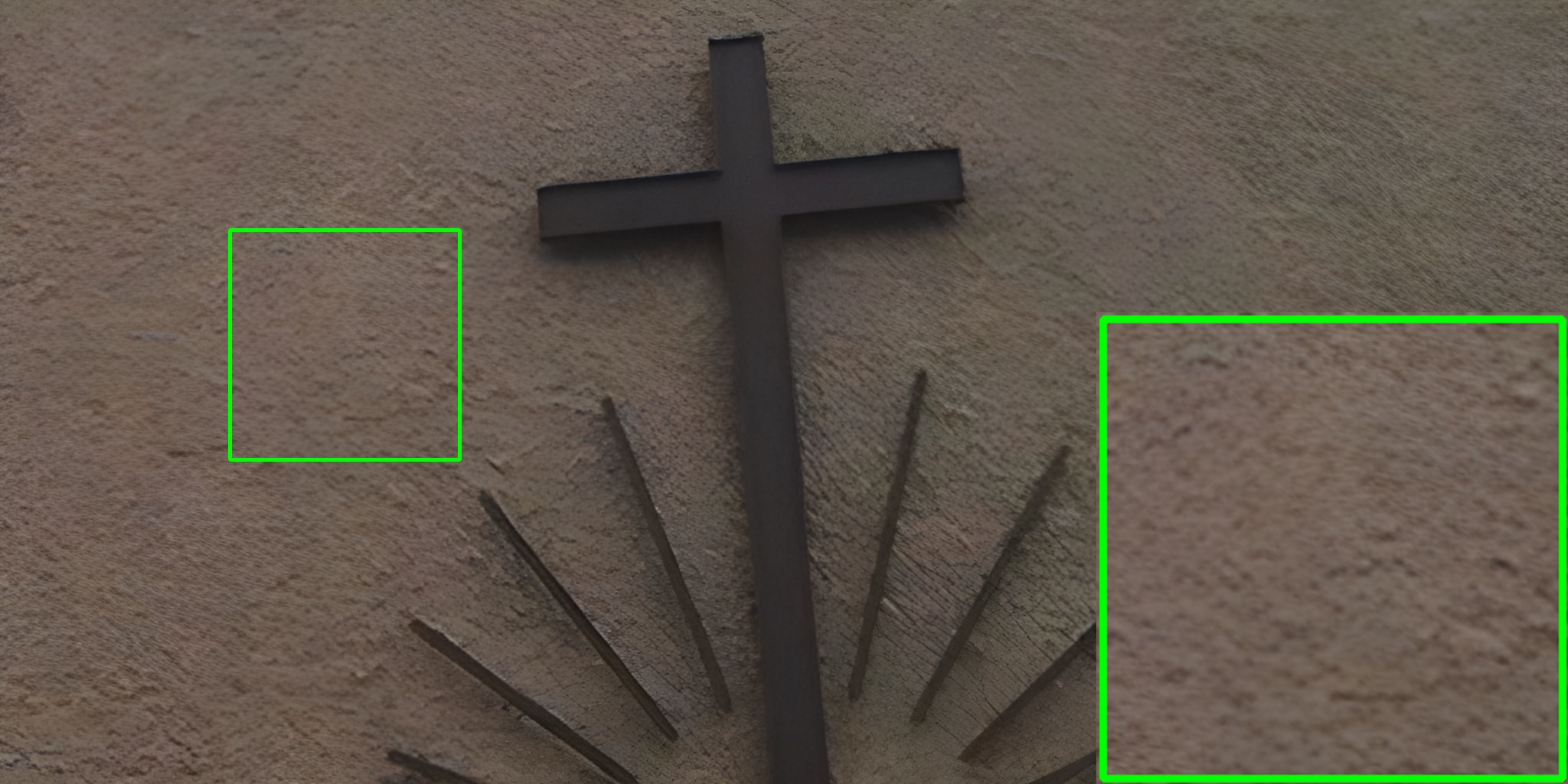}
    \includegraphics[width=\imgwidth]{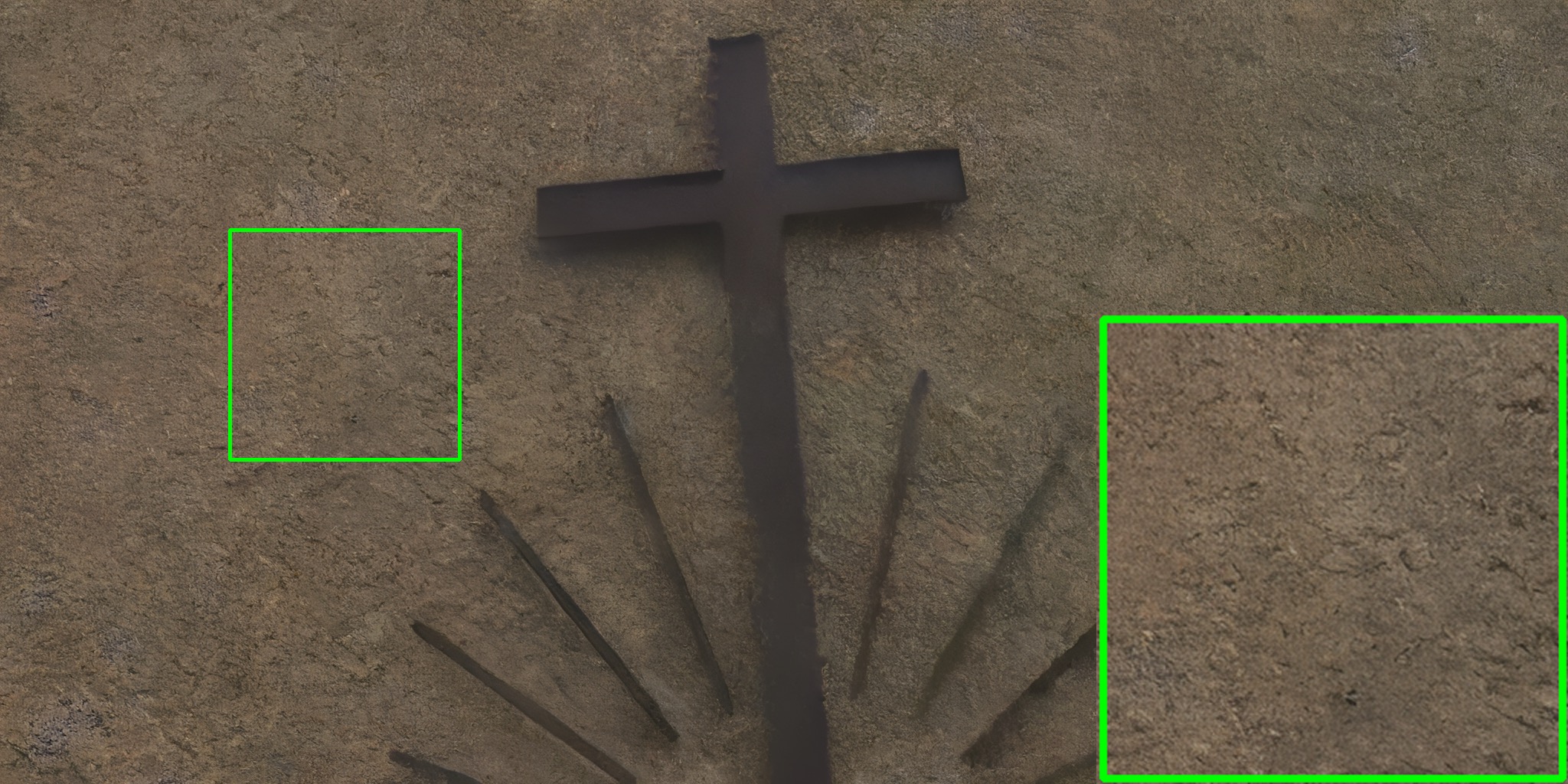}
    \\
    \makebox[\imgwidth]{(d) LDM-BSR \cite{rombach2022latentdiffusion}}
    \makebox[\imgwidth]{(e) MM-RealSR \cite{mou2022mmrealsr}}
    \makebox[\imgwidth]{(f) \textbf{ITER (Ours)}}
    \\ \vspace{1em} 
    \includegraphics[width=\imgwidth]{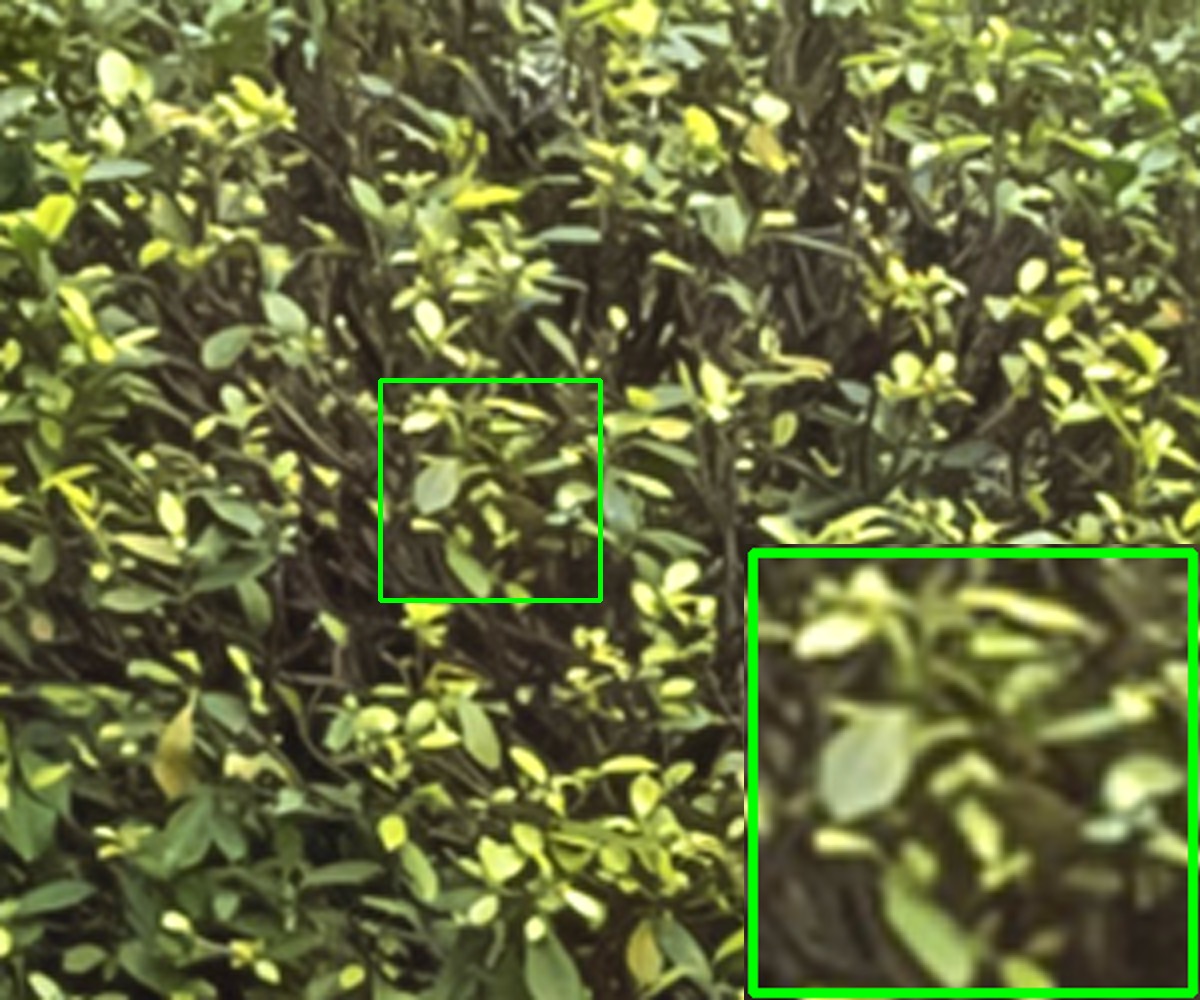}
    \includegraphics[width=\imgwidth]{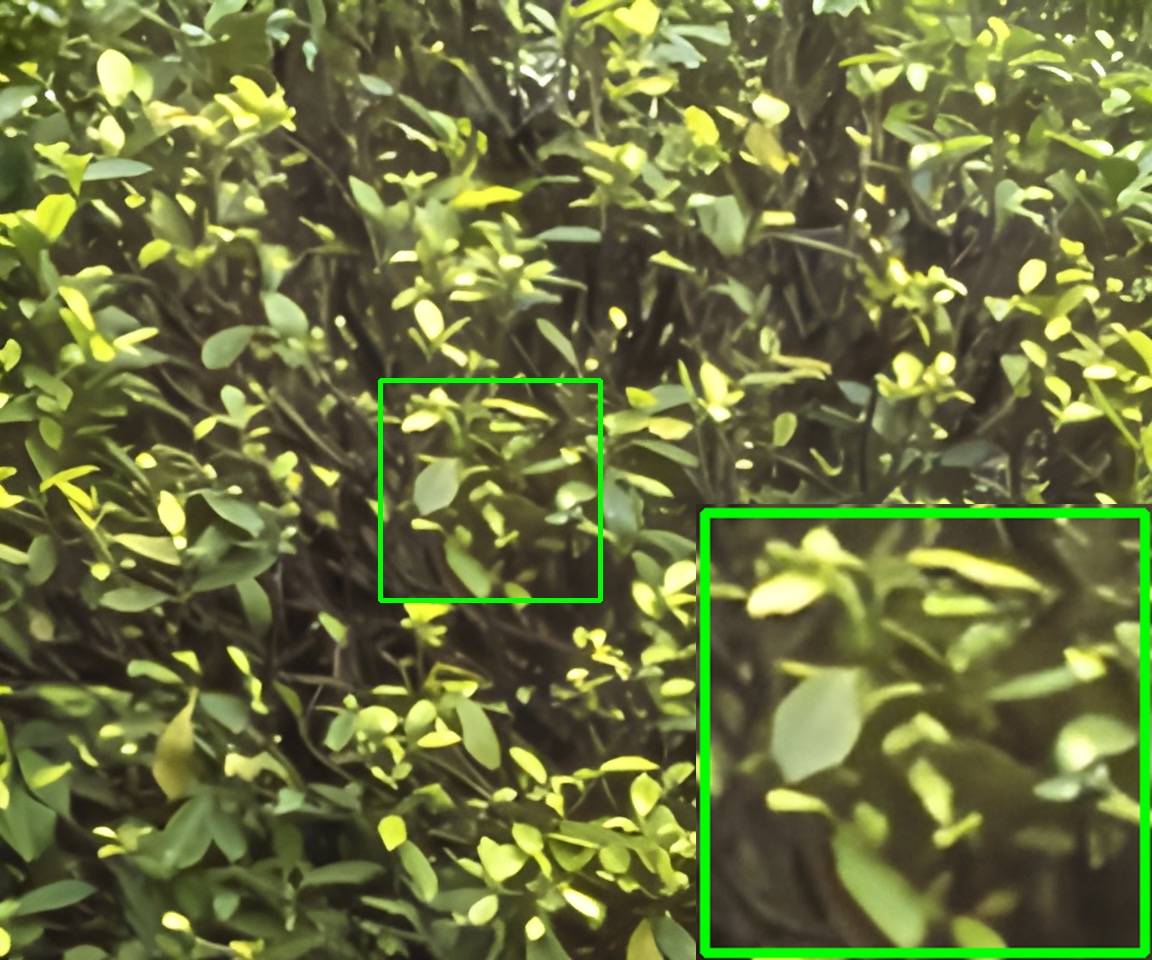}
    \includegraphics[width=\imgwidth]{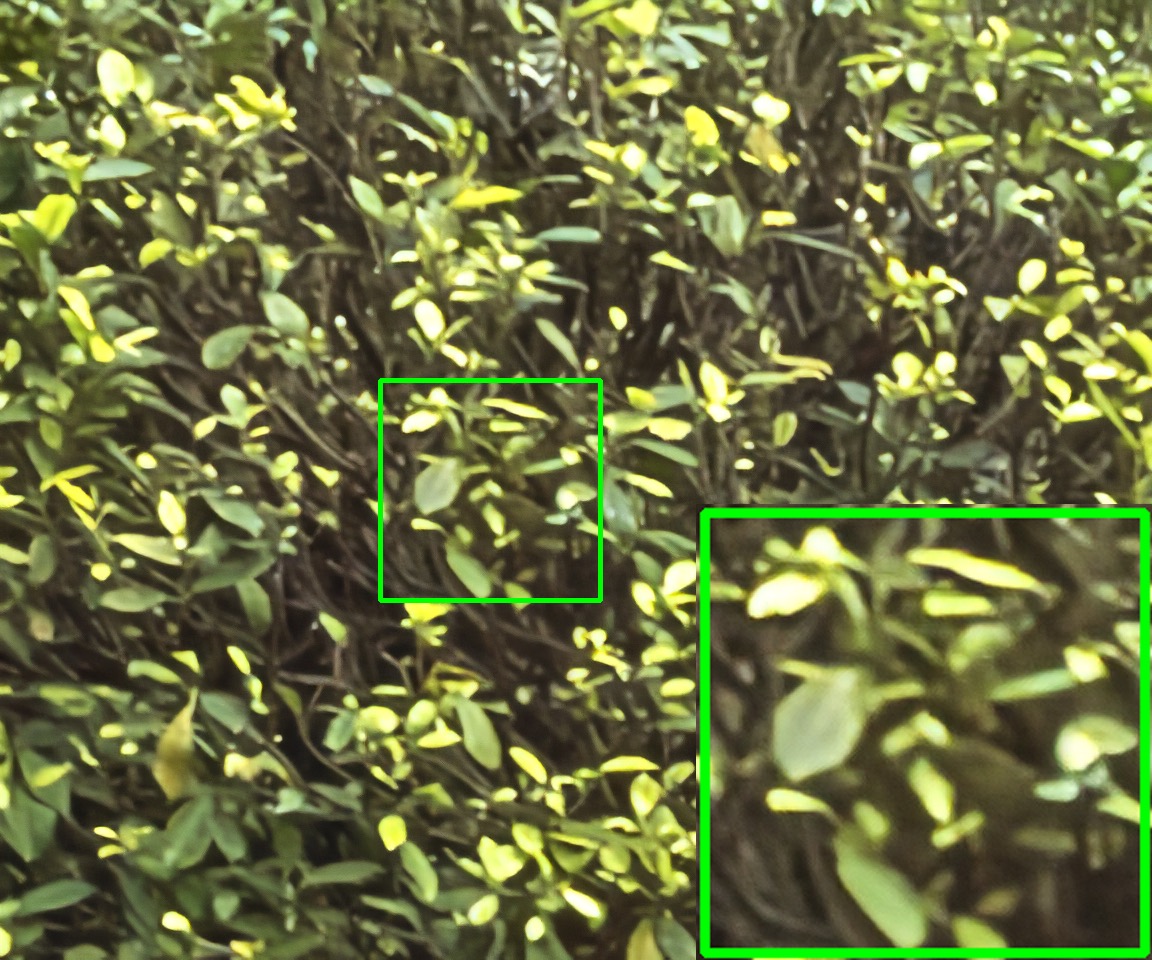}
    \\
    \makebox[\imgwidth]{(a) Real-world LQ input }
    \makebox[\imgwidth]{(b) Real-ESRGAN \cite{wang2021real}}
    \makebox[\imgwidth]{(c) FeMaSR \cite{chen2022femasr}}
    \\
    \includegraphics[width=\imgwidth]{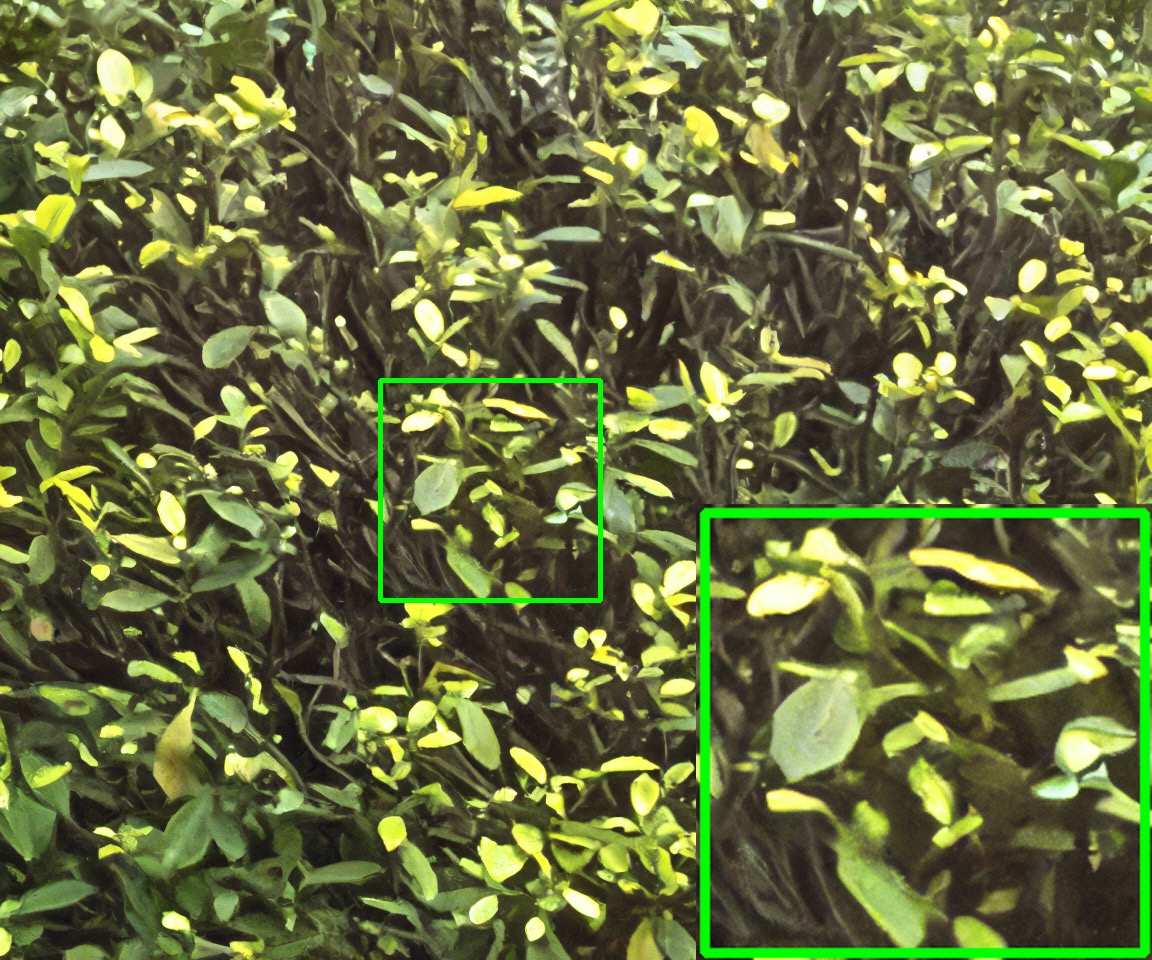}
    \includegraphics[width=\imgwidth]{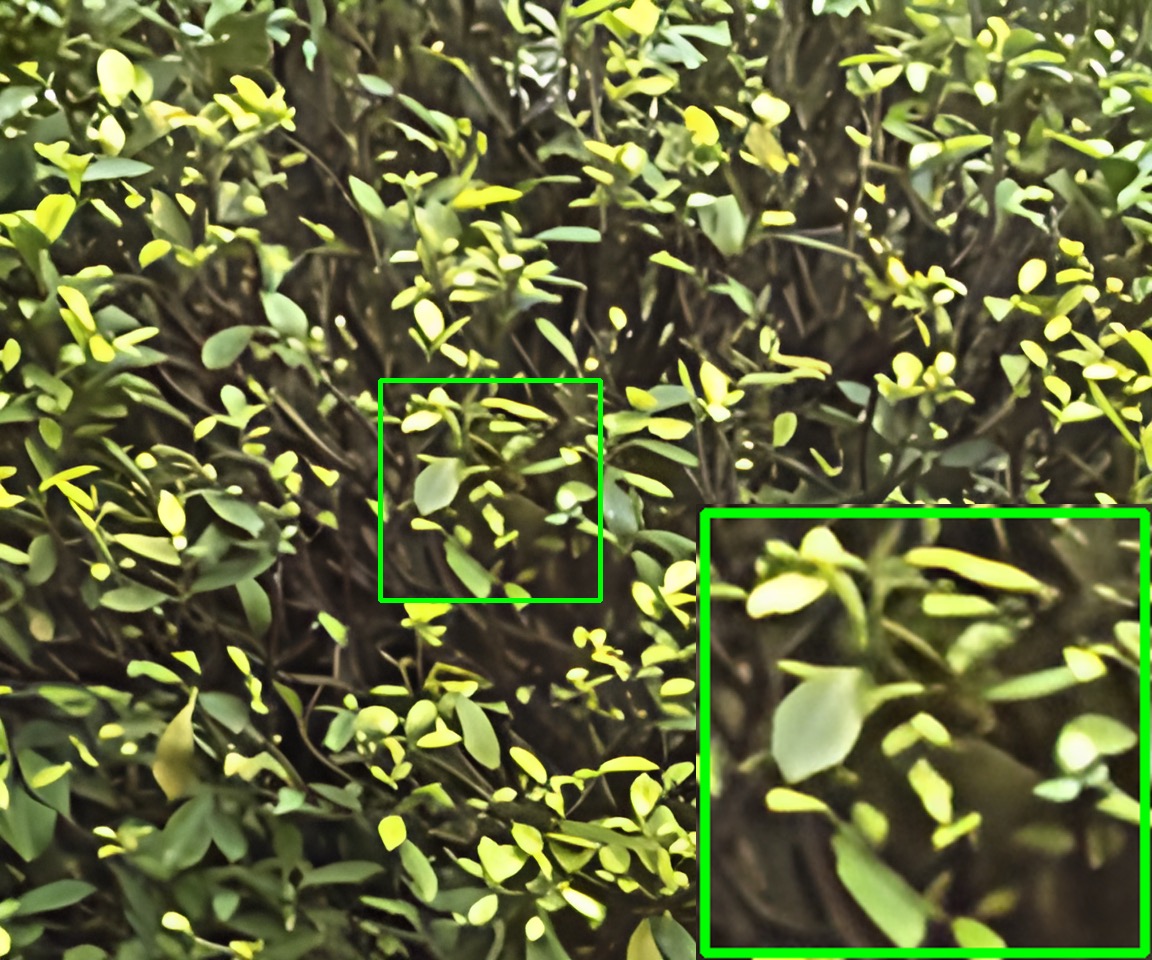}
    \includegraphics[width=\imgwidth]{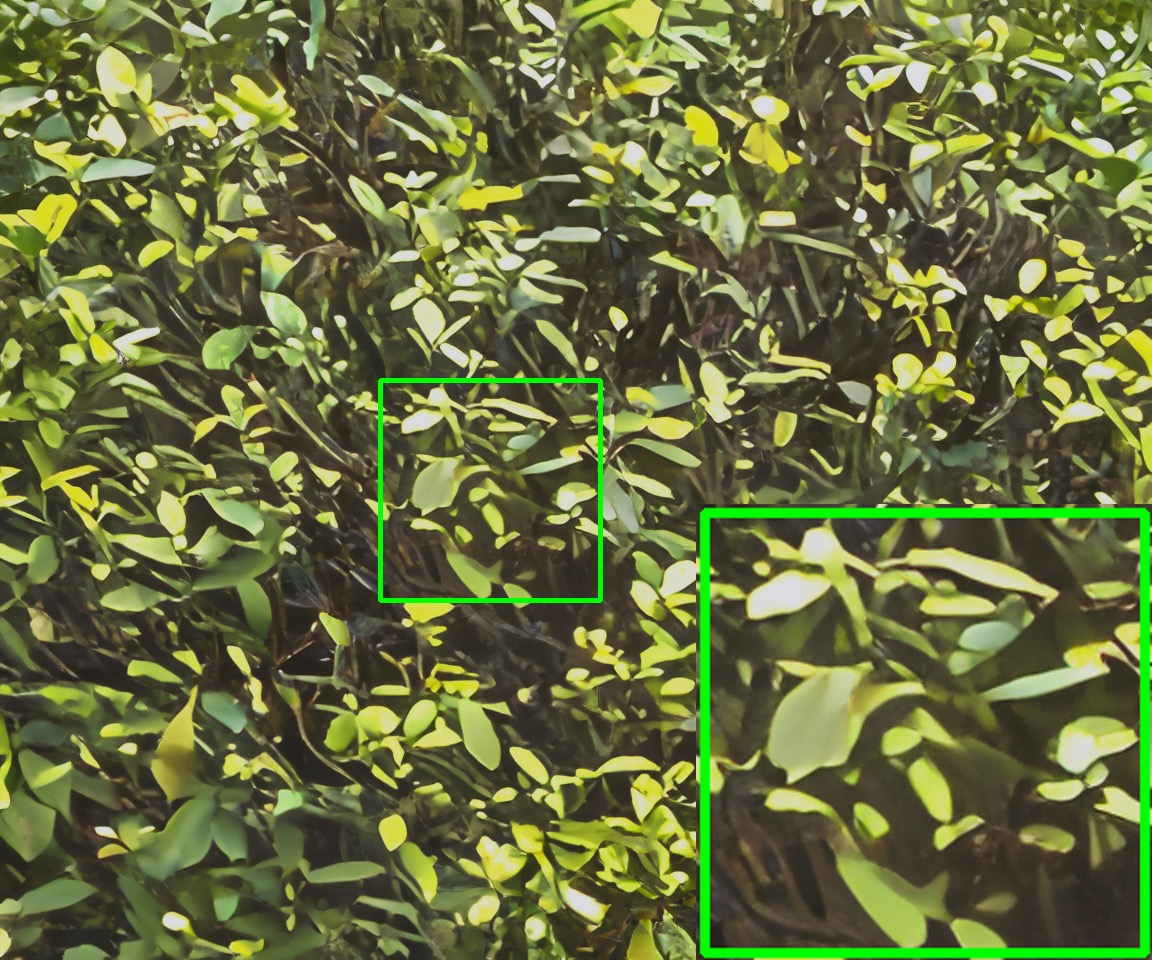}
    \\
    \makebox[\imgwidth]{(d) LDM-BSR \cite{rombach2022latentdiffusion}}
    \makebox[\imgwidth]{(e) MM-RealSR \cite{mou2022mmrealsr}}
    \makebox[\imgwidth]{(f) \textbf{ITER (Ours)}}
    \caption{Additional results from real-world benchmarks.}
    \label{fig:supp_vis2}
\end{figure*}

We show more results on real-world benchmarks in \cref{fig:supp_vis1,fig:supp_vis2}. We can observe that the proposed ITER generates sharper and more realistic textures than competitive approaches.

\section{Limitations}

\begin{figure*}[h]
    \centering
    \newcommand{\imgwidth}{0.24\linewidth}
    \includegraphics[width=\imgwidth]{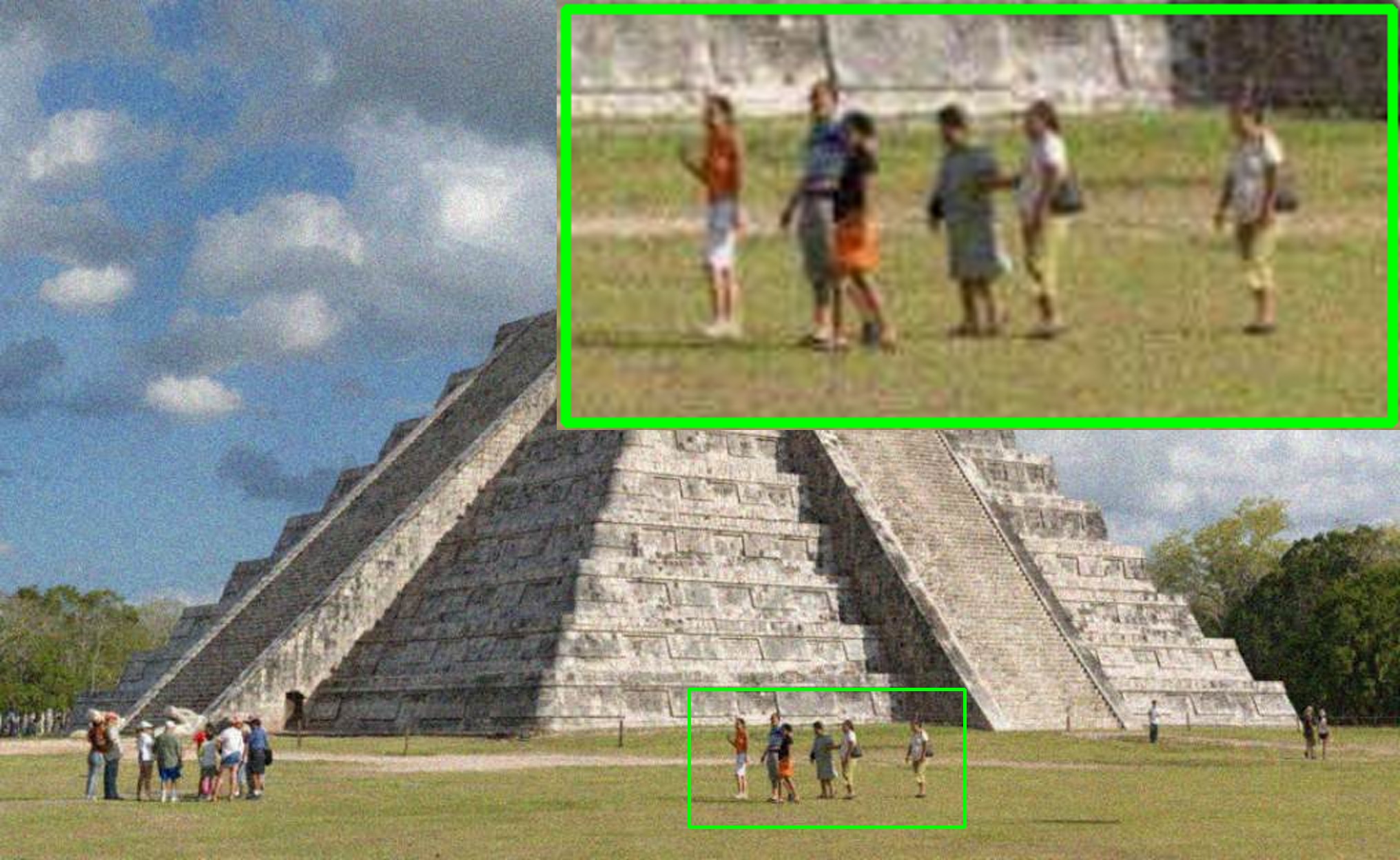}
    \includegraphics[width=\imgwidth]{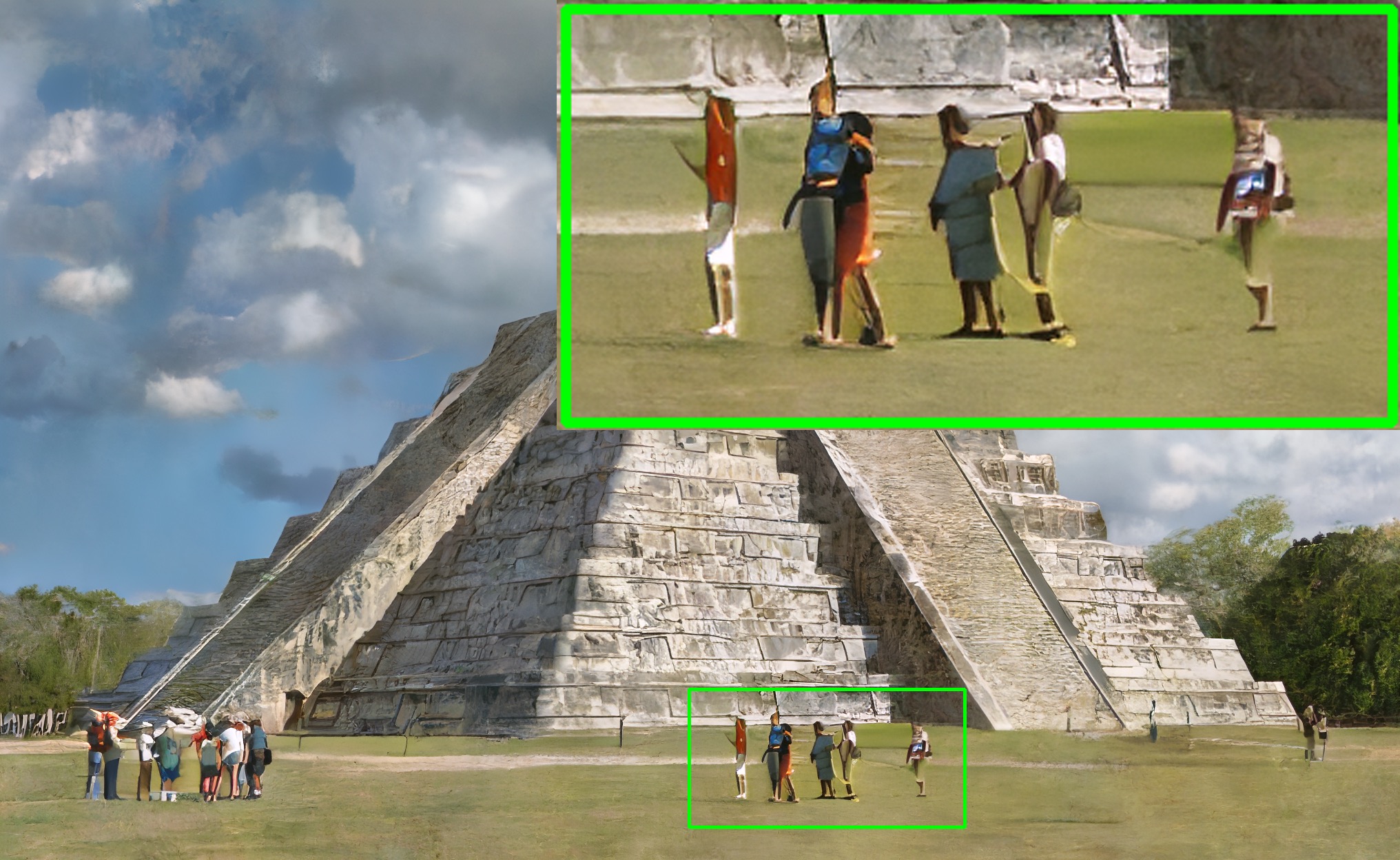}
    \includegraphics[width=\imgwidth]{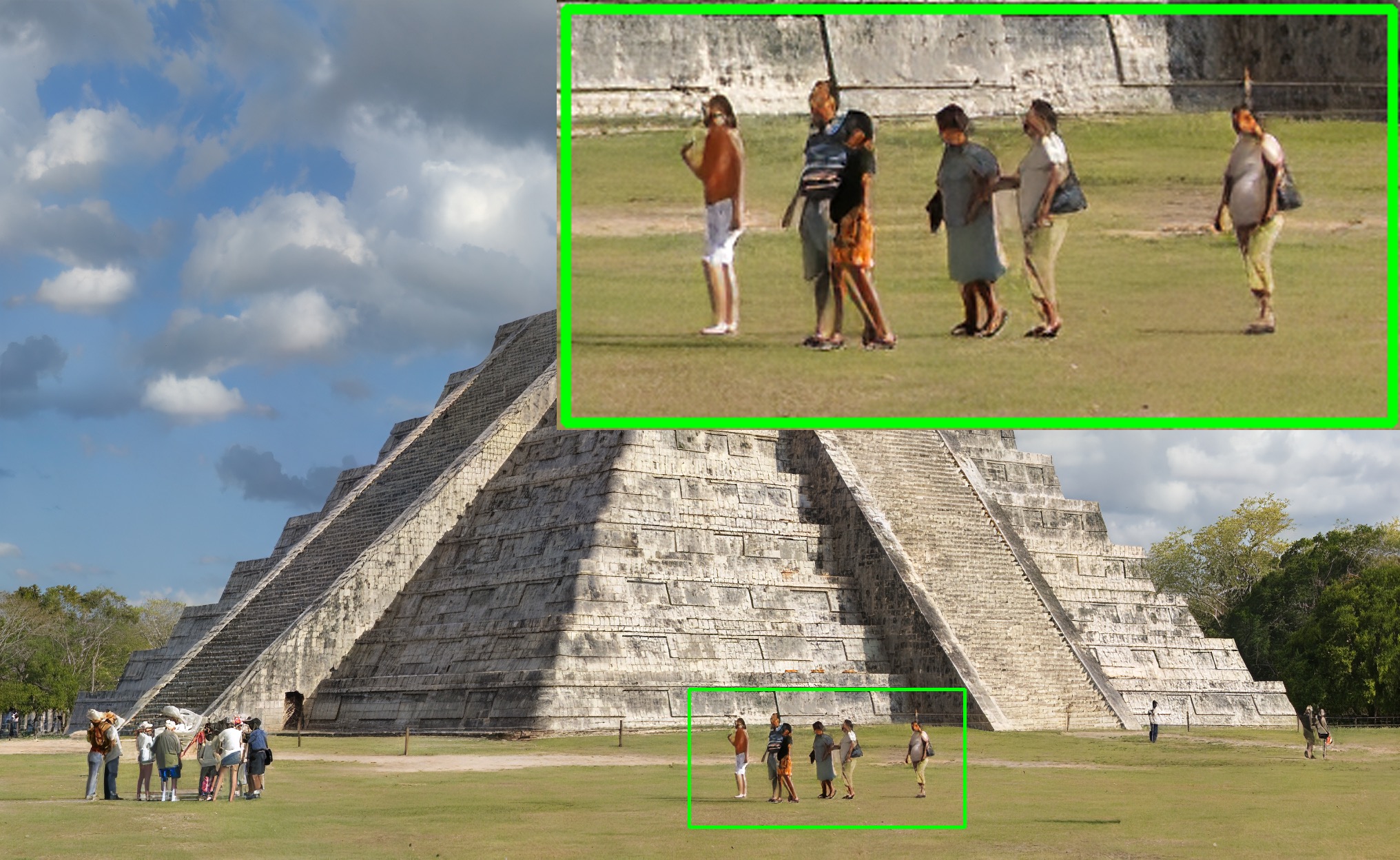}
    \includegraphics[width=\imgwidth]{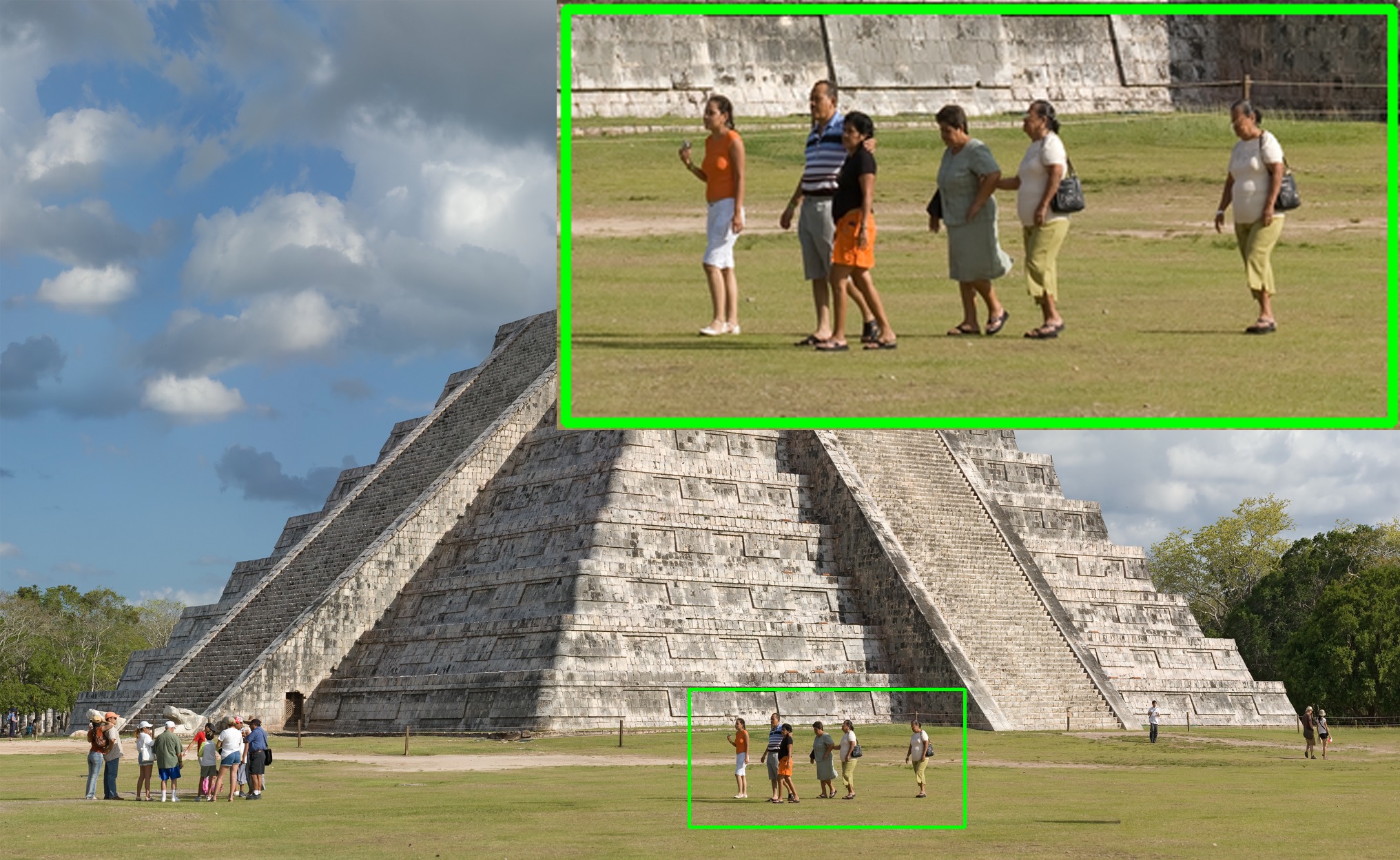}
    \makebox[\imgwidth]{\small LQ input}
    \makebox[\imgwidth]{\small Our result}
    \makebox[\imgwidth]{\small \makecell[c]{Swin-VQGAN Reconstruction \\ (\textbf{Upper Bound})}}
    \makebox[\imgwidth]{\small Ground truth}
    \caption{Limitation of the proposed method.}
    \label{fig:limitation}
\end{figure*}

The upper bound of ITER is limited by the reconstruction performance of VQGAN, \ie, 0.088 LPIPS score in our experiments. This is because VQGAN cannot perfectly reconstruct the HQ images and has information loss when compressing the image to tokens. As shown in \cref{fig:limitation}, the VQGAN has difficulties to reconstruct the small humans at the bottom of the image. Therefore, our method is also not able to recover them even with HQ input.

\end{document}